\documentclass[10pt,journal,compsoc]{IEEEtran}
\ifCLASSOPTIONcompsoc
  \usepackage[nocompress]{cite}
\else
  \usepackage{cite}
\fi
\ifCLASSINFOpdf
\else
\fi
\usepackage{color}
\usepackage{hyperref}       
\usepackage{url}            
\usepackage{booktabs}       
\usepackage{amsfonts}       
\usepackage{nicefrac}       
\usepackage{microtype}      
\usepackage{xcolor}   
\usepackage{amsmath}
\usepackage{amssymb}
\usepackage{graphicx}
\usepackage{wrapfig}
\usepackage{algorithm}
\usepackage{algpseudocode}
\usepackage{soul}
\usepackage{circledsteps}
\usepackage{pifont}
\usepackage{multirow}
\usepackage{svg}

\newcommand{\ye}[1]{\textcolor{black}{\@#1}}
\newcommand{\re}[1]{\textcolor{black}{\@#1}}
\newcommand{\minr}[1]{\textcolor{black}{\@#1}}

\begin{document}
%
\title{Vision + X: A Survey on \\ Multimodal Learning in the Light of Data}
%
%
%
%

\author{Ye Zhu,~\IEEEmembership{Member,~IEEE,}
        Yu~Wu,~\IEEEmembership{Member,~IEEE,}
        Nicu~Sebe,~\IEEEmembership{Senior Member,~IEEE,}
        and~Yan~Yan,~\IEEEmembership{Senior Member,~IEEE}
\IEEEcompsocitemizethanks{\IEEEcompsocthanksitem Y. Zhu is with the Department
of Computer Science, Princeton University, Princeton, NJ 08540 USA. E-mail: yezhu@princeton.edu
\IEEEcompsocthanksitem Y. Wu is with the School
of Computer Science, Wuhan University, China.
\IEEEcompsocthanksitem N. Sebe is with the Department
of Information Engineering and Computer Science, University of Trento, Italy.
\IEEEcompsocthanksitem Y. Yan is with the Department
of Computer Science, Illinois Institute of Technology, Chicago,
IL 60616, USA.
}
\thanks{Manuscript received October, 2022.}}

%
%

\markboth{Journal of \LaTeX\ Class Files, 2022}%
{Shell \MakeLowercase{\textit{et al.}}: Bare Demo of IEEEtran.cls for Computer Society Journals}
%



\IEEEtitleabstractindextext{%
\begin{abstract}
We are perceiving and communicating with the world in a multisensory manner, where different information sources are sophisticatedly processed and interpreted by separate parts of the human brain to constitute a complex, yet harmonious and unified sensing system. 
To endow the machines with true intelligence, multimodal machine learning that incorporates data from various \re{sources} has become an increasingly popular research area with emerging technical advances in recent years.
In this paper, we present a survey on multimodal machine learning from a novel perspective considering not only the purely technical aspects but also the intrinsic nature of different data modalities. 
We analyze the commonness and uniqueness of each data format \re{mainly} ranging from vision, audio, text, and motions, and then present the \re{methodological advancements} categorized by the combination of \re{data modalities, such as \emph{Vision+Text}, with slightly inclined emphasis on the visual data}. 
We investigate the existing literature on multimodal learning from both the representation learning and downstream application levels, and provide an additional comparison in the light of their technical connections with the data nature, \textit{e.g.}, the semantic consistency between image objects and textual descriptions, and the rhythm correspondence between video dance moves and musical beats.
We hope that the exploitation of the alignment as well as the existing gap between the intrinsic nature of data modality and the technical designs, will benefit future research studies to better address a specific challenge related to the concrete multimodal task, prompting a unified multimodal machine learning framework closer to a real human intelligence system. 
\end{abstract}

}

\maketitle

\IEEEdisplaynontitleabstractindextext

%
\IEEEpeerreviewmaketitle

\section{Introduction}
\label{sec:intro}

\IEEEPARstart{W}{e} perceive and communicate with the world through a multisensory human system, by seeing objects, hearing sounds, speaking languages, as well as writing and reading texts. 
The information from these varied sources is processed by different parts of the human brain, as indicated by~\cite{wallace2002histochemical,alain2001and,bornkessel2015neurobiological}. 
For instance, the occipital lobe acts as the primary center for visual processing, interpreting the distance and locations of objects, while the temporal lobe processes auditory information, helping us understand sounds. Language comprehension, facilitated by Wernicke's area in the posterior superior temporal lobe, is crucial for decoding both written and spoken words. Other sensory information, such as touch and movement, is processed by distinct brain areas.
These integrated yet distinct functions form a complex and harmonious human sensing system.
The specialized divisions in human neural processing, which highlight both unique and shared characteristics across different modalities, inspire us to think about the multimodal machine learning problem in the light of data in this paper.

Historically, the vision, audio and textual data were usually studied in separate research fields (\textit{i.e.}, computer vision, digital signal processing, and natural language processing).
With the ultimate objective to bring true intelligence to machines, the research in Artificial Intelligence (AI) nowadays has gone far beyond the exploitation of a single perception perspective but entered an era where the interplay of multiple sensing systems is studied in a collaborative way just as in human brain systems.
As the research of multimodal learning has become increasingly popular in recent years, we present a survey that not only studies the technical development of recent literature but also elaborates on the data characteristics, as well as examines the connections between the logic of such technical designs and their respective data natures. 

\begin{figure*}[th]
    \centering
    \includegraphics[width=0.98\textwidth]{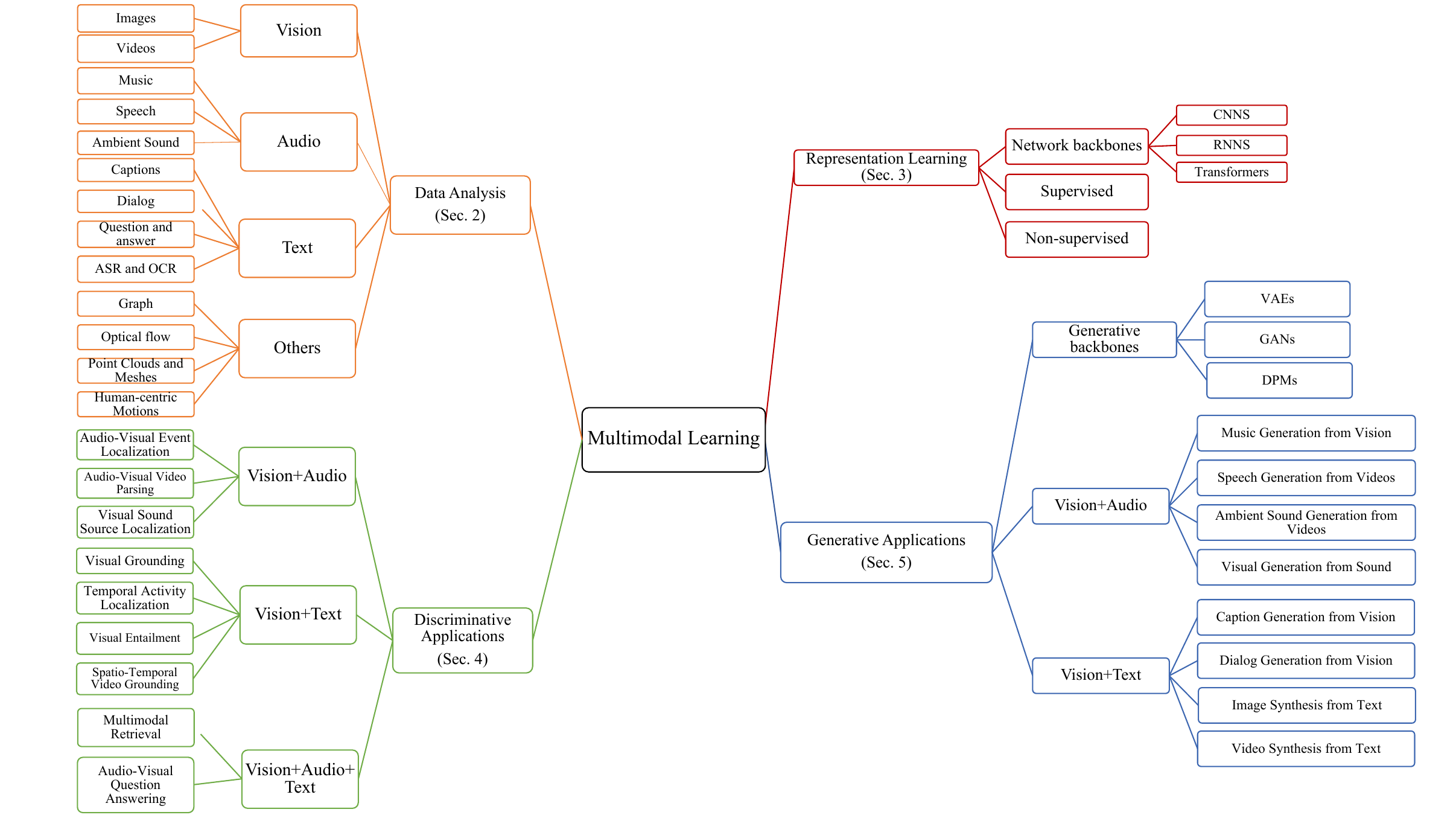}
    \caption{\textbf{Overall structure of our survey.} We first present different data modalities and their characteristics, along with examples of multimodal datasets. We then introduce the representation learning area categorized with learning settings. Next, we mainly divide the application area into the discriminative and the generative directions, with more detailed classifications following the combination of data modalities.}
    \label{fig:structure}
    \vspace{-0.1in}
\end{figure*}

To better structure the paper, we adopt a taxonomy centered on computer vision literature, using vision as the primary data modality while incorporating others, including \textit{audio}, \textit{text}, and others.
These modalities share commonalities but also have unique characteristics in terms of nature, format, and assessment criteria. For example, audio data can be categorized into music, speech, or ambient sounds, with speech closely linked to text and music often associated with motion in a more subjective manner. We then discuss multimodal representation learning, differentiating between \textit{supervised} and \textit{unsupervised} settings, alongside popular network architectures for processing various modalities. This categorization underscores a shift in research focus from traditional supervised learning with manually annotated data to large-scale pre-training on unlabeled data.

Subsequently, we delve into the downstream aspects of multimodal learning, categorizing multimodal applications into two primary directions: discriminative and generative applications.
For each direction, we group the existing literature in the form of \textbf{vision+X}, where \textit{X} primarily represents non-visual data modalities. This framework highlights the adaptability and practicality of multimodal learning across diverse scenarios.
For instance, akin to human multisensory perception, combining vision and language is essential for tasks like captioning, which provides textual descriptions of visual content, or imagining sounds based on visual cues. Reviewing various prominent multimodal tasks reveals that, despite differing data modalities and objectives, shared technical approaches emerge. Our detailed analysis examines these technical intricacies and their reflection on underlying data properties, strengthening the link between data types and model strategies. This exploration also addresses prevailing challenges and future directions in multimodal learning.

Compared to other surveys on multimodal learning~\cite{baltruvsaitis2018multimodal,guo2019deep,jaimes2007multimodal,liang2022foundations,bayoudh2021survey,xu2023multimodal}, we approach the problem from the unique perspective of data itself.
This novel perspective allows us to establish connections between the inherent characteristics of multimodal data and the design of methodologies, leading to a profound discussion on the future of multimodal research in two major aspects.
On the one hand, we believe that emphasizing and exploiting the unique characteristics of specific data modalities will contribute to solving concrete application problems associated with those modalities.
On the other hand, recognizing the commonalities among different modalities will enable researchers to construct a more unified and collaborative framework that mirrors the capabilities of a real human intelligence system.

The overall structure of the paper, illustrated in Fig.~\ref{fig:structure}, is as follows:
In Sec.~\ref{sec:data}, we first provide an analysis in terms of data characteristics for different modalities with a focus on vision, audio, and text.
Next, we explore the multimodal representation learning in Sec.~\ref{sec:representation}, sub-categorized by the current popular model architectures and different learning settings.
In Sec.~\ref{sec:discriminative} and~\ref{sec:generative}, we present concrete multimodal applications with discriminative and generative tasks, respectively.
In addition to the task-wise and technical introduction, we also make extra efforts by connecting the existing literature with their data characteristics as mentioned in Sec.~\ref{sec:data}, revealing which data attribute is handled and addressed in specific methods and models.
The above revisit forms the basis for our discussions on the existing challenges and possible future directions in Sec.~\ref{sec:future}.
Sec.~\ref{sec:conclusion} includes final remarks and conclusions.

\vspace{-0.15in}
\section{Data Analysis}
\label{sec:data}
In this section, we elaborate on the intrinsic nature of multiple data modalities by analyzing their characteristics and commonalities.
A list of commonly used multimodal datasets is included in Appendix~\ref{sec:dataset} with detailed descriptions.

\vspace{-0.1in}
\subsection{Vision}
\label{subsec:vision}

We categorize visual data into images and videos. As a primary information source in both human sensory systems and computer vision literature, visual data is often considered "raw data" with its high dimensionality. It encompasses a wealth of features and details, representing rich visual content. However, the redundancy in continuous spatial and temporal aspects poses challenges for processing, analysis, and efficient utilization in multimodal learning tasks.

\noindent \textbf{Images.}
Images are fundamental to computer vision research, characterized by their inherent invariance to transformations. This key attribute drives the development of classic image processing methods and deep learning techniques like CNNs to extract meaningful visual features.

In the pre-deep learning time, image processing and computer vision research primarily aimed at deciphering image content and patterns through a manual feature extraction and analysis pipeline using machine learning techniques.
For example, the scale-invariant feature transform (SIFT)~\cite{lowe2004SIFT} and the histogram of oriented gradients (HOG)~\cite{dalal2005histograms}, and Speeded-Up Robust Feature (SURF)~\cite{bay2006surf} are three examples of popular image feature descriptors largely used for computer vision and image processing. After having extracted those descriptive features, some machine learning algorithms such as Support-Vector Machines (SVM)~\cite{cortes1995support} and Principal Component Analysis (PCA)~\cite{wold1987principal} are used to further analyze the feature data.
With the rapid development in the deep neural network architectures~\cite{krizhevsky2012imagenet,simonyan2014very,he2016deep} and the availability of large-scale image datasets such as ImageNet~\cite{deng2009imagenet,russakovsky2015imagenet}, computer vision entered a new era where the classic procedure of feature extraction and analysis has been automatically integrated into neural network designs. 

Moreover, the applications of computer vision in the image domain have been extensively extended and enriched from simple image classifications~\cite{simonyan2014very,simonyan2014very} to various task scenarios like object detection~\cite{zhao2019object} and segmentation~\cite{long2015fully} within the image.
In addition to the above-mentioned discriminative task applications that aim to mine the data patterns from existing images, there is another branch of applications that aim to synthesize the image data using generative neural networks.

\noindent \textbf{Videos.}
Video is another form of common visual data that has been extensively studied in the computer vision community~\cite{perazzi2016benchmark,xu2016msr,oh2011large}.
\re{Unlike static images, videos encapsulate information across the temporal dimension.}
For instance, human actions in videos are usually defined by a series of specific movements depicted in consecutive video frames over time, as such consistency and transformation in the visual context can only be presented in the format of videos. 
This temporal characteristic of video data also influences \re{video-based applications, which often require additional understanding and analysis of temporal elements} (\textit{e.g.}, actions, motions, optical flow)~\cite{cheng2017segflow,sun2018optical}.
While the conventional image representation encoded by neural networks can be applied to individual frames, extracting video representations necessitates addressing the connections between temporally related frames.
An intuitive and classic approach to learning video data representation is to extend the conventional 2D convolutional neural network into 3D architectures with an additional temporal dimension, a notable example is the I3D model~\cite{i3d} proposed for action recognition in videos.

Regarding video-based applications, the tasks are similar to those in the image domain, where the most popular discriminative tasks include video classification (sometimes referred to as action recognition)~\cite{i3d} and segmentation~\cite{tsai2016video}, and the generative tasks that seek to directly synthesize videos~\cite{tulyakov2018mocogan}.
For the latter, Sora from OpenAI~\cite{sora2024} stands out as a most recent large video generator.

\vspace{-0.1in}
\subsection{Audio}
\label{subsec:audio}

Traditionally, the study of audio processing has predominately resided within the research field of digital signal processing. In this \re{survey}, we focus on introducing three primary types of audio data: speech, music, and ambient sound. Each of these audio types holds relevance and applicability in various multimodal task applications, further emphasizing the diverse nature of audio data within the context of multimodal learning.
Similar to visual data, audio signals are a form of ``raw data" that can be directly captured from the environment. 
However, unlike static images, audio signals possess inherent continuity in the temporal dimension.

\begin{figure}[t]
    \centering
    \includegraphics[width=0.48\textwidth]{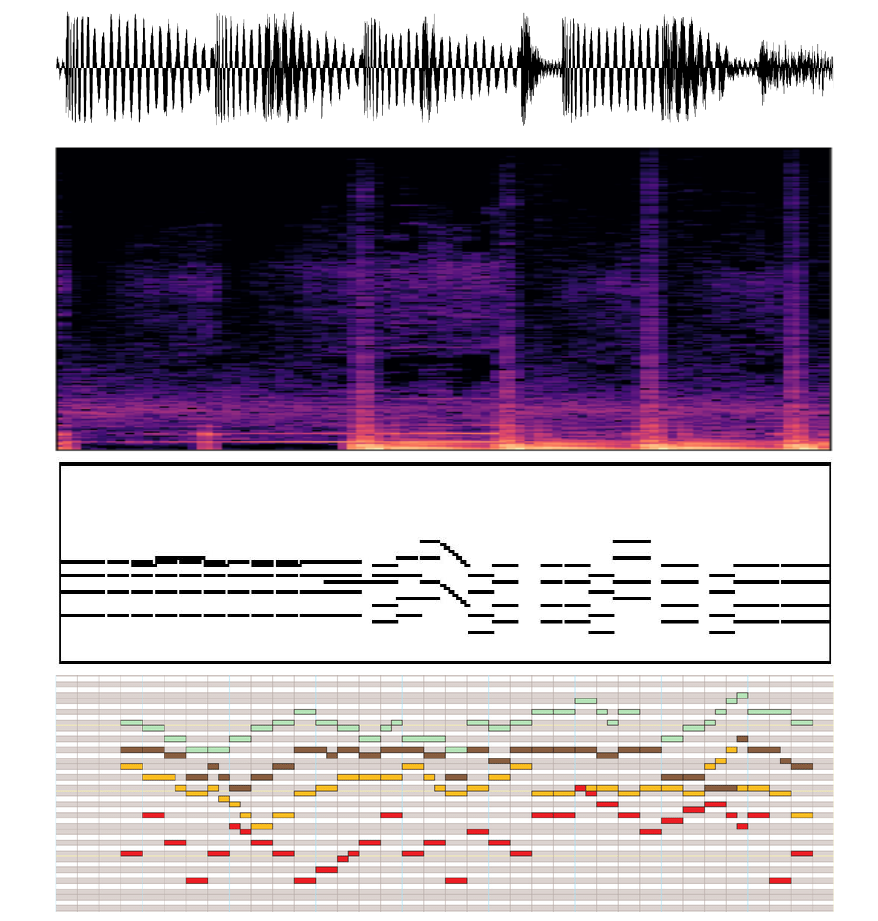}
    \caption{\textbf{Illustrations of different audio data representations.} From top to bottom: (a) raw audio data in waveform; (b) audio data in mel-spectrogram; (c) music piece in 1D piano-roll from~\cite{dong2018pypianoroll}, where the horizontal and vertical axis represent the timestamps and the audio pitch, respectively; (d) music piece in MIDI from~\cite{inbook}, where colors represent different instrument types.}
    \label{fig:audio_representation}
    \vspace{-0.15in}
\end{figure}

\noindent \textbf{Music.}
Music is a specific type of audio data that plays a significant role in our daily lives. As a form of expressive art, music is considered as the carrier and reflection of one's inner world. 
Generally speaking, the music itself has various genres such as traditional classical music, symphony, modern pop, country music, etc. 
\re{Furthermore, music can also be classified into diegetic and incidental categories. Diegetic music is integral to the narrative, existing within the story's universe and perceived by its characters. In contrast, incidental music is intended solely for the audience's experience, underscoring emotions and scenes without being part of the story's world.}
The music categorization can often be subjective, with less strict and rigorous taxonomy for a specific genre. 
One common feature of those genres is that a music piece with high auditory quality has a relatively large sampling rate. For example, for the music with CD quality, the sampling rate is 44.1kHz~\cite{jukebox}, which leads to over 2 million data points for a one-minute musical piece. 

From the perspective of scientific research, the high dimensionality of musical audio waveform imposes difficulties in data processing, therefore, researchers have developed different forms of musical data representations.
In this survey, we classify the musical data representations into \textit{``non learning-based"}  and \textit{``learning-based"} depending on whether the representations are obtained via deep learning techniques. 
For the \textit{``non learning-based"} musical representations, we can further divide them into continuous and discrete subcategories.
The most general non-learning based continuous data format for audio, including the music, is the waveform as shown in Fig.~\ref{fig:audio_representation}(a).
The waveform is a two-dimensional data that depicts the sound pressure variation measured by the air vibration in the time domain. 
Another popular and general type of audio representation is the spectrogram. \re{Compared to the waveform that emphasizes the temporal variations of the audio signals, spectrogram also reflects the frequency content of sounds over time, as illustrated in Fig.~\ref{fig:audio_representation}(b).}
In most cases, we refer to the waveform as the raw audio data.
1D piano-roll~\cite{dong2018pypianoroll} and 2D  Musical Instrument Digital Interface (MIDI) are classic discrete representations~\cite{briot2017deep}.
As illustrated in Fig.~\ref{fig:audio_representation}(c), piano-roll is a sparse data representation format, where the horizontal axis is the time stamps, and the vertical axis represents the acoustic pitch.
2D MIDI can be interpreted as a composed piano-roll format with instrument types, as represented by different colors in Fig.~\ref{fig:audio_representation}(d).
Both 1D piano-roll and 2D MIDI discrete forms can be decoded back to raw audio space by pre-defined music synthesizers. 
On the other hand, the \textit{``learning-based"} representations can be similarly specified by their discrete and continuous natures.
Recent progress in deep learning has introduced novel learning-based discrete representations, namely the \textit{\re{vector quantization} (VQ)}, to further reduce the high dimensional data into a discrete token space~\cite{vqvae,vqvae2}. 
Continuous learning-based musical representations share similar attributes as those in vision data, where we usually adopt \re{neural networks such as CNNs} to encode the raw audio signals to an embedding feature with desired dimensions. 

Compared to other data modalities, music audio signals have several unique features to consider while applied to specific downstream tasks. 
Firstly, the musical data is a sequence where the temporal coherence within a complete music piece should be emphasized.
Also, in addition to the temporal dimension, the audio data are often characterized by their frequency features in the form of spectrogram representation.
Other than temporal coherence, rhythm is another important unique music characteristic to consider while assessing the music quality.

\noindent \textbf{Speech.}
Speech primarily refers to audio signals of spoken languages, closely related to natural languages with their intrinsic correspondence. The data representations for speech are similar to music, where the waveform and the spectrogram are commonly used types in the non-learning based category.
However, one noticeable uniqueness of speech audio lies in its natural correlation to language, where discrete representations of speech audio align with language tokens. As a result, the discrete language token-driven representation of speech features a more unified format compared to the learning-based VQ representation used in music audio. This characteristic also influences the methodology design when it comes to architecture selection, as detailed in the following of the paper.



\re{In the application areas of speech, classic tasks like speech separation, which aims to isolate individual speech tracks from a composite audio mix, are well-studied~\cite{bahmaninezhad2019comprehensive}. Another pivotal task is automatic speech recognition (ASR)~\cite{yu2016automatic}, which focuses on converting spoken language into text. ASR systems are designed to accurately transcribe human speech, making them essential for voice-activated interfaces and transcription services.}
At the same time, due to the intrinsic correspondence nature between speech and language natures, speech data are often applied in multilingual translation problems~\cite{di2019must} or cross-modal translation between the audio and text~\cite{chung2018unsupervised}.
More recent works in the multimodal generative area also look into the speech generation from visual input with talking lips~\cite{kim2021lip}.

\re{There exists another special type of speech-language for the hard-of-hearing and speech-impaired community, which is the ``sign language''. 
Unlike audible speech, sign language necessitates the interpretation of visual signals from gestures, thereby exhibiting natural connections with visual and motion data. Research focused on sign languages explores tasks such as sign language recognition and generation~\cite{bragg2019sign,rastgoo2021sign}. Essentially, sign language recognition aims to translate specific hand gestures into textual data, whereas generation addresses the reverse process. Although sign language is discussed within the ``speech'' category, the datasets~\cite{akash_nagaraj_2018,duarte2021how2sign} typically comprise visual data, such as images and videos, accompanied by language annotations.
}

\noindent \textbf{Ambient Sound.}
Beyond speech and music, there are other types of audio signals, such as the sound accompanying certain events, which we refer to as \textit{``ambient sound"} in this survey.
Compared to music, which is subjective, and speech, which is closely related to natural languages, ambient sound is more frequently encountered in conjunction with videos to characterize specific actions and events. 
For instance, we naturally relate the sound of a crying baby with a video showing the corresponding visual scenes. This unique correspondence enables ambient sound to provide additional information to conventional video action recognition tasks, leveraging the audio modality~\cite{tian2018audio,huh2023epic}. The aforementioned audio representations are also applicable to ambient sound.

However, in contrast to music and speech audio signals, ambient sound exhibits a more noisy nature with less preprocessing. Unlike music, which can be represented using highly processed data formats like musical MIDI, and speech, which benefit from natural correspondence to text, the representation of ambient sound is more ambiguous. It lacks explicit formats like discrete token representations in speech or specific features like rhythm in music. These characteristics contribute to the inherent ambiguity and challenges in representing ambient audio. 


\vspace{-0.1in}
\subsection{Text}
\label{subsec:text}
Text has been studied within the Natural Language Processing (NLP) communities from the early years. 
\re{While there exist diverse formats of textual data, in this survey of multimodal learning, we mainly focus on introducing several types of textual data that exhibit close connections with other data modalities.}
The NLP community has witnessed significant attention in recent years, especially with the remarkable success in developing Large Language Models (LLMs) such as GPT-3~\cite{radford2019gpt3}. The tremendous achievements in NLP are closely linked to the nature of textual data and language. Unlike visual and audio information, which can be considered as ``raw data'', textual data undergoes substantial processing. More specifically, it is a data type that has evolved through human civilizations, characterized by a highly unified format and precise semantics despite linguistic differences. It signifies the fact that text is highly informative and compact, while visual and audio signals usually contain abundant information redundancy. 
Another unique characteristic of text on the application side is that the problem formulation of most NLP tasks can be unified under the notion of ``next word token prediction". This formulation represents a common underlying structure in various NLP tasks, which contributes to the coherence and consistency within the field and its potential to solve multiple tasks via a large foundation model~\cite{bommasani2021opportunities}.


\noindent \textbf{Captions.}
\re{Captions provide sentence descriptions that summarize either the entirety or a portion of the visual content in vision and text-related multimodal works. They may consist of single sentences or extend into longer paragraphs composed of multiple sentences.}
Bag-of-Words (BOW)~\cite{mikolov2013efficient} is a classic form of text representation, representing a text corpus as a multiset (\textit{i.e.}, a bag) of its words.
Following the development in deep learning, the captions are also frequently processed by Recurrent Neural Networks (RNNs) (\textit{e.g.}, LSTM~\cite{hochreiter1997long}) with an internal memory state to obtain the learning-based representations. 
Compared to the visual data processed by CNNs, the memory state design in RNNs allows for establishing the recurrent connections among consecutive words in a given sentence to better interpret the overall textual features.
A more recent breakthrough in the NLP community is the success of BERT (Bidirectional Encoder Representations from Transformers)~\cite{devlin2018bert}, which is a large-scale pre-trained model for word embedding.
This unique characteristic has been widely addressed in the subsequent research studies involved with captions.

\noindent \textbf{Dialogue.}
\re{Dialogue is another common form of textual data in multimodal machine learning, distinct from captions due to its inherently interactive nature, which involves conversation among participants with logical coherence rather than a unilateral description of visual content.
}
Therefore, when handling dialogue data, special attention should be paid not only to the words within a sentence like plain captions but also to the connections between different sentences within a complete dialogue.
In the literature on multimodal learning for vision and language, these unique characteristics of dialogue are often addressed by constituting an additional data component - history - in the design of the framework. This component typically captures the flow of the conversation, including prior exchanges, and is processed by dedicated mechanisms that operate alongside the processing module for individual sentences.

\noindent \textbf{\re{Question and Answer.}}
\re{A more specific categorization of textual data is \emph{question and answer}.  While its representation is overall similar to its other textual counterparts (tokens), they are often leveraged in vision-language tasks as an approach to study the visual reasoning ability of networks or to evaluate specific task performance. 
Visual question answering (VQA)~\cite{antol2015vqa} is a representative task that uses question and answer to reason the visual context. Question and answer are often closely related to the dialogue, as the interactions within the dialogue can take the form of questions and answers.}

\noindent \textbf{\re{ASR and OCR Text.}}
\re{While captions and dialogue are correlated with the visual context in a more high-level and semantic manner, automatic speech recognition (ASR) and optical character recognition (OCR) based texts present a slightly different connection to the audio and visual information. Specifically, ASR and OCR are foundational multimodal research topics that have matured over decades, they exhibit precise correspondence between text and other data modalities~\cite{gandhi2022esb,javed2022towards}.  
In addition, OCR also serves as a technique to obtain textual data from textual corpus.
}


\vspace{-0.12in}
\subsection{Other Modalities}
\label{subsec:others}

Multimodal learning encompasses a wide array of data modalities beyond vision, audio, and text. For instance, 3D data represents a significant category, including subcategories like point clouds and meshes. This survey focuses on exploring data modalities with cognitive significance that mirror the human perceptual system. Therefore, we categorize data modalities other than vision, audio, and text together, highlighting their relationships with these primary modalities for a more integrated understanding and presentation structure.

\noindent
\textbf{\re{Graph.}}
\re{Graph data offers structured representations of relational information via nodes and edges which captures the connections and interactions between elements. 
While it may not be a data modality that naturally exists through a human perception system, it plays a significant role in machine learning when connected to other data modalities. For example, the scene graph establishes a graphical representation from the images to interpret the connections among objects. A typical example of a multimodal application that leverages both visual and graph data is the scene graph generation from the visual context~\cite{lu2016visual,li2017scene}.  
The non-Euclidean nature of graph data also inspires the design of graph neural networks~\cite{wu2020comprehensive,yuan2022explainability}, which serve as a powerful model architecture to process graph data.
}

\noindent \textbf{Optical Flow.}
The concept of optical flow has been first proposed in the last century as a measurement to characterize the movement of objects in a visual scene caused by the relative motion between an observer and a scene~\cite{horn1981determining}. 
With the advancement of computer vision, especially with the deep learning techniques, the optical flow has also been studied together with visual data~\cite{walker2015dense,wang2019hallucinating,wang2019unos,teed2020raft}.
Compared to other motion data formats, optical flow is usually defined in a more precise manner via the pixel-level change within the consecutive image sequences. 
However, the calculation of optical flow itself has been a quite challenging research problem, due to the fact that environmental lighting can also impose large effects on the pixel values of images. 
Overall speaking, optical flow can be considered as a specific motion presentation explicitly derived from visual information.

\noindent
\textbf{\re{Point Clouds and Meshes.}}
\re{Point clouds and meshes are both important forms of 3D data, providing spatial and structural information that enriches our understanding of physical environments. While point clouds are collections of vertices in a three-dimensional coordinate system, meshes further build on this by connecting points with edges and faces, creating a comprehensive model that represents the shape and topology of 3D objects.
Like the other data modalities discussed in this section, point clouds and meshes are not directly captured by our sensory systems but are constructed through processes that often incorporate human insights. }

\noindent \textbf{Human-Centric Motions.}
Human motions are often used to define various daily activities. 
2D skeleton data of human body is a common representation form for human motions, which captures the keypoints of the human body and represents them as axis coordinates within an image.
They can be used to define various daily activities, which can be practically applied in real-life applications. 
For instance, automatic detection of human actions is particularly valuable in human-centered assistant systems, such as the health assistant to detect falls for elderly people.
Typically, we can extract 2D skeleton data for each frame via pre-trained networks such as OpenPose~\cite{openpose,openpose-cao2017realtime}. 3D human motion data can usually provide richer information with the extra data dimension. 
A classic form of 3D motion data involves incorporating depth information alongside the conventional 2D keypoints data. 
\re{Besides the RGB camera-based methods mentioned for keypoint acquisition, keypoints can also be derived through alternative approaches within the keypoint detection domain, such as geometric reasoning from 3D data~\cite{suwajanakorn2018discovery} and SLAM techniques applied to lidar sensor data~\cite{streiff20213d3l}.}
Furthermore, there are other forms of 3D motion representations that are more frequently adopted in the Computer Graphics (CG) field, such as the Skinned Multi-Person Linear Model (SMPL)~\cite{SMPL:2015}. 
SMPL integrates skinning and blends shapes to represent human bodies.
\re{Compared to optical flow, which captures the motion of all pixels between two frames, keypoint movement, on the other hand, tracks specific points of interest across frames, allowing for a more focused analysis of object or feature dynamics. 
Meanwhile, 3D video features extend this analysis into the spatial domain, integrating depth information with motion to provide a richer, more detailed representation of the visual structure and movement patterns. }



\vspace{-0.15in}
\section{Multimodal Representation Learning}
\label{sec:representation}

In this section, we focus on multimodal representation learning studies.
We structure this section into three parts: the introduction to several popular network architectures and evaluation, the supervised learning setting, and the non-supervised setting.
The rationale behind this categorization is based on the fact that the multimodal representation learning field has gone through a shift from conventional supervised representation to large-scale pretraining. 
Classic methods under the supervised learning setting usually require fully annotated data to train the networks, thus imposing limitations on the size of available training datasets due to the tedious human labor work for labeling. 

To overcome the bottleneck, the research trend in the multimodal representation learning has turned to \textit{``non-supervised"} setting, using data that do not necessarily require human annotations. 
These datasets are often collected directly from the Internet and consist of paired data from different modalities. It is important to note that while these datasets possess intrinsic correspondence between modalities, they are considered non-supervised in this survey due to the lack of manual labeling. Notably, these non-supervised approaches benefit from larger dataset sizes and have witnessed an increase in model scale.
Therefore, in Section~\ref{subsec:non-supervised} for non-supervised representation learning works, we mainly focus on introducing the large-scale pre-training studies, which have been attracting much research attention in recent years.
The \ye{primary research objective in} representation learning under the multimodal context is to learn an effective and discriminative mapping between the corresponding data representations from multiple modalities.


\vspace{-0.12in}
\subsection{Network Architectures}
\label{subsec:rep_learning_main}


\re{We introduce several popular network architecture backbones for learning data representations of the above-mentioned principal data modalities (\textit{i.e.}, visual, audio and text). 
Nevertheless, we note that there exists other popular network designs for specified data modality, \textit{e.g.}, Graph Neural Networks (GNNs)~\cite{scarselli2008gnns} for graph data and PointNet~\cite{qi2017pointnet} for point clouds.
}

\noindent \textbf{Convolutional Neural Networks (CNNs).}
As one of the most classic network architectures in the computer vision field, CNNs~\cite{krizhevsky2012imagenet,simonyan2014very,he2016deep} have been widely adopted as the backbone architecture in representation learning for visual data.
The core idea of CNNs is to extract high level data representations from the raw data via complex functions composed of convolutional layers and activation functions. Similarly, the same idea has also been adapted in learning data representations for audio signals~\cite{hershey2017cnn}.
\re{In the context of representation learning for classic visual and audio data, the training of CNNs often leverages the multi-class cross-entropy loss for classification tasks using:
\begin{equation}
    l = - \sum^N_{c=1} y_c \mathrm{log}(p_c),
\end{equation}
where $y_c$ is the class label, and $p_c$ denotes the predicted probability.
Then features extracted from the last layer of the CNNs are further used as the actual data representation.}

\noindent \ye{\textbf{Recurrent Neural Networks (RNNs).}
A specific demand for learning data representations for natural languages is to consider its temporal correlations with sequential order for words. 
Therefore, NLP community follows a different vein of network architectures to address this challenge using recurrent neural networks (RNNs) and LSTM~\cite{graves2013generating,hochreiter1997long}. 
Efforts have also been made to learn audio data representations via RNNs~\cite{freitag2017audeep}.
}

\noindent \textbf{Transformers.} 
Transformers~\cite{vaswani2017attention} have gained great popularity in machine learning community in both computer vision and natural language processing areas.
The core technical design of Transformers is the self-attention mechanism, which operates on sequential data to learn the overall information.
Compared to CNNs and RNNs, Transformers have several distinctive advantages in model designs: the flexibility to deal with sequential data with variant lengths; the efficiency to allow for parallel computing instead of following the sequential processing as for RNNs.
Despite the fact that the Transformers are initially designed for NLP tasks~\cite{brown2020language}, it has been successfully applied in representation learning for visual~\cite{he2022masked} and audio data~\cite{truong2021right}.

\re{Recently, Mamba~\cite{gu2023mamba} comes out as a new popular model that shows promising downstream performance on long language and audio sequence processing compared to Transformers~\cite{vaswani2017attention}. One of its key strengths is to address the computational challenges by incorporating a selective mechanism into the state space models.}

\re{For the representation learning of textual data, commonly referred to as Language Model Pre-training, a widely used problem formulation involves ``next token prediction'', which frames learning as a joint conditional probability challenge. As a result, the foundational approach for optimizing neural networks in this context frequently involves maximizing the likelihood, typically achieved through the use of cross-entropy loss. However, it is worth noting that various other objective functions have been introduced to further enhance the modeling ability of neural networks.
}

Overall, data representation learning has always been an important research direction in machine learning, and it is a topic that lies within the upstream of the research pipeline.
Consequently, the evaluation of multimodal representation learning methods usually relies on concrete downstream tasks, which we will present in detail in Section~\ref{sec:discriminative}.

\vspace{-0.12in}
\subsection{Supervised Learning}
\label{subsec:supervised}
 Supervised setting requires the annotations from multimodal sources to guide the learning process, which is also the most classic representation learning setting~\cite{yuan2021multimodal,desai2021virtex}.
 
Generally speaking, there are two approaches that are widely adopted for the supervised representation learning.
One possible way is to establish mapping after having obtained the data representations from their respective feature space, which can be considered as a two-stage method denoted `` representation learning in individual modality domain + mapping among modalities", usually with fixed backbone models for the first feature extraction stage~\cite{liu2019use,miech2018learning,wang2021t2vlad}.
Alternatively, another way to tackle the multimodal representation problem is to learn a unified representation for a given data pair in an end-to-end manner, freely optimizing the feature extraction backbones~\cite{wang2018learning,av4-2017look}. 
 
For the first approach, Liu \textit{et al.}~\cite{liu2019use} exploit the existing pre-trained semantic embeddings from visual content, and propose a collaborative experts model to aggregate the multimodal information. 
~\cite{miech2018learning} learns the text-video embedding from heterogeneous data with a Mixture-of-Embedding-Experts (MEE) model.
Wang \textit{et al.}~\cite{wang2021t2vlad} focus on global-local sequence alignment of video and textual representations.
For the second approach,~\cite{wang2018learning} proposes to learn two-branch neural networks for matching the text and image data. In the audio-visual domain,~\cite{av4-2017look} learns the mutual representation via a ``audio-visual correspondence" learning task.

\vspace{-0.12in}
\subsection{Non-supervised Learning}
\label{subsec:non-supervised}

In contrast to supervised learning where exhaustive manual annotations are required in training, there are other paradigms to learn data representation in the multimodal context. In the existing literature, ``unsupervised'', ``weakly-supervised'', and ``self-supervised'' are representative terminologies to describe the setting with slight nuance. 
Specifically, ``unsupervised'' often refers to the network training without human supervision, ``weakly-supervised'' describes the case where the supervisions may be noisy, limited, or imprecise; ``self-supervised'' is used to describe the model trains itself to learn one part of the input from another part of the input.
In this section, we refer to them as ``non-supervised'' only for easy structure and presentation purposes.

The fundamental idea for non-supervised learning setting relies on the premise of the intrinsic synchronization nature among paired data from multiple modalities~\cite{zhu2021learning,yang2017deep,radford2021clip}. 
For example, certain video actions are naturally accompanied with characteristic sounds as described in the ambient sound section in Section~\ref{subsec:audio}.
The images and captions are also paired to train the vision and language models.

Nowadays, there are several large-scale pre-trained models in multimodal learning research area, especially in the text-image domain, that have been attracting much attention, due to their impressive performance as well as the wide downstream applications~\cite{ramesh2021zero,luo2020univl,ni2021m3p}.
We can consider the large-scale pre-training as a specific type of multimodal representation learning, since the primary goal of pre-training is to learn a joint and unified cross-modal representation that can be flexibly transferred to other domain or downstream tasks.

There are in general two \re{popular} methods for the pre-training field, which are contrastive learning based~\cite{radford2021clip,akbari2021vatt} and mask reconstruction based~\cite{chen2020uniter,lu2019vilbert,lei2021less}. 
One of the most popular examples of such models include CLIP (Contrastive Language-Image Pre-Training)~\cite{radford2021clip} for the vision and language pre-training.
Most of those models are developed following the BERT~\cite{devlin2018bert} and GPT (Generative Pre-trained Transformer) models for natural language~\cite{brown2020language} and images~\cite{chen2020generative}, whose the core design consists of transformer architectures~\cite{vaswani2017attention} pre-trained for text and image generation tasks.
Inspired by the success of GPT models in showing the potential to use the language or image to guide a large neural network to accomplish a variety of generation tasks in their respective domains, researchers naturally proceed to the multimodal area to bridge these modalities.
The CLIP model is trained on 400 millions text-image pairs, and is considered as one of the first large-scale pre-trained models for the multimodal learning area to bridge the text and image data space. 
Another example is VATT (Video-Audio-Text Transformer)~\cite{akbari2021vatt}, which is a transformer-based self-supervised large-scale model for learning representations from raw video, audio and text. It first processes raw data from different modalities via linear projection, and trains the model to learn a semantically latent space via the Noise Contrastive Estimation (NCE).
One commonness among those pre-training works is that the proposed models are trained with enormous amount of data using extensive computational resources. 
From the technical point of view, CLIP follows the general idea to align the embedding space of paired images and the corresponding textual descriptions. It adopts the batch construction technique~\cite{sohn2016improved} by encoding the entire sentence description as an entirety, instead of processing the textual word by word. CLIP jointly trains a text and an image encoder by optimizing the similarity scores of a given pair. During the inference time, the model can be used for zero-shot prediction by embedding the names or the descriptions of the target dataset's classes in textual form via the learned text encoder. 



It is worth noting that while those large-scale models~\cite{radford2019gpt3,radford2021clip,akbari2021vatt} are able to achieve very impressive results, there are few radical novelties in the model architecture and training techniques. Therefore, while they have received large attention, there are also controversies regarding those works. One of the discussions over such large-scale pre-trained models is that the impressive results are largely due to the diverse and enormous data that have been carefully designed to train the model, as well as their scale on the existing models on large. Other concerns over the privacy and ethic issues have also been raised against these works. 
Overall speaking, despite the controversial voices over the topic, those models do help with building a more unified toolkit in connecting the visual and textual space in the multimodal learning area, which also promotes large number of following works that are developed based on the aligned feature space for various downstream tasks.

\vspace{-0.1in}
\subsection{Trend in Representation Learning}
\minr{The machine learning and computer vision research community is rapidly advancing, with a trend towards scaling up data representation learning using emerging foundation models, empowered by the upgrade on the dataset scale and computational resources. The multimodal representations learned by large pre-trained models such as CLIP~\cite{radford2021clip} have been successfully applied in various multimodal downstream tasks, boosting the performance, especially in the axes of generalization ability of models.}

\minr{However, we also want to emphasize that scaling up is not a panacea. Despite the benefits, fundamental issues persist, such as out-of-distribution challenges and amplified model bias~\cite{wang2023overwriting}. While large pre-trained models are powerful for many multimodal tasks, future research needs to focus on real-world scenarios with more edge cases and complex data formats for safe and responsible deployment.
}

\vspace{-0.15in}
\section{Discriminative Applications}
\label{sec:discriminative}

In this section, we \ye{discuss} the multimodal learning works for discriminative task applications, with subsections categorized with specific data modality combination in form of ``Vision+X", where X stands for \ye{the additional} data modalities.


For discriminative applications, \re{popular} approaches usually inherit the neural networks from general representation learning in Sec.~\ref{subsec:rep_learning_main}, with additional modules to adapt to task-specific objectives. 
A general methodological design in multimodal learning follows the ideas of \textit{``separate processing"} and \textit{``unified fusion"}. To be more specific, data of different modalities are first processed with respective network branches, and then the inter-modality learning is further performed by extra mutual modules before outputting the final results for different tasks.
Since the exact objectives depend on the task scenarios, we leave the detailed introduction in the following subsections.
In terms of the evaluations, different multimodal tasks have their corresponding evaluation protocols. Similarly to the specific methodology design, we detail the evaluations in the subsections below.

\vspace{-0.12in}
\subsection{Vision+Audio}

\noindent \textbf{\re{Audio-Visual Event Localization (AVEL).}}
\re{
The Audio-Visual Event (AVE) is defined as an event that is both audible and visible in a video segment~\cite{tian2018audio}, and the AVEL task aims to localize the AVE within an unconstrained video~\cite{tian2018audio,wu2019dual,xuan2020cross,duan2021audio,lin2020audiovisual}. 
This task was first proposed in~\cite{tian2018audio}, together with the AVE video dataset (details in Table~\ref{tab:data} and Appendix~\ref{sec:dataset}). 
The overall task objective resembles the action recognition with ambient audio data and the requirement for temporal localization under supervised or weakly-supervised settings.
To tackle the additional ambient sound accompanying an event, a common method is to achieve the cross-modal interactions via different attention modules~\cite{tian2018audio,wu2019dual,duan2021audio,yu2021mm}.  
Many of the existing works follow the framework that processes audio and visual data with separate encoders, and fuses the processed information for temporal localization and activity classification. 
The temporal connection from the video stream is often addressed via the model backbone such as LSTM~\cite{hochreiter1997long}. 
Evaluation of the AVEL task usually utilizes the prediction accuracy metric.}

\noindent \textbf{Audio-Visual Video Parsing (AVVP).}
The AVVP problem aims to parse a video into temporal segments and label them as either audible, visible or both~\cite{tian2020avvp,wu2021explore,lin2021exploring}.
\re{It is initially developed from the AVEL task, with its task emphasis on the recognition, while the AVEL focuses more the temporal localization.}
As a variant of the AVEL task, several works have been developed with the common core idea that seeks to learn an effective audio-visual feature as the foundation, and then to incorporate further refined technical designs to address specific task requirements.
For instance, Lin \textit{et al.}~\cite{lin2021exploring} introduced a sequence-to-sequence manner integration of audio and visual features. Yu \textit{et al.}~\cite{yu2021mm} explore the AVVP task by taking the potential audio-visual asynchrony into account.

\noindent \textbf{Visual Sound Source Localization (VSSL).}
Visual Sound Source Localization (VSSL) task aims to locate the corresponding visual locations within an image given the sound~\cite{rachavarapu2021localize,oya2020we,song2022self,qian2020multiple,senocak2022less,senocak2018learning}.
While the original sound source localization task (SSL) has been widely studied in the signal processing field~\cite{grumiaux2022survey}, the deep learning based visual localization was first proposed in~\cite{senocak2018learning}.
The high-level idea also focuses on learning the correlations between paired audio and visual data, except for the visual part, the VSSL task tends to switch the regions of interest within the visual data given different ambient audio signals.
Overall pipeline often consists of separate encoders for visual and audio input, and then fuses the audio-visual information for learning a localization module during training.
More technical details may differ in terms of how to perform the fusion via attention mechanisms~\cite{senocak2018learning,qian2020multiple}, the training technique using various localization or contrastive loss~\cite{senocak2018learning,rachavarapu2021localize,qian2020multiple}.
To assess the performance of the VSSL task, metrics such as the cIoU (Complete IOU) and AUC (Area under the ROC Curve) scores are often used to quantify the precision of the predicted areas for sound sources.



\vspace{-0.12in}
\subsection{Vision+Text}

\noindent \textbf{Visual Grounding.}
As a popular discriminative vision and language task, visual grounding aims to localize the object within an image given a text description as input~\cite{fukui2016multimodal,yang2020improving,liu2020learning,liu2019learning,yu2018rethinking,sigurdsson2020visual,hong2019learning,xiao2017weakly,deng2021transvg,yang2019fast,huang2022deconfounded,li2020visual,deng2018visual}. 
The idea of realizing cross-reference between sentences and visual context was first proposed and studied in~\cite{karpathy2014deep}, where the task was also known as \textit{``referring expression comprehension"}.
Pioneering works on the referring usually only require the grounding of a single object from the description sentence input~\cite{hu2016natural,yu2016modeling,mao2016generation}, working on the premise that the region of interest is expected to achieve the maximum posterior probability for the given textual description. 
More recent works tackle a more challenging visual grounding setting, where the task is refined into two sub-objectives: phrasing and grounding.
For the first sub-objective, models are expected to localize all the objects mentioned in the given textual description, and then to individually detect their corresponding boxes in the image~\cite{plummer2017phrase,liu2021relation}. 

In terms of \re{methodology designs} for the visual grounding task, most works~\cite{fukui2016multimodal,yang2020improving,liu2020learning,liu2019learning,yu2018rethinking,sigurdsson2020visual,hong2019learning,xiao2017weakly,deng2021transvg,yang2019fast,huang2022deconfounded,li2020visual,deng2018visual,hu2016natural,yu2016modeling,mao2016generation} can be categorized into supervised, weakly-supervised and unsupervised settings. 
The supervised setting refers to the condition where the annotation of phrase-object pairs is provided, the weakly supervised removes the phrase annotations for the textual description input, and the unsupervised setting completely removes annotations for both data modalities. 
As for general pipelines, most methods follow either a two-stage or one-stage framework. 
For the two-stage frameworks, models first extract region proposals for potential objects within the image, and then rank and match the proposals with language phrases. 
For the one-stage frameworks, visual objects and textual phrases are aligned and connected during the learning process to avoid redundant region proposals as in two-stage designs.
In the case of weakly- or unsupervised settings, some extra regularization losses such as structural loss and discriminative loss~\cite{xiao2017weakly,sun2021discriminative} are usually needed to better learn the correlations between corresponding object regions and textual phrases.   
The evaluation for the visual grounding resembles other visual localization tasks, which often use the IoU (intersection over union) between the predicted and ground truth boxes as a quantitative measurement, with a threshold value of 0.5. Another unique metric for the visual grounding task is \textit{PointIt} (pointing game metric)~\cite{xiao2017weakly}, which computes the pixel location with maximum predicted attention weight, if the selected hit point lies within the ground truth box area, the prediction is counted as valid.


\noindent \textbf{Temporal Activity Localization (TAL).}
Activity localization task (TAL) is also known as the video grounding, which seeks to locate the temporal segment of a video clip given the language description of a certain activity as query~\cite{gao2017tall,chen2019semantic,zhang2021multi,zhang2021multi,wang2019language,chen2020hierarchical,anne2017localizing}. 
Compared to the visual grounding within the image, TAL requires additional reasoning and matching along the temporal direction as indicated by its name. 
For this task, the models are expected not only to capture the correlations between the visual activity and the language, but also able to temporally localize the segment among consecutive video frames. 
While the high-level framework structure remains similar to previous multimodal discriminative tasks with separate encoders, a multimodal processing module for fusing the features, a decoder module adapted for specific task objectives, and early representative works for TAL task introduce different techniques to emphasize the network ability for temporal reasoning.
Gao \textit{et al.}~\cite{gao2017tall} propose a Cross-modal Temporal Regression Localizer (CTRL) with a Temporal localization regression network to align the fused visual-textual information with video temporal locations. 
The popular quantitative metrics used for evaluation include the \textit{mean IoU} and \textit{IoU@a}, with \textit{a} standing for the percentage of overlap between the predicated segment and the ground truth annotations.

\noindent
\textbf{Visual Entailment (VE).}
Visual entailment (VE) seeks to predict the logical relationship of a piece of text to an image~\cite{xie2019visual,xie2018visual,thomas2022fine}. It is developed from the textual entailment task~\cite{dagan2005pascal}, whose initial objective is to decide if a hypothesis can be logically deduced from the premise.
Xie \textit{et al.}~\cite{xie2018visual} extends the textual entailment to the multimodal context, which replaces the textual premise with an image.
Thomas \textit{et al.}~\cite{thomas2022fine} further refines the task by introducing different levels of granularity.
The emphasis of the VE task lies within the multimodal reasoning ability of networks. 
To achieve the reasoning between the image and textual hypothesis, early methods~\cite{xie2018visual,xie2019visual} adopt separate network branches to process visual and textual data and leverage the attention interaction for interactions. A refined framework~\cite{thomas2022fine} further disentangles the textual hypothesis into its constituent parts and proposes to enhance the reasoning by introducing an abstract meaning representation (AMR) graph for the decomposed textual components. 
The performance is often evaluated via prediction accuracy given the premise as input.

\noindent
\textbf{Spatio-Temporal Video Grounding (STVG).}
Spatio-Temporal Video Grounding (STVG) is a recent multimodal task that lies at the intersection of visual grounding and temporal localization, integrating reasoning among space, time, and language within the visual context of videos~\cite{su2021stvgbert,yang2022tubedetr,zhang2020does,jin2022embracing}.
Specifically, given an untrimmed video and a textual description of an object, the task seeks to localize a spatio-temporal tube (\textit{i.e.}, a sequence of bounding boxes) for the target object described.
Most existing methods for STVG either build upon the idea from visual grounding or focus on temporal localization designs.
One popular paradigm for tackling the task adopts the two-stage design, which leverages pre-extracted object proposals and then integrates the temporal localization via attention mechanisms~\cite{yang2019dynamic,wang2019neighbourhood}.
In the meantime, another thread of works proposes one-stage frameworks and does not rely on prior for object proposals~\cite{kamath2021mdetr,yang2022tubedetr}.
In terms of network architectures, Transformers are widely adopted as the backbone for such methodology designs~\cite{su2021stvgbert,yang2022tubedetr,zhang2020does}.
STVG is usually evaluated via IoU metrics by comparing the frame overlap between the ground truth and the predicted timestamps.

\vspace{-0.12in}
\subsection{Vision+Audio+Text}



\noindent \textbf{Multimodal Retrieval.}
Another multimodal discriminative task that has been widely studied is the retrieval~\cite{chun2021probabilistic,gu2018look,zhen2019deep,wang2016comprehensive,wang2017adversarial,wang2022multi}. 
Most retrieval works operate on the representation space by measuring the similarities among learned representations from different modalities.
As consequences, retrieval task is also one of the most frequently used downstream tasks in representation learning works.

While the retrieval task can be conducted within a single modality of data, multimodal retrieval seeks to extend its original setting to cross-modality scenarios, where we want to retrieve items that match the input from a different data modality, \textit{e.g.,} text based vision retrieval, audio based vision retrieval.
For example, CAMP~\cite{wang2019camp} learns the text and image embedding via message passing across modalities.
Gu \textit{et al.}~\cite{gu2018look} propose to improve the text-visual retrieval with auxiliary generative models.
~\cite{wang2021t2vlad} looks into the task of text based video retrieval by additionally looking into the local details via a global-local alignment method in the learned representation space.
~\cite{zhu2021learning} learns a mutual audio-visual latent space via VAE-based framework for audio-visual cross-modality retrieval.
Oncescu \textit{et al.}~\cite{oncescu2021audio} propose to retrieve audio signals given natural language queries.

\noindent
\textbf{Audio-Visual Question Answering.}
\re{Audio-Visual Question Answering, as indicated by its name, is an extension based on visual question answering with integrated audio modality~\cite{li2023progressive,li2022learning,yun2021pano,yang2022avqa}.
Specifically, AVQA often involves questions regarding different visual objects, sounds, and their associations in videos.
Existing methodology designs are often extended from VQA frameworks with extra interactions with audio data.
For instance, an intuitive framework~\cite{li2022learning} expands the two-branch encoder design into three-branch and separately processes video, audio, and textual data before introducing interactions via the attention mechanism.
Answer prediction accuracy is often used for evaluation.
}

\vspace{-0.15in}

\section{Generative Applications}
\label{sec:generative}

In this section, our focus is on cross-modality synthesis tasks for generative applications. These tasks involve generating data from a specific modality or multiple modalities as input.

There are usually typically two high-level approaches to generating data in cross-modality synthesis tasks: retrieving an item from a given database, or directly synthesizing and decoding data via the neural networks.
For the retrieval-based generations, the core idea follows the logic to search for an item or several items that mostly resemble the ``generated'' data. A large percentage of the retrieval-based works perform the similarity measurement on the data representation level without actually considering the decoding part. Technically, we argue that such works are categorized in the \textit{Representation Learning} section.
As consequence, we mainly focus on introducing the works that ``truly generate'' data instead of retrieving items in this section for generative applications.

\vspace{-0.12in}
\subsection{Generative Networks}
\label{subsec:generative_backbone}

\re{Before diving into the concrete application tasks, we first introduce three popular backbone models for general generative tasks that have been widely adopted in multimodal generative literature.}

\noindent \textbf{VAE-Based Models.}
Variational Auto Encoders (VAEs)~\cite{vae-kingma2013auto} are classic generative models proposed based on the deep neural autoencoders~\cite{hinton2006reducing} under the unsupervised learning setting. 
The core of autoencoders relies on the premise that an effectively trained encoder should learn the data representation in a way that the encoded representations can be decoded to reconstruct the original data input by a decoder.
Compared to conventional autoencoders, VAEs introduce regularizations on the bettleneck level by re-parametering the latent space using the Gaussian priors, where the learned Gaussian parameters allow for sampling new data.
Typical training of VAEs often includes two types of losses, which are the variational loss (ELBO)~\cite{vae-kingma2013auto} that consists of a regularization loss on the latent representation space (\textit{e.g.}, Kullback–Leibler divergence) and the reconstruction losses on the output data (\textit{e.g.}, Mean Square Errors (MSE)). 
\re{The classic variational objective can be formulated and derived from the following equation:
\begin{equation}
    \mathrm{log} \: p(x) = \mathbb{E}_{q(x|z)}[\mathrm{log}\:p(x|z)] - D_{KL}[q(z|x)||q(z)],
\end{equation}
where $p$ represents the decoder, $q$ is the encoder, $x$ and $z$ denote the original raw data and the learned latent embedding, respectively.
In the actual implementation, the re-parameterization technique, which assumes $z \sim \mathcal{N}(\mu, \sigma)$, allows us to minimize the KL divergence based on sampling from N samples:
\begin{equation}
    \sum^n_{i=1}\sigma^2_i+\mu^2_i-\mathrm{log}(\sigma_i)-1.
\end{equation}}
VAEs have been widely exploited in various generative tasks in audio and images~\cite{vae-kingma2013auto,higgins2016beta} as well as in the multimodal context for cross-modality generations~\cite{spurr2018cross,zhu2021learning}.


\noindent \textbf{GAN-Based Models.}
The generative adversarial networks (GANs)~\cite{gan} are another mainstream backbone type for various generative models. From a high-level perspective, GANs involve two agents (\textit{i.e.}, the generator $\mathcal{G}$ and the discriminator $\mathcal{D}$) playing an adversarial game. The generator aims to synthesize realistic data that resemble real data for fooling the discriminator, while the goal of the discriminator is to distinguish between the synthesized data by $\mathcal{G}$ from the real data.
Similar to VAEs, the training of GAN-based models does not require external annotations but only the real raw data, and therefore is frequently used in unsupervised or weakly-supervised settings. 
The standard training of GANs also minimizes losses from two aspects with latent space regularization (also known as adversarial losses) and reconstruction optimizations~\cite{gan,salimans2016improved}.
Following the original work, multiple variants of the GAN models and adversarial losses have been proposed such as the Wasserstein GANs with the Wasserstein loss~\cite{arjovsky2017wasserstein,gulrajani2017improved} and conditional GANs~\cite{mirza2014conditional}.
The classic GAN loss is formulated as follows:
\begin{equation}
\begin{split}
    & \mathrm{min}_{G}\mathrm{max}_{D}\:V(D,G) = \\
    & \mathbb{E}_{x\sim p_{data}(x)}[\mathrm{log}D(x)] + \mathbb{E}_{z \sim p_Z(z)}[\mathrm{log}(1-D(G(z)))],
\end{split}
\end{equation}
where $G$ and $D$ are the generator and discriminator, respectively. $x$ and $z$ denotes the original raw data and the learned latent embedding.
On the application level, GANs are first widely applied in image generations~\cite{isola2017image,brock2018large}, later studies also explore the GAN-based models for audio synthesis~\cite{melgan,hifigan} or cross-modality areas~\cite{yezhu2022quantizedgan,zhu2019dm,zhan2019spatial}.


\noindent \textbf{DPM-Based Models.}
Compare to VAEs and GANs, the diffusion probabilistic models (DPMs)~\cite{sohl2015dpm_thermo} are another type of generative backbones that have been very popular in more recent years. 
In principal, the DPMs include a Markov chain of a finite steps in two opposite directions. The forward direction, also known as ``diffusion" process seeks to gradually add noises to a given data at each diffusion step, while the inverse denoising process aims to remove the noises added in the forward steps and recover the actual data from a non-informative noisy distribution. 
There are two variants of conventional DPMs that differ in the state space formulations of the Markov chain.
Classic DPMs suppose the state space to be continuous, and parameterize the diffusion process with Gaussian noises~\cite{ho2020dpm,nichol2021improveddmp,dhariwal2021diff-img1,kong2020diffwave,theo-diff1,theo-diff2,theo-diff3,theo-diff4}, while another variant of DPMs consider the discrete state space and formulate the diffusion process with state transition matrix~\cite{austin2021structured,gu2021vector,yezhu2022cdcd}.
The variational lower bound~\cite{ho2020dpm} is the classic loss function used for effective DPMs learning, other practical losses such as the auxiliary loss~\cite{austin2021structured,gu2021vector}, classifier-free guidance~\cite{ho2022classifier} and contrastive diffusion losses~\cite{yezhu2022cdcd} are also proposed to further improve the generation performance. 
Vanilla DPMs are trained on a variational lower bound defined as follows:
\begin{equation}
     \begin{split}
     & \mathcal{L}_\mathrm{vb} = \mathbb{E}_q[\underbrace{D_\mathrm{KL}(q(x_T|x_0)||p(x_T))}_{\mathcal{L}_T} + \\
     & \sum_{t>1}\underbrace{D_\mathrm{KL}(q(x_{t-1}|x_t,x_0)||p_{\theta}(x_{t-1}|x_t))}_{\mathcal{L}_{t-1}}-\underbrace{\mathrm{log}\:p_{\theta}(x_0|x_1)}_{\mathcal{L}_0}],
     \end{split}
    \label{eq:vlb}
\end{equation}
where $q$ and $p$ represent the diffusion and denoising processes, respectively. $x_{i}$ denotes the data at diffusion step $t$. 
DPMs have received much research attention due to the competitive performance in generative tasks for images~\cite{ho2020dpm,nichol2021improveddmp,dhariwal2021diff-img1,ho2022diff-img2,hu2021global}, audio~\cite{kong2020diffwave,mittal2021symbolic-diff,lee2021nu}, as well as in the cross-modality scenarios such as text-to-image~\cite{nichol2021glide,gu2021vector,yezhu2022cdcd} and dance-to-music generations~\cite{yezhu2022cdcd}.

\ye{The evaluations for the generative tasks have always been an important aspect to consider. Typically, the assessment of synthesized data in multimodal setting considers both unimodal and multimodal criteria.
The unimodal metrics are utilized not only for multimodal scenarios but also for generative tasks in general, such as the fidelity in image generation. 
In addition to the general quality, the multimodal generations also take the cross-modality correspondence into account, such as the beat correspondence between video and music.
We summarize the common evaluation metrics for various synthesized data in generative tasks in Appendix~\ref{app_sec:evaluation_generation}.
}

\vspace{-0.13in}
\subsection{Vision+Audio}

\noindent \textbf{Music Generation from Vision.}
Recent studies that seek to generate music from visual data (usually from videos) can be categorized by their adopted music representations.
One branch of the music generation works~\cite{foley,aggarwal2021dance2music,di2021video,su2020audeo} rely on the symbolic audio representations such as 1D piano-roll and 2D midi and as we have introduced in Section~\ref{subsec:audio}.
The symbolic musical representations can be decoded back to the raw audio waveform by pre-defined synthesizers that introduces no additional noises, thus enduring the high-quality of the generated music.
This is especially true when compared to the learning-based musical representations and decoders~\cite{jukebox,yezhu2022quantizedgan,yezhu2022cdcd}, where the synthesized music usually has relatively high noise levels.
Secondly, the computational cost of symbolic representation based works is generally lower than the pure learning-based methods, since the symbolic musical representations are very sparse and low dimensional, which facilitate the learning and inference process.
However, such symbolic based music generation methods are also restricted in terms of the music diversity and flexibility. 
Especially in current research works, the generated music are usually limited to a certain specific pre-defined instrumental sound~\cite{di2021video,foley,aggarwal2021dance2music}. 
It is worth noting that most symbolic based music generation works, despite the fact that the framework output is raw music, do not directly use the generative backbones as we have introduced in Sec.~\ref{subsec:generative_backbone}. Technically, most of them are trained based on the ground truth MIDI annotations in form of cross-entropy loss.  
In contrast, another branch of research works deploy the learning-based musical representations either in form of continuous or discrete.
However, despite the continuous music representations have been exploited in the music synthesis field~\cite{melgan}, most recent cross-modal music generations adopt the discrete form of learned musical feature - \re{vector quantization} (VQ) - as the intermediate representations~\cite{yezhu2022quantizedgan,yezhu2022cdcd}, leveraging the large-scale pretrained music synthesis model JukeBox~\cite{jukebox}. 
For example, D2M-GAN~\cite{yezhu2022quantizedgan} proposes a GAN-based framework that takes the human body motion data and dance video frames as input, and generates musical VQ representation.
CDCD~\cite{yezhu2022cdcd} builds upon the diffusion probabilistic model with a discrete state space represented by the VQ, and incorporates a contrastive diffusion loss to train the network to improve the input-output correspondence for cross-modality applications.

\noindent \textbf{Speech Generation from Videos.}
In addition to the music audio, another specific generation task seeks to synthesize speech audio from videos of human speaking~\cite{ephrat2017vid2speech,kim2021lip,salik2019lipper,ephrat2017improved,vougioukas2019video,yadav2021speech,michelsanti2020vocoder,Prajwal_2020_CVPR,mira2022svts}.
One unique aspect about this audio generation task is that the speech largely relies on the movement of lips while speaking.
Based on this characteristic, many works on this direction focus on reading and interpreting visual lip movement from video input and then convert it to audio waveform, which also explains the reason why this ``video-to-speech" synthesis task is also known as ``lip-to-speech" generation.
Therefore, despite the topic of audio generation from videos, large percentage of the works in this area rather focus on the ``motions in the videos", instead of the raw videos.
To enhance the correlations between lip movement and speech audio, the audio-visual cross-modal attention mechanisms are further adopted to improve the generation quality.
Kim \textit{et al.}~\cite{kim2021lip} propose an attentional GAN with visual context to read lips for speech synthesis.
Yadav \textit{et al.}~\cite{yadav2021speech} use the VAE generative backbone with the stochastic modelling approach.
At the same time, more refined variant of this problem with disentangled speech features such as the individual speaking styles have also been studied~\cite{Prajwal_2020_CVPR}.

\noindent \textbf{Ambient Sound Generation from Videos.}
Research works that seek to generate sound from natural videos~\cite{chen2020generating,zhou2018visual,chen2018visually} put special emphasis on the alignment between the generated sound and the visual context, which consists of both the semantic and temporal alignments.
Chen \textit{et al.}~\cite{chen2018visually} address the semantic alignment problem by adopting a perceptual loss and considering the sound categories in the optimization process.
Zhou \textit{et al.}~\cite{zhou2018visual} follows a rather classic encoder-decoder framework for video input and audio decoder, propose three methods including the frame-to-frame, sequence-to-sequence, and flow-based variants.
In~\cite{chen2020generating}, the authors tackle both semantic and temporal alignment with the proposed \textit{REGNET} framework, whose core technical design includes a visual encoder and an audio forwarding regularizer. 
Generally speaking, compared to speech and music, the ambient sound has relatively fewer unique attributes other than its correspondence with certain activity. 
Summarizing the above works, we notice the high-level technical ideas are rather general and resemble to the standard pipeline designs.

\noindent \textbf{Visual Generation from Sound.}
As the inverse direction of the sound generation from vision, to directly generate the pixel-wise natural images or videos solely from the audio modality is a challenging problem. 
However, as a specific type of visual generation from sound tasks, synthesizing talking faces from speech audio~\cite{zhou2021pose,song2018talking,chen2019hierarchical,zhang2021flow} is a relatively well-studied sub-field. 
Similar to the \textit{``video-to-speech"} task, the visual information in this reverse direction rather emphasizes the motions of lips in a video clip.
In most cases, the input for the talking faces generation task includes a reference image and a driving audio track. 
Early works~\cite{song2018talking,vougioukas2020realistic,chen2018lip} adopt a general pipeline with two separate encoders for the input and a decoder for synthesizing the talking videos mostly via the GANs-based generative backbones. 
More recent works seeks to refine and improve the synthesis results by splitting the previous architecture into hierarchical structures~\cite{chen2019hierarchical,das2020speech}.  
In addition to the raw videos, more specific motion data such as the flow is also used to further enable high-resolution generations~\cite{zhang2021flow}.
Some works also seek to generate natural videos by reformulating the video generation problem into an motion generation task in the form of optical flow~\cite{chatterjee2020sound2sight,denton2018stochastic}.

\vspace{-0.12in}
\subsection{Vision+Text}

\noindent \textbf{Caption Generation from Vision.}
One of the classic vision and language generative tasks is the image and video captioning~\cite{you2016image,wang2017diverse,xu2015show,chen2017sca,lu2017knowing,wang2018reconstruction,chen2019variational,wu2022snoc}, which aims to generate a language textual description of the given visual data.
In the early stage of deep learning based visual captioning~\cite{mao2014deep,vinyals2015show}, the general pipeline usually consists of encoder-decoder framework with the CNNs and RNNs as the backbone architectures for the image encoder and text decoder. 
Later on, the attention mechanism becomes a popular technique~\cite{xu2015show,lu2017knowing,wu2018decoupled} to enhance the correlations between the sentence descriptions and the corresponding visual concept. A large amount of research works follows the updated general encoder-decoder pipeline with an additional attention modules. 
Other than the general encoder-decoder framework, there are some other works that tackle the captioning task using techniques such as adversarial learning~\cite{dai2017towards}, or reinforcement learning~\cite{liu2017improved}.

In addition to the technical progress, there are also some works that bring novel insights on the task setting of caption generations. For example, ~\cite{hori2019end,schwartz2019simple,alamri2019audio} incorporates ambient audio from the video as an additional input data modalities to further assist the video captioning. Following the above works,~\cite{zhu2020describing,zhu2021saying} propose a new setting where part of the visual data becomes inaccessible as input, they further propose a dialog process between two agents as a supplement to the missing visual input, whose final goal remains to be generating a precise and complete textual description for the video.

\noindent \textbf{Dialog Generation from Vision.}
In addition to the captioning tasks, another form of text generation from visual input focus on the dialog texts instead of plain descriptions.
We can further classify the dialog generation from vision into visual question answering (VQA)~\cite{das2017human,xiao2021next,antol2015vqa,xu2016ask,Agarwal_2020_CVPR,chen2020counterfactual,shih2016look} and visual dialog ~\cite{das2017visual,de2017guesswhat,jain2018two,seo2017visual,wu2018you,zhu2020describing,zhu2021saying,qi2020two}.
As described in Sec.~\ref{sec:data}, the main difference between the two subcategories lies within the fact that the former VQA task aims to answer a single question related to the visual input, while the latter expect to maintain the question-answer interactions for multiple rounds with internal logic.

VQA task was first introduced in~\cite{antol2015vqa}, whose task objective is to answer a language question related to the visual content. Similarly to the captioning task, the mainstream frameworks also follow the encoder-decoder pipeline, usually equipped with separate encoders for the visual and textual data, and a decoder for generating the words in language. Attention mechanisms have also been widely used in literature for the similar purpose to enhance the correlations between corresponding features from visual and language domains. 
One uniqueness of the VQA task compared to the previous visual captioning is the potential bias problem in the task setting. Specifically, one (models) may not truly rely on the visual context to answer the raised questions. For example, given a question such as `` what is the color of the sky ?", the answer is likely to be ``blue" in most cases, or given a question that the answer is expected to be ``yes" or ``no", one (models) can simply guess between two options. Therefore, to address such spurious pattern and bias problem, recent works have been seeking to migrate the issue by analyzing the causal relations~\cite{Niu_2021_CVPR,chen2020counterfactual}.

As another popular vision and language task, visual dialog follows a similar development path after being proposed in ~\cite{das2017visual}, where the general encoder-decoder framework is further extended to include an additional encoder for the dialog history data. 
From a high-level point of view, the challenge of visual dialog compared to the captioning and VQA is that the multiple rounds of question-answer interactions may refer to different parts of visual context with an image or video as the dialog goes on, leading to a higher requirement for precisely referencing the key information between two visual and textual data. In the meanwhile, the bias issue for VQA remains in the visual dialog.

\noindent \textbf{Image Synthesis from Text.}
The inverse direction that seeks to generate image data from given text conditioning has also been a very active research topic in multimodal learning field within recent years since 2016~\cite{li2019object,reed2016generative,zhang2017stackgan,zhang2018stackgan++,xu2018attngan,gu2021vector,yezhu2022cdcd,liu2022compositional}.
It is a more challenging task compared to the text generation from visions, due to the reason that the visual data usually are richer in context, with demanding requirement for pixel-level synthesis.

Following the preliminary exploitation in the text-to-image task from~\cite{reed2016generative}, the recent literature in the area can be divided into different stages following the chronological order.
Before the year of 2020, the mainstream approaches use the GAN based architectures to tackle the problem~\cite{yang2016stacked,zhang2018stackgan++,xu2018attngan}.
Later with the development of multimodal learning, researchers start to get inspiration from the NLP society by adapting techniques originated from language processing, one concrete example is the Auto-Regressive models for image generation~\cite{ramesh2021zero,ding2021cogview}.
Starting from recent several years since 2020, DPMs have gradually become one of the most active methods for vision generation tasks~\cite{theo-diff1,ho2022diff-img2,ho2020dpm,yezhu2022cdcd,gu2021vector,ramesh2022dalle2}, due to its impressive performance and better tractability compared to GANs.

On the other hand, it is worth mentioning that several large-scale models have been recently introduced in the text-to-image field such as \re{DALL$\cdot$E}~\cite{ramesh2021zero} and \re{DALL$\cdot$E 2}~\cite{ramesh2022dalle2}.
\re{DALL$\cdot$E} model includes two separate stages to train for the raw image generation objective. In the first stage, \re{DALL$\cdot$E} learns the visual concept from images via the discrete VAE~\cite{ramesh2021zero}, and then fuse the learned discrete image embedding with the textual tokens to train the transformer~\cite{vaswani2017attention} in the second stage. In inference, \re{DALL$\cdot$E} first obtains the fused text-image embedding for the given text description and potential image candidates, and then use the pre-trained CLIP to do image re-ranking to get the generated images with higher similarities.  
\re{DALL$\cdot$E 2}, as an improved text-to-image generator, uses the Diffusion Probabilistic Models (DPMs) (see details in Section~\ref{subsec:generative_backbone}) on the image embedding space from CLIP to do the image synthesis.

\noindent \ye{\textbf{Text-Guided Image Editing.}
As a step further from the image synthesis from text, another popular generative task that combines the vision and text is to perform editing based on text prompt for the given raw images~\cite{avrahami2022blended,nichol2021glide,aghajanyan2022cm3,kim2022diffusionclip,zhu2023boundary,kwon2022diffusion}.
Compared to the text to image generation, the setup of text-guided image editing using the generative models takes not only the text prompt, but also raw real images as part of the input. The task objectives are usually two-fold: achieve the target editing effect, and preserve the remaining features of the given images.
}

\ye{Similar to the text-to-image (T2I) literature, as a refined task with additional raw image data as input, the community has explored several mainstream generative approaches using different architectures such as GANs~\cite{zhu2020domain,pan2023drag} and DPMs~\cite{kim2022diffusionclip,zhu2023boundary,avrahami2022blended,nichol2021glide}.
As the state-of-the-art approaches, one way to categorize the DPM-based image editing methods is to see whether the proposed methods require additional learning given a pre-trained generative model.
In most intuitive and straightforward cases, image editing via text prompt requires fine-tuning the parameters of given pre-trained models~\cite{kim2022diffusionclip}, or learning extra neural network modules~\cite{kwon2022diffusion} to achieve the target editing effect.
Another category of image editing methods propose to solve the editing objective in a leaning-free manner~\cite{zhu2023boundary}, by explicitly leveraging the intrinsic abilities of DPMs in exhibiting semantics along the generation trajectories.
}

\noindent \textbf{Video Synthesis from Text.}
There are also works tacking the more challenging vision generation from text that focus on the videos~\cite{li2018video,ho2022video,hu2022make}. Compared to the images, videos often consist of multiple consecutive frames that are temporally and spatially correlated, in addition to the pixel-level computational extensive synthesis for a single visual frame.
Existing works usually deploy VAE or GAN-based generative backbones to tackle the task~\cite{li2018video} with language priors. 
Alternatively, Hu \textit{et al.}~\cite{hu2022make} modifies the task formulation by providing a reference image and a textual description, and synthesize videos based on the given input.
More recent work~\cite{ho2022video} introduces a video diffusion model for the task via modified 3D U-Net architecture equipped with an extra temporal dimension. 
\textit{Make-A-Video} is one of the first large-scale video generation models. 
It leverages the recent advances in text-to-image field to learn the visual information, and then proposes to learn the temporal motions from unlabelled large-scale video data. In inference, the model composes the visual content with learned motions to synthesize a realistic video.

\vspace{-0.15in}
\section{Further Discussion}
\label{sec:future}

\subsection{Insights from Data and Methodology Design}
\label{subsec:revisit}

Our paper aims to provide a novel perspective to understand the multimodal learning in the light of data, we revisit and discuss the correlation between the data nature and the methodology design as part of discussion from two aspects: the semantics of data modalities, and their specific formats.

For data semantics, as explained in Sec.~\ref{sec:data}, visual data and several types of audio data, such as the ambient sound, can be considered as raw information sources. These modalities include sensory information directly captured from the environment that is usually high-dimensional and can be further processed and analyzed with information redundancy.
In contrast to raw information sources, text data and certain types of audio signals, such as speech, have undergone extensive processing throughout the evolution of human civilization. These data modalities already possess meaningful semantics and are more information compact with a unified representation in the form of tokens. Additionally, most NLP tasks have a rather unified problem formation under the notion of "next word token prediction". This distinction in data nature, particularly in terms of semantics, has played a significant role in shaping the methodological and technical advancements in their respective research domains. 
In the field of NLP, the highly processed nature of text data and its consistent problem formulation have paved the way for the development of large-scale foundational models. These models, such as GPT-3, have demonstrated remarkable performance across a wide range of NLP tasks. The unified nature of text data allows for the application of these models to various tasks without significant modifications, leveraging the semantic richness and the consistent problem formulation.
On the other hand, the CV community faces different challenges. Visual data, being raw information sources, require extensive representation learning and specific downstream application stages to obtain effective and processed visual representations. The complexity of visual data and the diversity of visual tasks make it more challenging to develop a unified foundational model that can be applied across the board. As a result, researchers in the CV domain are constantly exploring novel representation learning techniques and task-specific approaches to address the intricacies of visual data and achieve state-of-the-art performance in complex visual tasks.
As for audio data, researchers are following either community based on the specific audio types as well as the specific task requirements.

Another aspect to understand the logic between data nature and the methodology design focuses on the data format. The format, whether continuous or discrete, plays a crucial role in determining the suitable model architecture for effective processing. 
For continuous data, such as images and ambient sound, the continuity in the spatial or temporal dimensions usually benefit from model architectures like CNNs that are specifically designed to handle spatial and temporal dependencies and to capture the local and global correlations within the data.
As for data with discrete formats, such as MIDI musical representations or textual word tokens, models like Transformers are more appropriate to model the dependencies among discrete elements.

\vspace{-0.1in}
\subsection{Future Directions and Challenge}
\label{subsec:future}

As we have extensively discussed in this survey paper, the research in multimodal machine learning is diversely spread from general representation learning to detailed downstream tasks within a specific field. 

After having introduced various discriminative and generative multimodal applications that involve the vision and data of other modalities, we revisit and summarize the existing works from the perspective of their technique designs and the connections with respect to the data attributes.

For the discriminative tasks that involve the visual and audio data, we can observe from the introduced existing works that majority of them follow the general pipeline that contains separate data encoders, cross-modality attention feature fusions, as well as a decoder module designed for various task objectives. It is worth noting that all of the existing works process the ambient audio data as an entirety without specifically looking into the acoustic features of the audio signals. For example, certain type of ambient audio signals may include higher pitch and frequency than others, which can be used as a strong supplementary indicator for purely vision based recognition. 
In contrast, existing audio-involved generative works have explored more the disentangled features such as rhythms, pitches, and genres for either both synthesis and editing purposes.
As for the vision and text (natural language) combination, representative early \re{classic} methods often use the LSTM model to deal with the textual language data with word orderings.
Later, the success of Transformer model has encouraged the fast technical transition from LSTM to Transformers for the text processing branch in the multimodal learning context.


Return to current multimodal studies, while great success has been achieved within recent years, the challenges for the future research remain.
From the technical perspective, we believe that we can summarize the future research directions into two direction with regard to the connections with the data modalities.
On the one hand, the research community is seeking to establish a unified and general model that efficiently learns the representation of all the modalities of interest. Such a unified model, similar to the large-scale pre-training models we have introduced in Section~\ref{subsec:non-supervised}, should greatly help with various downstream applications such as specific cross-modality generations, interactive editing, and evaluations.
On the other hand, with increasing demand of more fine-grained and detailed applications in our daily life, we also expect to develop and achieve better performance for more specific and crafted tasks.

Another possible future direction for the multimodal learning could be the human intervention for ultimate multimodal perception AI systems. 
As our ultimate objective for multimodal learning is to bring intelligence to machines as real humans, human intervention could be a critical part to guide the general research direction in this fast developing area.   
A concrete example could be involving the human to provide more controls on the cross-modality generations and several downstream tasking such as editing~\cite{menapace2021playable,menapace2022playable} .

\vspace{-0.17in}
\section{Conclusions}
\label{sec:conclusion}

In this paper, we present a survey on the multimodal learning domain from an unique perspective of data characteristics. 
We start by mainly analyzing the intrinsic natures of different data modalities for the vision, audio, and text. 
We proceed to introduce the multimodal representation learning, in which we mainly categorize the current literature by their learning settings. 
Following the general representation learning in the multimodal field, we then introduce the concrete task applications from both discriminative and generative natures, each structured into sub-classes with specific data combinations in the form of ``Vision + X''.
For the discriminative tasks, we also include a revisit in light of data after presenting the task-specific works, where we provide our analysis to bridge the existing technique designs and their connections to the data nature from different modalities.
For the generative tasks, we include an introduction to the popular generative backbone models before diving into the detailed task explanations.  
Finally, we provide our discussions in terms of the challenges and future directions for the multimodal learning domain.


\vspace{-0.1in}
{
\bibliographystyle{ref.bst}
\bibliography{egbib}

\begin{thebibliography}{100}\itemsep=-1pt

\bibitem{Agarwal_2020_CVPR}
Vedika Agarwal, Rakshith Shetty, and Mario Fritz.
\newblock Towards causal vqa: Revealing and reducing spurious correlations by invariant and covariant semantic editing.
\newblock In {\em CVPR}, 2020.

\bibitem{aggarwal2021dance2music}
Gunjan Aggarwal and Devi Parikh.
\newblock Dance2music: Automatic dance-driven music generation.
\newblock {\em arXiv:2107.06252}, 2021.

\bibitem{aghajanyan2022cm3}
Armen Aghajanyan, Bernie Huang, Candace Ross, Vladimir Karpukhin, Hu Xu, Naman Goyal, Dmytro Okhonko, Mandar Joshi, Gargi Ghosh, Mike Lewis, et~al.
\newblock Cm3: A causal masked multimodal model of the internet.
\newblock {\em arXiv:2201.07520}, 2022.

\bibitem{akbari2021vatt}
Hassan Akbari, Liangzhe Yuan, Rui Qian, Wei-Hong Chuang, Shih-Fu Chang, Yin Cui, and Boqing Gong.
\newblock Vatt: Transformers for multimodal self-supervised learning from raw video, audio and text.
\newblock {\em NeurIPS}, pages 24206--24221, 2021.

\bibitem{alain2001and}
Claude Alain, Stephen~R Arnott, Stephanie Hevenor, Simon Graham, and Cheryl~L Grady.
\newblock “what” and “where” in the human auditory system.
\newblock {\em Proceedings of the national academy of sciences}, 98(21):12301--12306, 2001.

\bibitem{alamri2019audio}
Huda Alamri, Vincent Cartillier, Abhishek Das, Jue Wang, Anoop Cherian, Irfan Essa, Dhruv Batra, Tim~K Marks, Chiori Hori, Peter Anderson, et~al.
\newblock Audio visual scene-aware dialog.
\newblock In {\em CVPR}, 2019.

\bibitem{anderson2016spice}
Peter Anderson, Basura Fernando, Mark Johnson, and Stephen Gould.
\newblock Spice: Semantic propositional image caption evaluation.
\newblock In {\em ECCV}, 2016.

\bibitem{anne2017localizing}
Lisa Anne~Hendricks, Oliver Wang, Eli Shechtman, Josef Sivic, Trevor Darrell, and Bryan Russell.
\newblock Localizing moments in video with natural language.
\newblock In {\em ICCV}, pages 5803--5812, 2017.

\bibitem{antol2015vqa}
Stanislaw Antol, Aishwarya Agrawal, Jiasen Lu, Margaret Mitchell, Dhruv Batra, C Lawrence~Zitnick, and Devi Parikh.
\newblock Vqa: Visual question answering.
\newblock In {\em ICCV}, 2015.

\bibitem{av4-2017look}
Relja Arandjelovic and Andrew Zisserman.
\newblock Look, listen and learn.
\newblock In {\em ICCV}, 2017.

\bibitem{arjovsky2017wasserstein}
Martin Arjovsky, Soumith Chintala, and L{\'e}on Bottou.
\newblock Wasserstein generative adversarial networks.
\newblock In {\em ICML}, pages 214--223. PMLR, 2017.

\bibitem{austin2021structured}
Jacob Austin, Daniel Johnson, Jonathan Ho, Daniel Tarlow, and Rianne van~den Berg.
\newblock Structured denoising diffusion models in discrete state-spaces.
\newblock In {\em NeurIPS}, 2021.

\bibitem{avrahami2022blended}
Omri Avrahami, Dani Lischinski, and Ohad Fried.
\newblock Blended diffusion for text-driven editing of natural images.
\newblock In {\em CVPR}, 2022.

\bibitem{bahmaninezhad2019comprehensive}
Fahimeh Bahmaninezhad, Jian Wu, Rongzhi Gu, Shi-Xiong Zhang, Yong Xu, Meng Yu, and Dong Yu.
\newblock A comprehensive study of speech separation: spectrogram vs waveform separation.
\newblock {\em arXiv:1905.07497}, 2019.

\bibitem{baltruvsaitis2018multimodal}
Tadas Baltru{\v{s}}aitis, Chaitanya Ahuja, and Louis-Philippe Morency.
\newblock Multimodal machine learning: A survey and taxonomy.
\newblock {\em IEEE TPAMI}, 41(2):423--443, 2018.

\bibitem{banerjee2005meteor}
Satanjeev Banerjee and Alon Lavie.
\newblock Meteor: An automatic metric for mt evaluation with improved correlation with human judgments.
\newblock In {\em Proceedings of the acl workshop on intrinsic and extrinsic evaluation measures for machine translation and/or summarization}, pages 65--72, 2005.

\bibitem{bay2006surf}
Herbert Bay, Tinne Tuytelaars, and Luc~Van Gool.
\newblock Surf: Speeded up robust features.
\newblock In {\em ECCV}. Springer, 2006.

\bibitem{bayoudh2021survey}
Khaled Bayoudh, Raja Knani, Fay{\c{c}}al Hamdaoui, and Abdellatif Mtibaa.
\newblock A survey on deep multimodal learning for computer vision: advances, trends, applications, and datasets.
\newblock {\em The Visual Computer}, pages 1--32, 2021.

\bibitem{bommasani2021opportunities}
Rishi Bommasani, Drew~A Hudson, Ehsan Adeli, Russ Altman, Simran Arora, Sydney von Arx, Michael~S Bernstein, Jeannette Bohg, Antoine Bosselut, Emma Brunskill, et~al.
\newblock On the opportunities and risks of foundation models.
\newblock {\em arXiv:2108.07258}, 2021.

\bibitem{bornkessel2015neurobiological}
Ina Bornkessel-Schlesewsky, Matthias Schlesewsky, Steven~L Small, and Josef~P Rauschecker.
\newblock Neurobiological roots of language in primate audition: common computational properties.
\newblock {\em Trends in cognitive sciences}, 19(3):142--150, 2015.

\bibitem{bragg2019sign}
Danielle Bragg, Oscar Koller, Mary Bellard, Larwan Berke, Patrick Boudreault, Annelies Braffort, Naomi Caselli, Matt Huenerfauth, Hernisa Kacorri, Tessa Verhoef, et~al.
\newblock Sign language recognition, generation, and translation: An interdisciplinary perspective.
\newblock In {\em Proceedings of the 21st International ACM SIGACCESS Conference on Computers and Accessibility}, 2019.

\bibitem{briot2017deep}
Jean-Pierre Briot, Ga{\"e}tan Hadjeres, and Fran{\c{c}}ois-David Pachet.
\newblock Deep learning techniques for music generation--a survey.
\newblock {\em arXiv preprint arXiv:1709.01620}, 2017.

\bibitem{brock2018large}
Andrew Brock, Jeff Donahue, and Karen Simonyan.
\newblock Large scale gan training for high fidelity natural image synthesis.
\newblock {\em arXiv:1809.11096}, 2018.

\bibitem{sora2024}
Tim Brooks, Bill Peebles, Connor Holmes, Will DePue, Yufei Guo, Li Jing, David Schnurr, Joe Taylor, Troy Luhman, Eric Luhman, Clarence Ng, Ricky Wang, and Aditya Ramesh.
\newblock Video generation models as world simulators.
\newblock 2024.

\bibitem{brown2020language}
Tom Brown, Benjamin Mann, Nick Ryder, Melanie Subbiah, Jared~D Kaplan, Prafulla Dhariwal, Arvind Neelakantan, Pranav Shyam, Girish Sastry, Amanda Askell, et~al.
\newblock Language models are few-shot learners.
\newblock {\em NeurIPS}, pages 1877--1901, 2020.

\bibitem{caba2015activitynet}
Fabian Caba~Heilbron, Victor Escorcia, Bernard Ghanem, and Juan Carlos~Niebles.
\newblock Activitynet: A large-scale video benchmark for human activity understanding.
\newblock In {\em CVPR}, pages 961--970, 2015.

\bibitem{cao2014crema}
Houwei Cao, David~G Cooper, Michael~K Keutmann, Ruben~C Gur, Ani Nenkova, and Ragini Verma.
\newblock Crema-d: Crowd-sourced emotional multimodal actors dataset.
\newblock {\em IEEE transactions on affective computing}, 2014.

\bibitem{openpose}
Z. {Cao}, G. {Hidalgo Martinez}, T. {Simon}, S. {Wei}, and Y.~A. {Sheikh}.
\newblock Openpose: Realtime multi-person 2d pose estimation using part affinity fields.
\newblock {\em IEEE TPAMI}, 2019.

\bibitem{openpose-cao2017realtime}
Zhe Cao, Tomas Simon, Shih-En Wei, and Yaser Sheikh.
\newblock Realtime multi-person 2d pose estimation using part affinity fields.
\newblock In {\em CVPR}, 2017.

\bibitem{i3d}
Joao Carreira and Andrew Zisserman.
\newblock Quo vadis, action recognition? a new model and the kinetics dataset.
\newblock In {\em CVPR}, 2017.

\bibitem{chang2015shapenet}
Angel~X Chang, Thomas Funkhouser, Leonidas Guibas, Pat Hanrahan, Qixing Huang, Zimo Li, Silvio Savarese, Manolis Savva, Shuran Song, Hao Su, et~al.
\newblock Shapenet: An information-rich 3d model repository.
\newblock {\em arXiv:1512.03012}, 2015.

\bibitem{changpinyo2021conceptual}
Soravit Changpinyo, Piyush Sharma, Nan Ding, and Radu Soricut.
\newblock Conceptual 12m: Pushing web-scale image-text pre-training to recognize long-tail visual concepts.
\newblock In {\em CVPR}, 2021.

\bibitem{chatterjee2020sound2sight}
Moitreya Chatterjee and Anoop Cherian.
\newblock Sound2sight: Generating visual dynamics from sound and context.
\newblock In {\em ECCV}, pages 701--719. Springer, 2020.

\bibitem{chen2011collecting}
David Chen and William~B Dolan.
\newblock Collecting highly parallel data for paraphrase evaluation.
\newblock In {\em Proceedings of the 49th annual meeting of the association for computational linguistics: human language technologies}, pages 190--200, 2011.

\bibitem{chen2019variational}
Fuhai Chen, Rongrong Ji, Jiayi Ji, Xiaoshuai Sun, Baochang Zhang, Xuri Ge, Yongjian Wu, Feiyue Huang, and Yan Wang.
\newblock Variational structured semantic inference for diverse image captioning.
\newblock In {\em NeurIPS}, 2019.

\bibitem{chen2018visually}
Kan Chen, Chuanxi Zhang, Chen Fang, Zhaowen Wang, Trung Bui, and Ram Nevatia.
\newblock Visually indicated sound generation by perceptually optimized classification.
\newblock In {\em ECCV Workshops}, pages 0--0, 2018.

\bibitem{chen2018lip}
Lele Chen, Zhiheng Li, Ross~K Maddox, Zhiyao Duan, and Chenliang Xu.
\newblock Lip movements generation at a glance.
\newblock In {\em ECCV}, pages 520--535, 2018.

\bibitem{chen2019hierarchical}
Lele Chen, Ross~K Maddox, Zhiyao Duan, and Chenliang Xu.
\newblock Hierarchical cross-modal talking face generation with dynamic pixel-wise loss.
\newblock In {\em CVPR}, pages 7832--7841, 2019.

\bibitem{chen2020counterfactual}
Long Chen, Xin Yan, Jun Xiao, Hanwang Zhang, Shiliang Pu, and Yueting Zhuang.
\newblock Counterfactual samples synthesizing for robust visual question answering.
\newblock In {\em CVPR}, 2020.

\bibitem{chen2017sca}
Long Chen, Hanwang Zhang, Jun Xiao, Liqiang Nie, Jian Shao, Wei Liu, and Tat-Seng Chua.
\newblock Sca-cnn: Spatial and channel-wise attention in convolutional networks for image captioning.
\newblock In {\em CVPR}, 2017.

\bibitem{chen2020generative}
Mark Chen, Alec Radford, Rewon Child, Jeffrey Wu, Heewoo Jun, David Luan, and Ilya Sutskever.
\newblock Generative pretraining from pixels.
\newblock In {\em ICML}, pages 1691--1703. PMLR, 2020.

\bibitem{chen2020generating}
Peihao Chen, Yang Zhang, Mingkui Tan, Hongdong Xiao, Deng Huang, and Chuang Gan.
\newblock Generating visually aligned sound from videos.
\newblock {\em IEEE TIP}, 29:8292--8302, 2020.

\bibitem{chen2019semantic}
Shaoxiang Chen and Yu-Gang Jiang.
\newblock Semantic proposal for activity localization in videos via sentence query.
\newblock In {\em AAAI}, pages 8199--8206, 2019.

\bibitem{chen2020hierarchical}
Shaoxiang Chen and Yu-Gang Jiang.
\newblock Hierarchical visual-textual graph for temporal activity localization via language.
\newblock In {\em ECCV}, pages 601--618. Springer, 2020.

\bibitem{chen2020uniter}
Yen-Chun Chen, Linjie Li, Licheng Yu, Ahmed El~Kholy, Faisal Ahmed, Zhe Gan, Yu Cheng, and Jingjing Liu.
\newblock Uniter: Universal image-text representation learning.
\newblock In {\em ECCV}, pages 104--120. Springer, 2020.

\bibitem{cheng2017segflow}
Jingchun Cheng, Yi-Hsuan Tsai, Shengjin Wang, and Ming-Hsuan Yang.
\newblock Segflow: Joint learning for video object segmentation and optical flow.
\newblock In {\em ICCV}, pages 686--695, 2017.

\bibitem{chun2021probabilistic}
Sanghyuk Chun, Seong~Joon Oh, Rafael~Sampaio De~Rezende, Yannis Kalantidis, and Diane Larlus.
\newblock Probabilistic embeddings for cross-modal retrieval.
\newblock In {\em CVPR}, pages 8415--8424, 2021.

\bibitem{chung2017lip}
Joon~Son Chung and Andrew Zisserman.
\newblock Lip reading in the wild.
\newblock In {\em ACCV}. Springer, 2017.

\bibitem{chung2018unsupervised}
Yu-An Chung, Wei-Hung Weng, Schrasing Tong, and James Glass.
\newblock Unsupervised cross-modal alignment of speech and text embedding spaces.
\newblock {\em NeurIPS}, 31, 2018.

\bibitem{cobbe2019quantifying}
Karl Cobbe, Oleg Klimov, Chris Hesse, Taehoon Kim, and John Schulman.
\newblock Quantifying generalization in reinforcement learning.
\newblock In {\em ICML}. PMLR, 2019.

\bibitem{cortes1995support}
Corinna Cortes and Vladimir Vapnik.
\newblock Support-vector networks.
\newblock {\em Machine learning}, 20(3):273--297, 1995.

\bibitem{dagan2005pascal}
Ido Dagan, Oren Glickman, and Bernardo Magnini.
\newblock The pascal recognising textual entailment challenge.
\newblock In {\em Machine learning challenges workshop}, 2005.

\bibitem{dai2017towards}
Bo Dai, Sanja Fidler, Raquel Urtasun, and Dahua Lin.
\newblock Towards diverse and natural image descriptions via a conditional gan.
\newblock In {\em ICCV}, pages 2970--2979, 2017.

\bibitem{dalal2005histograms}
Navneet Dalal and Bill Triggs.
\newblock Histograms of oriented gradients for human detection.
\newblock In {\em CVPR}, 2005.

\bibitem{damen2018scaling}
Dima Damen, Hazel Doughty, Giovanni~Maria Farinella, Sanja Fidler, Antonino Furnari, Evangelos Kazakos, Davide Moltisanti, Jonathan Munro, Toby Perrett, Will Price, et~al.
\newblock Scaling egocentric vision: The epic-kitchens dataset.
\newblock In {\em ECCV}, 2018.

\bibitem{das2017human}
Abhishek Das, Harsh Agrawal, Larry Zitnick, Devi Parikh, and Dhruv Batra.
\newblock Human attention in visual question answering: Do humans and deep networks look at the same regions?
\newblock {\em CVIU}, 163:90--100, 2017.

\bibitem{das2017visual}
Abhishek Das, Satwik Kottur, Khushi Gupta, Avi Singh, Deshraj Yadav, Jos{\'e}~MF Moura, Devi Parikh, and Dhruv Batra.
\newblock Visual dialog.
\newblock In {\em CVPR}, 2017.

\bibitem{das2020speech}
Dipanjan Das, Sandika Biswas, Sanjana Sinha, and Brojeshwar Bhowmick.
\newblock Speech-driven facial animation using cascaded gans for learning of motion and texture.
\newblock In {\em ECCV}, pages 408--424, 2020.

\bibitem{davis2018visual}
Abe Davis and Maneesh Agrawala.
\newblock Visual rhythm and beat.
\newblock In {\em ACM Transactions on Graphics (TOG)}, 2018.

\bibitem{de2017guesswhat}
Harm De~Vries, Florian Strub, Sarath Chandar, Olivier Pietquin, Hugo Larochelle, and Aaron Courville.
\newblock Guesswhat?! visual object discovery through multi-modal dialogue.
\newblock In {\em CVPR}, 2017.

\bibitem{deng2018visual}
Chaorui Deng, Qi Wu, Qingyao Wu, Fuyuan Hu, Fan Lyu, and Mingkui Tan.
\newblock Visual grounding via accumulated attention.
\newblock In {\em CVPR}, pages 7746--7755, 2018.

\bibitem{deng2009imagenet}
Jia Deng, Wei Dong, Richard Socher, Li-Jia Li, Kai Li, and Li Fei-Fei.
\newblock Imagenet: A large-scale hierarchical image database.
\newblock In {\em CVPR}, 2009.

\bibitem{deng2021transvg}
Jiajun Deng, Zhengyuan Yang, Tianlang Chen, Wengang Zhou, and Houqiang Li.
\newblock Transvg: End-to-end visual grounding with transformers.
\newblock In {\em ICCV}, pages 1769--1779, 2021.

\bibitem{denton2018stochastic}
Emily Denton and Rob Fergus.
\newblock Stochastic video generation with a learned prior.
\newblock In {\em ICML}, pages 1174--1183. PMLR, 2018.

\bibitem{desai2021virtex}
Karan Desai and Justin Johnson.
\newblock Virtex: Learning visual representations from textual annotations.
\newblock In {\em CVPR}, pages 11162--11173, 2021.

\bibitem{devlin2018bert}
Jacob Devlin, Ming-Wei Chang, Kenton Lee, and Kristina Toutanova.
\newblock Bert: Pre-training of deep bidirectional transformers for language understanding.
\newblock {\em arXiv:1810.04805}, 2018.

\bibitem{jukebox}
Prafulla Dhariwal, Heewoo Jun, Christine Payne, Jong~Wook Kim, Alec Radford, and Ilya Sutskever.
\newblock Jukebox: A generative model for music.
\newblock {\em arXiv:2005.00341}, 2020.

\bibitem{dhariwal2021diff-img1}
Prafulla Dhariwal and Alexander Nichol.
\newblock Diffusion models beat gans on image synthesis.
\newblock In {\em NeurIPS}, 2021.

\bibitem{di2021video}
Shangzhe Di, Zeren Jiang, Si Liu, Zhaokai Wang, Leyan Zhu, Zexin He, Hongming Liu, and Shuicheng Yan.
\newblock Video background music generation with controllable music transformer.
\newblock In {\em ACMMM}, 2021.

\bibitem{di2019must}
Mattia~A Di~Gangi, Roldano Cattoni, Luisa Bentivogli, Matteo Negri, and Marco Turchi.
\newblock Must-c: a multilingual speech translation corpus.
\newblock In {\em 2019 Conference of the North American Chapter of the Association for Computational Linguistics: Human Language Technologies}, pages 2012--2017. Association for Computational Linguistics, 2019.

\bibitem{ding2021cogview}
Ming Ding, Zhuoyi Yang, Wenyi Hong, Wendi Zheng, Chang Zhou, Da Yin, Junyang Lin, Xu Zou, Zhou Shao, Hongxia Yang, et~al.
\newblock Cogview: Mastering text-to-image generation via transformers.
\newblock {\em NeurIPS}, pages 19822--19835, 2021.

\bibitem{dong2018pypianoroll}
Hao-Wen Dong, Wen-Yi Hsiao, and Yi-Hsuan Yang.
\newblock Pypianoroll: Open source python package for handling multitrack pianoroll.
\newblock {\em Proc. ISMIR. Late-breaking paper}, 2018.

\bibitem{duan2021audio}
Bin Duan, Hao Tang, Wei Wang, Ziliang Zong, Guowei Yang, and Yan Yan.
\newblock Audio-visual event localization via recursive fusion by joint co-attention.
\newblock In {\em WACV}, pages 4013--4022, 2021.

\bibitem{duarte2021how2sign}
Amanda Duarte, Shruti Palaskar, Lucas Ventura, Deepti Ghadiyaram, Kenneth DeHaan, Florian Metze, Jordi Torres, and Xavier Giro-i Nieto.
\newblock How2sign: a large-scale multimodal dataset for continuous american sign language.
\newblock In {\em CVPR}, 2021.

\bibitem{ephrat2017improved}
Ariel Ephrat, Tavi Halperin, and Shmuel Peleg.
\newblock Improved speech reconstruction from silent video.
\newblock In {\em ICCV Workshops}, pages 455--462, 2017.

\bibitem{ephrat2017vid2speech}
Ariel Ephrat and Shmuel Peleg.
\newblock Vid2speech: speech reconstruction from silent video.
\newblock In {\em ICASSP}, pages 5095--5099. IEEE, 2017.

\bibitem{freitag2017audeep}
Michael Freitag, Shahin Amiriparian, Sergey Pugachevskiy, Nicholas Cummins, and Bj{\"o}rn Schuller.
\newblock audeep: Unsupervised learning of representations from audio with deep recurrent neural networks.
\newblock {\em The Journal of Machine Learning Research}, 2017.

\bibitem{fukui2016multimodal}
Akira Fukui, Dong~Huk Park, Daylen Yang, Anna Rohrbach, Trevor Darrell, and Marcus Rohrbach.
\newblock Multimodal compact bilinear pooling for visual question answering and visual grounding.
\newblock {\em arXiv:1606.01847}, 2016.

\bibitem{foley}
Chuang Gan, Deng Huang, Peihao Chen, Joshua~B Tenenbaum, and Antonio Torralba.
\newblock Foley music: Learning to generate music from videos.
\newblock In {\em ECCV}, 2020.

\bibitem{gandhi2022esb}
Sanchit Gandhi, Patrick Von~Platen, and Alexander~M Rush.
\newblock Esb: A benchmark for multi-domain end-to-end speech recognition.
\newblock {\em arXiv:2210.13352}, 2022.

\bibitem{gao2017tall}
Jiyang Gao, Chen Sun, Zhenheng Yang, and Ram Nevatia.
\newblock Tall: Temporal activity localization via language query.
\newblock In {\em ICCV}, pages 5267--5275, 2017.

\bibitem{gemmeke2017audioset}
Jort~F Gemmeke, Daniel~PW Ellis, Dylan Freedman, Aren Jansen, Wade Lawrence, R~Channing Moore, Manoj Plakal, and Marvin Ritter.
\newblock Audio set: An ontology and human-labeled dataset for audio events.
\newblock In {\em ICASSP}. IEEE, 2017.

\bibitem{gan}
Ian Goodfellow, Jean Pouget-Abadie, Mehdi Mirza, Bing Xu, David Warde-Farley, Sherjil Ozair, Aaron Courville, and Yoshua Bengio.
\newblock Generative adversarial nets.
\newblock In {\em NeurIPS}, 2014.

\bibitem{goyal2017making}
Yash Goyal, Tejas Khot, Douglas Summers-Stay, Dhruv Batra, and Devi Parikh.
\newblock Making the v in vqa matter: Elevating the role of image understanding in visual question answering.
\newblock In {\em CVPR}, 2017.

\bibitem{graves2013generating}
Alex Graves.
\newblock Generating sequences with recurrent neural networks.
\newblock {\em arXiv:1308.0850}, 2013.

\bibitem{grumiaux2022survey}
Pierre-Amaury Grumiaux, Srdjan Kitic, Laurent Girin, and Alexandre Gu{\'e}rin.
\newblock A survey of sound source localization with deep learning methods.
\newblock {\em The Journal of the Acoustical Society of America}, pages 107--151, 2022.

\bibitem{gu2023mamba}
Albert Gu and Tri Dao.
\newblock Mamba: Linear-time sequence modeling with selective state spaces.
\newblock {\em arXiv:2312.00752}, 2023.

\bibitem{gu2018look}
Jiuxiang Gu, Jianfei Cai, Shafiq~R Joty, Li Niu, and Gang Wang.
\newblock Look, imagine and match: Improving textual-visual cross-modal retrieval with generative models.
\newblock In {\em CVPR}, pages 7181--7189, 2018.

\bibitem{gu2021vector}
Shuyang Gu, Dong Chen, Jianmin Bao, Fang Wen, Bo Zhang, Dongdong Chen, Lu Yuan, and Baining Guo.
\newblock Vector quantized diffusion model for text-to-image synthesis.
\newblock In {\em CVPR}, 2022.

\bibitem{gulrajani2017improved}
Ishaan Gulrajani, Faruk Ahmed, Martin Arjovsky, Vincent Dumoulin, and Aaron~C Courville.
\newblock Improved training of wasserstein gans.
\newblock {\em NeurIPS}, 30, 2017.

\bibitem{guo2019deep}
Wenzhong Guo, Jianwen Wang, and Shiping Wang.
\newblock Deep multimodal representation learning: A survey.
\newblock {\em IEEE Access}, 7:63373--63394, 2019.

\bibitem{harwath2015deep}
David Harwath and James Glass.
\newblock Deep multimodal semantic embeddings for speech and images.
\newblock In {\em 2015 IEEE Workshop on Automatic Speech Recognition and Understanding (ASRU)}. IEEE, 2015.

\bibitem{hayes2022mugen}
Thomas Hayes, Songyang Zhang, Xi Yin, Guan Pang, Sasha Sheng, Harry Yang, Songwei Ge, Isabelle Hu, and Devi Parikh.
\newblock Mugen: A playground for video-audio-text multimodal understanding and generation.
\newblock {\em arXiv:2204.08058}, 2022.

\bibitem{he2022masked}
Kaiming He, Xinlei Chen, Saining Xie, Yanghao Li, Piotr Doll{\'a}r, and Ross Girshick.
\newblock Masked autoencoders are scalable vision learners.
\newblock In {\em CVPR}, 2022.

\bibitem{he2016deep}
Kaiming He, Xiangyu Zhang, Shaoqing Ren, and Jian Sun.
\newblock Deep residual learning for image recognition.
\newblock In {\em CVPR}, 2016.

\bibitem{hershey2017cnn}
Shawn Hershey, Sourish Chaudhuri, Daniel~PW Ellis, Jort~F Gemmeke, Aren Jansen, R~Channing Moore, Manoj Plakal, Devin Platt, Rif~A Saurous, Bryan Seybold, et~al.
\newblock Cnn architectures for large-scale audio classification.
\newblock In {\em ICASSP}. IEEE, 2017.

\bibitem{hessel2021clipscore}
Jack Hessel, Ari Holtzman, Maxwell Forbes, Ronan~Le Bras, and Yejin Choi.
\newblock {CLIPScore:} a reference-free evaluation metric for image captioning.
\newblock In {\em EMNLP}, 2021.

\bibitem{heusel2017gans}
Martin Heusel, Hubert Ramsauer, Thomas Unterthiner, Bernhard Nessler, and Sepp Hochreiter.
\newblock Gans trained by a two time-scale update rule converge to a local nash equilibrium.
\newblock {\em NeurIPS}, 30, 2017.

\bibitem{higgins2016beta}
Irina Higgins, Loic Matthey, Arka Pal, Christopher Burgess, Xavier Glorot, Matthew Botvinick, Shakir Mohamed, and Alexander Lerchner.
\newblock beta-vae: Learning basic visual concepts with a constrained variational framework.
\newblock In {\em ICLR}, 2017.

\bibitem{hinton2006reducing}
Geoffrey~E Hinton and Ruslan~R Salakhutdinov.
\newblock Reducing the dimensionality of data with neural networks.
\newblock {\em science}, 313(5786):504--507, 2006.

\bibitem{ho2020dpm}
Jonathan Ho, Ajay Jain, and Pieter Abbeel.
\newblock Denoising diffusion probabilistic models.
\newblock In {\em NeurIPS}, 2020.

\bibitem{ho2022diff-img2}
Jonathan Ho, Chitwan Saharia, William Chan, David~J Fleet, Mohammad Norouzi, and Tim Salimans.
\newblock Cascaded diffusion models for high fidelity image generation.
\newblock {\em Journal of Machine Learning Research}, 2022.

\bibitem{ho2022classifier}
Jonathan Ho and Tim Salimans.
\newblock Classifier-free diffusion guidance.
\newblock {\em arXiv:2207.12598}, 2022.

\bibitem{ho2022video}
Jonathan Ho, Tim Salimans, Alexey Gritsenko, William Chan, Mohammad Norouzi, and David~J Fleet.
\newblock Video diffusion models.
\newblock {\em arXiv:2204.03458}, 2022.

\bibitem{hochreiter1997long}
Sepp Hochreiter and J{\"u}rgen Schmidhuber.
\newblock Long short-term memory.
\newblock {\em Neural computation}, 9(8):1735--1780, 1997.

\bibitem{hong2019learning}
Richang Hong, Daqing Liu, Xiaoyu Mo, Xiangnan He, and Hanwang Zhang.
\newblock Learning to compose and reason with language tree structures for visual grounding.
\newblock {\em IEEE TPAMI}, 2019.

\bibitem{hori2019end}
Chiori Hori, Huda Alamri, Jue Wang, Gordon Wichern, Takaaki Hori, Anoop Cherian, Tim~K Marks, Vincent Cartillier, Raphael~Gontijo Lopes, Abhishek Das, et~al.
\newblock End-to-end audio visual scene-aware dialog using multimodal attention-based video features.
\newblock In {\em ICASSP}. IEEE, 2019.

\bibitem{horn1981determining}
Berthold~KP Horn and Brian~G Schunck.
\newblock Determining optical flow.
\newblock {\em Artificial intelligence}, 17(1-3):185--203, 1981.

\bibitem{hu2021global}
Minghui Hu, Yujie Wang, Tat-Jen Cham, Jianfei Yang, and PN Suganthan.
\newblock Global context with discrete diffusion in vector quantised modelling for image generation.
\newblock {\em arXiv:2112.01799}, 2021.

\bibitem{hu2016natural}
Ronghang Hu, Huazhe Xu, Marcus Rohrbach, Jiashi Feng, Kate Saenko, and Trevor Darrell.
\newblock Natural language object retrieval.
\newblock In {\em CVPR}, pages 4555--4564, 2016.

\bibitem{hu2022make}
Yaosi Hu, Chong Luo, and Zhenzhong Chen.
\newblock Make it move: controllable image-to-video generation with text descriptions.
\newblock In {\em CVPR}, pages 18219--18228, 2022.

\bibitem{huang2022deconfounded}
Jianqiang Huang, Yu Qin, Jiaxin Qi, Qianru Sun, and Hanwang Zhang.
\newblock Deconfounded visual grounding.
\newblock In {\em AAAI}, pages 998--1006, 2022.

\bibitem{huh2023epic}
Jaesung Huh, Jacob Chalk, Evangelos Kazakos, Dima Damen, and Andrew Zisserman.
\newblock Epic-sounds: A large-scale dataset of actions that sound.
\newblock In {\em ICASSP}. IEEE, 2023.

\bibitem{isola2017image}
Phillip Isola, Jun-Yan Zhu, Tinghui Zhou, and Alexei~A Efros.
\newblock Image-to-image translation with conditional adversarial networks.
\newblock In {\em CVPR}, 2017.

\bibitem{jaimes2007multimodal}
Alejandro Jaimes and Nicu Sebe.
\newblock Multimodal human--computer interaction: A survey.
\newblock {\em CVIU}, 108(1-2):116--134, 2007.

\bibitem{jain2018two}
Unnat Jain, Svetlana Lazebnik, and Alexander~G Schwing.
\newblock Two can play this game: visual dialog with discriminative question generation and answering.
\newblock In {\em CVPR}, 2018.

\bibitem{javed2022towards}
Tahir Javed, Sumanth Doddapaneni, Abhigyan Raman, Kaushal~Santosh Bhogale, Gowtham Ramesh, Anoop Kunchukuttan, Pratyush Kumar, and Mitesh~M Khapra.
\newblock Towards building asr systems for the next billion users.
\newblock In {\em AAAI}, 2022.

\bibitem{jensen2016algorithm}
Jesper Jensen and Cees~H Taal.
\newblock An algorithm for predicting the intelligibility of speech masked by modulated noise maskers.
\newblock {\em IEEE/ACM Transactions on Audio, Speech, and Language Processing}, 24(11):2009--2022, 2016.

\bibitem{jin2022embracing}
Yang Jin, Zehuan Yuan, Yadong Mu, et~al.
\newblock Embracing consistency: A one-stage approach for spatio-temporal video grounding.
\newblock {\em NeurIPS}, 2022.

\bibitem{johnson2017clevr}
Justin Johnson, Bharath Hariharan, Laurens Van Der~Maaten, Li Fei-Fei, C Lawrence~Zitnick, and Ross Girshick.
\newblock Clevr: A diagnostic dataset for compositional language and elementary visual reasoning.
\newblock In {\em CVPR}, 2017.

\bibitem{kamath2021mdetr}
Aishwarya Kamath, Mannat Singh, Yann LeCun, Gabriel Synnaeve, Ishan Misra, and Nicolas Carion.
\newblock Mdetr-modulated detection for end-to-end multi-modal understanding.
\newblock In {\em ICCV}, 2021.

\bibitem{karpathy2014deep}
Andrej Karpathy, Armand Joulin, and Li~F Fei-Fei.
\newblock Deep fragment embeddings for bidirectional image sentence mapping.
\newblock {\em NeurIPS}, 27, 2014.

\bibitem{kim2022diffusionclip}
Gwanghyun Kim, Taesung Kwon, and Jong~Chul Ye.
\newblock Diffusionclip: Text-guided diffusion models for robust image manipulation.
\newblock In {\em CVPR}, 2022.

\bibitem{kim2021lip}
Minsu Kim, Joanna Hong, and Yong~Man Ro.
\newblock Lip to speech synthesis with visual context attentional gan.
\newblock {\em NeurIPS}, pages 2758--2770, 2021.

\bibitem{theo-diff3}
Diederik~P Kingma, Tim Salimans, Ben Poole, and Jonathan Ho.
\newblock Variational diffusion models.
\newblock In {\em NeurIPS}, 2021.

\bibitem{vae-kingma2013auto}
Diederik~P Kingma and Max Welling.
\newblock Auto-encoding variational bayes.
\newblock In {\em ICLR}, 2014.

\bibitem{hifigan}
Jungil Kong, Jaehyeon Kim, and Jaekyoung Bae.
\newblock Hifi-gan: Generative adversarial networks for efficient and high fidelity speech synthesis.
\newblock In {\em NeurIPS}, 2020.

\bibitem{kong2020diffwave}
Zhifeng Kong, Wei Ping, Jiaji Huang, Kexin Zhao, and Bryan Catanzaro.
\newblock Diffwave: A versatile diffusion model for audio synthesis.
\newblock In {\em ICLR}, 2020.

\bibitem{krishna2017dense}
Ranjay Krishna, Kenji Hata, Frederic Ren, Li Fei-Fei, and Juan Carlos~Niebles.
\newblock Dense-captioning events in videos.
\newblock In {\em ICCV}, pages 706--715, 2017.

\bibitem{krishna2017visualgenome}
Ranjay Krishna, Yuke Zhu, Oliver Groth, Justin Johnson, Kenji Hata, Joshua Kravitz, Stephanie Chen, Yannis Kalantidis, Li-Jia Li, David~A Shamma, et~al.
\newblock Visual genome: Connecting language and vision using crowdsourced dense image annotations.
\newblock {\em IJCV}, 2017.

\bibitem{krizhevsky2012imagenet}
Alex Krizhevsky, Ilya Sutskever, and Geoffrey~E Hinton.
\newblock Imagenet classification with deep convolutional neural networks.
\newblock {\em NeurIPS}, 2012.

\bibitem{melgan}
Kundan Kumar, Rithesh Kumar, Thibault de Boissiere, Lucas Gestin, Wei~Zhen Teoh, Jose Sotelo, Alexandre de Br{\'e}bisson, Yoshua Bengio, and Aaron~C Courville.
\newblock Melgan: Generative adversarial networks for conditional waveform synthesis.
\newblock In {\em NeurIPS}, 2019.

\bibitem{kwon2022diffusion}
Mingi Kwon, Jaeseok Jeong, and Youngjung Uh.
\newblock Diffusion models already have a semantic latent space.
\newblock In {\em ICLR}, 2023.

\bibitem{lee2021nu}
Junhyeok Lee and Seungu Han.
\newblock Nu-wave: A diffusion probabilistic model for neural audio upsampling.
\newblock {\em Proc. Interspeech 2021}, 2021.

\bibitem{lei2021less}
Jie Lei, Linjie Li, Luowei Zhou, Zhe Gan, Tamara~L Berg, Mohit Bansal, and Jingjing Liu.
\newblock Less is more: Clipbert for video-and-language learning via sparse sampling.
\newblock In {\em CVPR}, pages 7331--7341, 2021.

\bibitem{li2023progressive}
Guangyao Li, Wenxuan Hou, and Di Hu.
\newblock Progressive spatio-temporal perception for audio-visual question answering.
\newblock In {\em ACMM}, 2023.

\bibitem{li2022learning}
Guangyao Li, Yake Wei, Yapeng Tian, Chenliang Xu, Ji-Rong Wen, and Di Hu.
\newblock Learning to answer questions in dynamic audio-visual scenarios.
\newblock In {\em CVPR}, 2022.

\bibitem{aistplus}
Ruilong Li, Shan Yang, David~A. Ross, and Angjoo Kanazawa.
\newblock Ai choreographer: Music conditioned 3d dance generation with aist++.
\newblock In {\em ICCV}, 2021.

\bibitem{li2019object}
Wenbo Li, Pengchuan Zhang, Lei Zhang, Qiuyuan Huang, Xiaodong He, Siwei Lyu, and Jianfeng Gao.
\newblock Object-driven text-to-image synthesis via adversarial training.
\newblock In {\em CVPR}, pages 12174--12182, 2019.

\bibitem{li2020visual}
Yunzhu Li, Toru Lin, Kexin Yi, Daniel Bear, Daniel Yamins, Jiajun Wu, Joshua Tenenbaum, and Antonio Torralba.
\newblock Visual grounding of learned physical models.
\newblock In {\em ICML}, pages 5927--5936. PMLR, 2020.

\bibitem{li2018video}
Yitong Li, Martin Min, Dinghan Shen, David Carlson, and Lawrence Carin.
\newblock Video generation from text.
\newblock In {\em AAAI}, 2018.

\bibitem{li2017scene}
Yikang Li, Wanli Ouyang, Bolei Zhou, Kun Wang, and Xiaogang Wang.
\newblock Scene graph generation from objects, phrases and region captions.
\newblock In {\em ICCV}, 2017.

\bibitem{liang2022foundations}
Paul~Pu Liang, Amir Zadeh, and Louis-Philippe Morency.
\newblock Foundations and recent trends in multimodal machine learning: Principles, challenges, and open questions.
\newblock {\em arXiv:2209.03430}, 2022.

\bibitem{lin2004rouge}
Chin-Yew Lin.
\newblock Rouge: A package for automatic evaluation of summaries.
\newblock In {\em Text summarization branches out}, pages 74--81, 2004.

\bibitem{lin2014microsoft}
Tsung-Yi Lin, Michael Maire, Serge Belongie, James Hays, Pietro Perona, Deva Ramanan, Piotr Doll{\'a}r, and C~Lawrence Zitnick.
\newblock Microsoft coco: Common objects in context.
\newblock In {\em ECCV}. Springer, 2014.

\bibitem{lin2021exploring}
Yan-Bo Lin, Hung-Yu Tseng, Hsin-Ying Lee, Yen-Yu Lin, and Ming-Hsuan Yang.
\newblock Exploring cross-video and cross-modality signals for weakly-supervised audio-visual video parsing.
\newblock {\em NeurIPS}, 34:11449--11461, 2021.

\bibitem{lin2020audiovisual}
Yan-Bo Lin and Yu-Chiang~Frank Wang.
\newblock Audiovisual transformer with instance attention for audio-visual event localization.
\newblock In {\em ACCV}, 2020.

\bibitem{liu2019learning}
Daqing Liu, Hanwang Zhang, Feng Wu, and Zheng-Jun Zha.
\newblock Learning to assemble neural module tree networks for visual grounding.
\newblock In {\em ICCV}, pages 4673--4682, 2019.

\bibitem{liu2022compositional}
Nan Liu, Shuang Li, Yilun Du, Antonio Torralba, and Joshua~B Tenenbaum.
\newblock Compositional visual generation with composable diffusion models.
\newblock In {\em ECCV}, 2022.

\bibitem{liu2017improved}
Siqi Liu, Zhenhai Zhu, Ning Ye, Sergio Guadarrama, and Kevin Murphy.
\newblock Improved image captioning via policy gradient optimization of spider.
\newblock In {\em ICCV}, pages 873--881, 2017.

\bibitem{liu2019use}
Yang Liu, Samuel Albanie, Arsha Nagrani, and Andrew Zisserman.
\newblock Use what you have: Video retrieval using representations from collaborative experts.
\newblock {\em arXiv:1907.13487}, 2019.

\bibitem{liu2021relation}
Yongfei Liu, Bo Wan, Lin Ma, and Xuming He.
\newblock Relation-aware instance refinement for weakly supervised visual grounding.
\newblock In {\em CVPR}, pages 5612--5621, 2021.

\bibitem{liu2020learning}
Yongfei Liu, Bo Wan, Xiaodan Zhu, and Xuming He.
\newblock Learning cross-modal context graph for visual grounding.
\newblock In {\em AAAI}, pages 11645--11652, 2020.

\bibitem{long2015fully}
Jonathan Long, Evan Shelhamer, and Trevor Darrell.
\newblock Fully convolutional networks for semantic segmentation.
\newblock In {\em CVPR}, 2015.

\bibitem{SMPL:2015}
Matthew Loper, Naureen Mahmood, Javier Romero, Gerard Pons-Moll, and Michael~J. Black.
\newblock {SMPL}: A skinned multi-person linear model.
\newblock {\em ACM Trans. Graphics (Proc. SIGGRAPH Asia)}, 34, 2015.

\bibitem{lowe2004SIFT}
David~G Lowe.
\newblock Distinctive image features from scale-invariant keypoints.
\newblock {\em IJCV}, 60(2):91--110, 2004.

\bibitem{lu2016visual}
Cewu Lu, Ranjay Krishna, Michael Bernstein, and Li Fei-Fei.
\newblock Visual relationship detection with language priors.
\newblock In {\em ECCV}. Springer, 2016.

\bibitem{lu2019vilbert}
Jiasen Lu, Dhruv Batra, Devi Parikh, and Stefan Lee.
\newblock Vilbert: Pretraining task-agnostic visiolinguistic representations for vision-and-language tasks.
\newblock {\em NeurIPS}, 2019.

\bibitem{lu2017knowing}
Jiasen Lu, Caiming Xiong, Devi Parikh, and Richard Socher.
\newblock Knowing when to look: Adaptive attention via a visual sentinel for image captioning.
\newblock In {\em CVPR}, 2017.

\bibitem{luo2020univl}
Huaishao Luo, Lei Ji, Botian Shi, Haoyang Huang, Nan Duan, Tianrui Li, Jason Li, Taroon Bharti, and Ming Zhou.
\newblock Univl: A unified video and language pre-training model for multimodal understanding and generation.
\newblock {\em arXiv:2002.06353}, 2020.

\bibitem{mao2016generation}
Junhua Mao, Jonathan Huang, Alexander Toshev, Oana Camburu, Alan~L Yuille, and Kevin Murphy.
\newblock Generation and comprehension of unambiguous object descriptions.
\newblock In {\em CVPR}, pages 11--20, 2016.

\bibitem{mao2014deep}
Junhua Mao, Wei Xu, Yi Yang, Jiang Wang, Zhiheng Huang, and Alan Yuille.
\newblock Deep captioning with multimodal recurrent neural networks (m-rnn).
\newblock {\em arXiv:1412.6632}, 2014.

\bibitem{marino2019ok}
Kenneth Marino, Mohammad Rastegari, Ali Farhadi, and Roozbeh Mottaghi.
\newblock Ok-vqa: A visual question answering benchmark requiring external knowledge.
\newblock In {\em CVPR}, 2019.

\bibitem{menapace2022playable}
Willi Menapace, St{\'e}phane Lathuili{\`e}re, Aliaksandr Siarohin, Christian Theobalt, Sergey Tulyakov, Vladislav Golyanik, and Elisa Ricci.
\newblock Playable environments: Video manipulation in space and time.
\newblock In {\em CVPR}, pages 3584--3593, 2022.

\bibitem{menapace2021playable}
Willi Menapace, St{\'e}phane Lathuili{\`e}re, Sergey Tulyakov, Aliaksandr Siarohin, and Elisa Ricci.
\newblock Playable video generation.
\newblock In {\em CVPR}, pages 10061--10070, 2021.

\bibitem{michelsanti2020vocoder}
Daniel Michelsanti, Olga Slizovskaia, Gloria Haro, Emilia G{\'o}mez, Zheng-Hua Tan, and Jesper Jensen.
\newblock Vocoder-based speech synthesis from silent videos.
\newblock {\em arXiv:2004.02541}, 2020.

\bibitem{miech2018learning}
Antoine Miech, Ivan Laptev, and Josef Sivic.
\newblock Learning a text-video embedding from incomplete and heterogeneous data.
\newblock {\em arXiv:1804.02516}, 2018.

\bibitem{mikolov2013efficient}
Tomas Mikolov, Kai Chen, Greg Corrado, and Jeffrey Dean.
\newblock Efficient estimation of word representations in vector space.
\newblock {\em arXiv preprint arXiv:1301.3781}, 2013.

\bibitem{mira2022svts}
Rodrigo Mira, Alexandros Haliassos, Stavros Petridis, Bj{\"o}rn~W Schuller, and Maja Pantic.
\newblock Svts: Scalable video-to-speech synthesis.
\newblock {\em arXiv:2205.02058}, 2022.

\bibitem{mirza2014conditional}
Mehdi Mirza and Simon Osindero.
\newblock Conditional generative adversarial nets.
\newblock {\em arXiv:1411.1784}, 2014.

\bibitem{mittal2021symbolic-diff}
Gautam Mittal, Jesse Engel, Curtis Hawthorne, and Ian Simon.
\newblock Symbolic music generation with diffusion models.
\newblock {\em arXiv:2103.16091}, 2021.

\bibitem{akash_nagaraj_2018}
Akash Nagaraj.
\newblock Asl alphabet, 2018.

\bibitem{ni2021m3p}
Minheng Ni, Haoyang Huang, Lin Su, Edward Cui, Taroon Bharti, Lijuan Wang, Dongdong Zhang, and Nan Duan.
\newblock M3p: Learning universal representations via multitask multilingual multimodal pre-training.
\newblock In {\em CVPR}, pages 3977--3986, 2021.

\bibitem{nichol2021glide}
Alex Nichol, Prafulla Dhariwal, Aditya Ramesh, Pranav Shyam, Pamela Mishkin, Bob McGrew, Ilya Sutskever, and Mark Chen.
\newblock Glide: Towards photorealistic image generation and editing with text-guided diffusion models.
\newblock {\em arXiv:2112.10741}, 2021.

\bibitem{nichol2021improveddmp}
Alexander~Quinn Nichol and Prafulla Dhariwal.
\newblock Improved denoising diffusion probabilistic models.
\newblock In {\em ICML}. PMLR, 2021.

\bibitem{Niu_2021_CVPR}
Yulei Niu, Kaihua Tang, Hanwang Zhang, Zhiwu Lu, Xian-Sheng Hua, and Ji-Rong Wen.
\newblock Counterfactual vqa: A cause-effect look at language bias.
\newblock In {\em CVPR}, pages 12700--12710, 2021.

\bibitem{oh2011large}
Sangmin Oh, Anthony Hoogs, Amitha Perera, Naresh Cuntoor, Chia-Chih Chen, Jong~Taek Lee, Saurajit Mukherjee, JK Aggarwal, Hyungtae Lee, Larry Davis, et~al.
\newblock A large-scale benchmark dataset for event recognition in surveillance video.
\newblock In {\em CVPR}, pages 3153--3160, 2011.

\bibitem{oncescu2021audio}
Andreea-Maria Oncescu, A Koepke, Joao~F Henriques, Zeynep Akata, and Samuel Albanie.
\newblock Audio retrieval with natural language queries.
\newblock In {\em INTERSPEECH}, 2021.

\bibitem{vqvae}
Aaron van~den Oord, Oriol Vinyals, and Koray Kavukcuoglu.
\newblock Neural discrete representation learning.
\newblock In {\em NeurIPS}, 2017.

\bibitem{oya2020we}
Takashi Oya, Shohei Iwase, Ryota Natsume, Takahiro Itazuri, Shugo Yamaguchi, and Shigeo Morishima.
\newblock Do we need sound for sound source localization?
\newblock In {\em ACCV}, 2020.

\bibitem{pan2023drag}
Xingang Pan, Ayush Tewari, Thomas Leimk{\"u}hler, Lingjie Liu, Abhimitra Meka, and Christian Theobalt.
\newblock Drag your gan: Interactive point-based manipulation on the generative image manifold.
\newblock {\em arXiv:2305.10973}, 2023.

\bibitem{papineni2002bleu}
Kishore Papineni, Salim Roukos, Todd Ward, and Wei-Jing Zhu.
\newblock Bleu: a method for automatic evaluation of machine translation.
\newblock In {\em ACL}, 2002.

\bibitem{perazzi2016benchmark}
Federico Perazzi, Jordi Pont-Tuset, Brian McWilliams, Luc Van~Gool, Markus Gross, and Alexander Sorkine-Hornung.
\newblock A benchmark dataset and evaluation methodology for video object segmentation.
\newblock In {\em CVPR}, pages 724--732, 2016.

\bibitem{inbook}
Sergio Picazo-Vela and Luis Hernandez.
\newblock {\em Technology, Science, and Culture: A Global Vision}.
\newblock 02 2019.

\bibitem{plapous2006improvedsnr}
Cyril Plapous, Claude Marro, and Pascal Scalart.
\newblock Improved signal-to-noise ratio estimation for speech enhancement.
\newblock {\em IEEE transactions on audio, speech, and language processing}, 2006.

\bibitem{plummer2017phrase}
Bryan~A Plummer, Arun Mallya, Christopher~M Cervantes, Julia Hockenmaier, and Svetlana Lazebnik.
\newblock Phrase localization and visual relationship detection with comprehensive image-language cues.
\newblock In {\em ICCV}, pages 1928--1937, 2017.

\bibitem{Prajwal_2020_CVPR}
K~R Prajwal, Rudrabha Mukhopadhyay, Vinay~P. Namboodiri, and C.V. Jawahar.
\newblock Learning individual speaking styles for accurate lip to speech synthesis.
\newblock In {\em CVPR}, 2020.

\bibitem{qi2017pointnet}
Charles~R Qi, Hao Su, Kaichun Mo, and Leonidas~J Guibas.
\newblock Pointnet: Deep learning on point sets for 3d classification and segmentation.
\newblock In {\em CVPR}, 2017.

\bibitem{qi2020two}
Jiaxin Qi, Yulei Niu, Jianqiang Huang, and Hanwang Zhang.
\newblock Two causal principles for improving visual dialog.
\newblock In {\em CVPR}, pages 10860--10869, 2020.

\bibitem{qian2020multiple}
Rui Qian, Di Hu, Heinrich Dinkel, Mengyue Wu, Ning Xu, and Weiyao Lin.
\newblock Multiple sound sources localization from coarse to fine.
\newblock In {\em ECCV}, pages 292--308. Springer, 2020.

\bibitem{rachavarapu2021localize}
Kranthi~Kumar Rachavarapu, Vignesh Sundaresha, AN Rajagopalan, et~al.
\newblock Localize to binauralize: Audio spatialization from visual sound source localization.
\newblock In {\em ICCV}, pages 1930--1939, 2021.

\bibitem{radford2021clip}
Alec Radford, Jong~Wook Kim, Chris Hallacy, Aditya Ramesh, Gabriel Goh, Sandhini Agarwal, Girish Sastry, Amanda Askell, Pamela Mishkin, Jack Clark, et~al.
\newblock Learning transferable visual models from natural language supervision.
\newblock In {\em ICML}. PMLR, 2021.

\bibitem{radford2019gpt3}
Alec Radford, Jeffrey Wu, Rewon Child, David Luan, Dario Amodei, Ilya Sutskever, et~al.
\newblock Language models are unsupervised multitask learners.
\newblock {\em OpenAI blog}, 2019.

\bibitem{ramesh2022dalle2}
Aditya Ramesh, Prafulla Dhariwal, Alex Nichol, Casey Chu, and Mark Chen.
\newblock Hierarchical text-conditional image generation with clip latents.
\newblock {\em arXiv:2204.06125}, 2022.

\bibitem{ramesh2021zero}
Aditya Ramesh, Mikhail Pavlov, Gabriel Goh, Scott Gray, Chelsea Voss, Alec Radford, Mark Chen, and Ilya Sutskever.
\newblock Zero-shot text-to-image generation.
\newblock In {\em ICML}. PMLR, 2021.

\bibitem{rastgoo2021sign}
Razieh Rastgoo, Kourosh Kiani, and Sergio Escalera.
\newblock Sign language recognition: A deep survey.
\newblock {\em Expert Systems with Applications}, 2021.

\bibitem{vqvae2}
Ali Razavi, Aaron van~den Oord, and Oriol Vinyals.
\newblock Generating diverse high-fidelity images with vq-vae-2.
\newblock In {\em NeurIPS}, 2019.

\bibitem{reed2016learning}
Scott Reed, Zeynep Akata, Honglak Lee, and Bernt Schiele.
\newblock Learning deep representations of fine-grained visual descriptions.
\newblock In {\em CVPR}, 2016.

\bibitem{reed2016generative}
Scott Reed, Zeynep Akata, Xinchen Yan, Lajanugen Logeswaran, Bernt Schiele, and Honglak Lee.
\newblock Generative adversarial text to image synthesis.
\newblock In {\em ICML}, pages 1060--1069. PMLR, 2016.

\bibitem{rix2001perceptual}
Antony~W Rix, John~G Beerends, Michael~P Hollier, and Andries~P Hekstra.
\newblock Perceptual evaluation of speech quality (pesq)-a new method for speech quality assessment of telephone networks and codecs.
\newblock In {\em ICASSP}, pages 749--752. IEEE, 2001.

\bibitem{Rombach_2022_CVPR}
Robin Rombach, Andreas Blattmann, Dominik Lorenz, Patrick Esser, and Bj\"orn Ommer.
\newblock High-resolution image synthesis with latent diffusion models.
\newblock In {\em CVPR}, 2022.

\bibitem{russakovsky2015imagenet}
Olga Russakovsky, Jia Deng, Hao Su, Jonathan Krause, Sanjeev Satheesh, Sean Ma, Zhiheng Huang, Andrej Karpathy, Aditya Khosla, Michael Bernstein, et~al.
\newblock Imagenet large scale visual recognition challenge.
\newblock {\em IJCV}, 2015.

\bibitem{salik2019lipper}
Khwaja~Mohd Salik, Swati Aggarwal, Yaman Kumar, Rajiv~Ratn Shah, Rohit Jain, and Roger Zimmermann.
\newblock Lipper: Speaker independent speech synthesis using multi-view lipreading.
\newblock In {\em AAAI}, pages 10023--10024, 2019.

\bibitem{salimans2016improved}
Tim Salimans, Ian Goodfellow, Wojciech Zaremba, Vicki Cheung, Alec Radford, and Xi Chen.
\newblock Improved techniques for training gans.
\newblock {\em NeurIPS}, 29, 2016.

\bibitem{scarselli2008gnns}
Franco Scarselli, Marco Gori, Ah~Chung Tsoi, Markus Hagenbuchner, and Gabriele Monfardini.
\newblock The graph neural network model.
\newblock {\em IEEE transactions on neural networks}, 2008.

\bibitem{schuhmann2022laion}
Christoph Schuhmann, Romain Beaumont, Richard Vencu, Cade Gordon, Ross Wightman, Mehdi Cherti, Theo Coombes, Aarush Katta, Clayton Mullis, Mitchell Wortsman, et~al.
\newblock Laion-5b: An open large-scale dataset for training next generation image-text models.
\newblock {\em arXiv:2210.08402}, 2022.

\bibitem{schuhmann2021laion}
Christoph Schuhmann, Richard Vencu, Romain Beaumont, Robert Kaczmarczyk, Clayton Mullis, Aarush Katta, Theo Coombes, Jenia Jitsev, and Aran Komatsuzaki.
\newblock Laion-400m: Open dataset of clip-filtered 400 million image-text pairs.
\newblock {\em arXiv:2111.02114}, 2021.

\bibitem{schwartz2019simple}
Idan Schwartz, Alexander~G Schwing, and Tamir Hazan.
\newblock A simple baseline for audio-visual scene-aware dialog.
\newblock In {\em CVPR}, 2019.

\bibitem{senocak2018learning}
Arda Senocak, Tae-Hyun Oh, Junsik Kim, Ming-Hsuan Yang, and In~So Kweon.
\newblock Learning to localize sound source in visual scenes.
\newblock In {\em CVPR}, pages 4358--4366, 2018.

\bibitem{senocak2022less}
Arda Senocak, Hyeonggon Ryu, Junsik Kim, and In~So Kweon.
\newblock Less can be more: Sound source localization with a classification model.
\newblock In {\em WACV}, pages 3308--3317, 2022.

\bibitem{seo2017visual}
Paul~Hongsuck Seo, Andreas Lehrmann, Bohyung Han, and Leonid Sigal.
\newblock Visual reference resolution using attention memory for visual dialog.
\newblock In {\em NeurIPS}, 2017.

\bibitem{sharma2018conceptual}
Piyush Sharma, Nan Ding, Sebastian Goodman, and Radu Soricut.
\newblock Conceptual captions: A cleaned, hypernymed, image alt-text dataset for automatic image captioning.
\newblock In {\em ACL}, 2018.

\bibitem{shih2016look}
Kevin~J Shih, Saurabh Singh, and Derek Hoiem.
\newblock Where to look: Focus regions for visual question answering.
\newblock In {\em CVPR}, 2016.

\bibitem{sigurdsson2020visual}
Gunnar~A Sigurdsson, Jean-Baptiste Alayrac, Aida Nematzadeh, Lucas Smaira, Mateusz Malinowski, Joao Carreira, Phil Blunsom, and Andrew Zisserman.
\newblock Visual grounding in video for unsupervised word translation.
\newblock In {\em CVPR}, pages 10850--10859, 2020.

\bibitem{sigurdsson2016hollywood}
Gunnar~A Sigurdsson, G{\"u}l Varol, Xiaolong Wang, Ali Farhadi, Ivan Laptev, and Abhinav Gupta.
\newblock Hollywood in homes: Crowdsourcing data collection for activity understanding.
\newblock In {\em ECCV}, 2016.

\bibitem{simonyan2014very}
Karen Simonyan and Andrew Zisserman.
\newblock Very deep convolutional networks for large-scale image recognition.
\newblock {\em arXiv preprint arXiv:1409.1556}, 2014.

\bibitem{sohl2015dpm_thermo}
Jascha Sohl-Dickstein, Eric Weiss, Niru Maheswaranathan, and Surya Ganguli.
\newblock Deep unsupervised learning using nonequilibrium thermodynamics.
\newblock In {\em ICML}. PMLR, 2015.

\bibitem{sohn2016improved}
Kihyuk Sohn.
\newblock Improved deep metric learning with multi-class n-pair loss objective.
\newblock {\em NeurIPS}, 29, 2016.

\bibitem{theo-diff4}
Yang Song, Conor Durkan, Iain Murray, and Stefano Ermon.
\newblock Maximum likelihood training of score-based diffusion models.
\newblock In {\em NeurIPS}, 2021.

\bibitem{theo-diff1}
Yang Song and Stefano Ermon.
\newblock Improved techniques for training score-based generative models.
\newblock In {\em NeurIPS}, 2020.

\bibitem{theo-diff2}
Yang Song, Jascha Sohl-Dickstein, Diederik~P Kingma, Abhishek Kumar, Stefano Ermon, and Ben Poole.
\newblock Score-based generative modeling through stochastic differential equations.
\newblock In {\em ICLR}, 2020.

\bibitem{song2018talking}
Yang Song, Jingwen Zhu, Dawei Li, Xiaolong Wang, and Hairong Qi.
\newblock Talking face generation by conditional recurrent adversarial network.
\newblock {\em arXiv:1804.04786}, 2018.

\bibitem{song2022self}
Zengjie Song, Yuxi Wang, Junsong Fan, Tieniu Tan, and Zhaoxiang Zhang.
\newblock Self-supervised predictive learning: A negative-free method for sound source localization in visual scenes.
\newblock In {\em CVPR}, pages 3222--3231, 2022.

\bibitem{spurr2018cross}
Adrian Spurr, Jie Song, Seonwook Park, and Otmar Hilliges.
\newblock Cross-modal deep variational hand pose estimation.
\newblock In {\em CVPR}, pages 89--98, 2018.

\bibitem{srinivasan2021wit}
Krishna Srinivasan, Karthik Raman, Jiecao Chen, Michael Bendersky, and Marc Najork.
\newblock Wit: Wikipedia-based image text dataset for multimodal multilingual machine learning.
\newblock In {\em Proceedings of the 44th International ACM SIGIR Conference on Research and Development in Information Retrieval}, 2021.

\bibitem{streiff20213d3l}
Dominc Streiff, Lukas Bernreiter, Florian Tschopp, Marius Fehr, and Roland Siegwart.
\newblock 3d3l: Deep learned 3d keypoint detection and description for lidars.
\newblock In {\em ICRA}. IEEE, 2021.

\bibitem{su2020audeo}
Kun Su, Xiulong Liu, and Eli Shlizerman.
\newblock Audeo: Audio generation for a silent performance video.
\newblock In {\em NeurIPS}, 2020.

\bibitem{su2021stvgbert}
Rui Su, Qian Yu, and Dong Xu.
\newblock Stvgbert: A visual-linguistic transformer based framework for spatio-temporal video grounding.
\newblock In {\em ICCV}, 2021.

\bibitem{sun2021discriminative}
Mingjie Sun, Jimin Xiao, Eng~Gee Lim, Si Liu, and John~Y Goulermas.
\newblock Discriminative triad matching and reconstruction for weakly referring expression grounding.
\newblock {\em IEEE TPAMI}, pages 4189--4195, 2021.

\bibitem{sun2018optical}
Shuyang Sun, Zhanghui Kuang, Lu Sheng, Wanli Ouyang, and Wei Zhang.
\newblock Optical flow guided feature: A fast and robust motion representation for video action recognition.
\newblock In {\em CVPR}, pages 1390--1399, 2018.

\bibitem{suwajanakorn2018discovery}
Supasorn Suwajanakorn, Noah Snavely, Jonathan~J Tompson, and Mohammad Norouzi.
\newblock Discovery of latent 3d keypoints via end-to-end geometric reasoning.
\newblock {\em NeurIPS}, 2018.

\bibitem{szegedy2016rethinking}
Christian Szegedy, Vincent Vanhoucke, Sergey Ioffe, Jon Shlens, and Zbigniew Wojna.
\newblock Rethinking the inception architecture for computer vision.
\newblock In {\em CVPR}, pages 2818--2826, 2016.

\bibitem{taal2010short}
Cees~H Taal, Richard~C Hendriks, Richard Heusdens, and Jesper Jensen.
\newblock A short-time objective intelligibility measure for time-frequency weighted noisy speech.
\newblock In {\em ICASSP}, pages 4214--4217. IEEE, 2010.

\bibitem{tan2022naturalspeech}
Xu Tan, Jiawei Chen, Haohe Liu, Jian Cong, Chen Zhang, Yanqing Liu, Xi Wang, Yichong Leng, Yuanhao Yi, Lei He, et~al.
\newblock Naturalspeech: End-to-end text to speech synthesis with human-level quality.
\newblock {\em arXiv:2205.04421}, 2022.

\bibitem{tang2022improved}
Zhicong Tang, Shuyang Gu, Jianmin Bao, Dong Chen, and Fang Wen.
\newblock Improved vector quantized diffusion models.
\newblock {\em arXiv:2205.16007}, 2022.

\bibitem{teed2020raft}
Zachary Teed and Jia Deng.
\newblock Raft: Recurrent all-pairs field transforms for optical flow.
\newblock In {\em ECCV}, pages 402--419. Springer, 2020.

\bibitem{thomas2022fine}
Christopher Thomas, Yipeng Zhang, and Shih-Fu Chang.
\newblock Fine-grained visual entailment.
\newblock In {\em ECCV}. Springer, 2022.

\bibitem{tian2020avvp}
Yapeng Tian, Dingzeyu Li, and Chenliang Xu.
\newblock Unified multisensory perception: Weakly-supervised audio-visual video parsing.
\newblock In {\em ECCV}, 2020.

\bibitem{tian2018audio}
Yapeng Tian, Jing Shi, Bochen Li, Zhiyao Duan, and Chenliang Xu.
\newblock Audio-visual event localization in unconstrained videos.
\newblock In {\em ECCV}, 2018.

\bibitem{truong2021right}
Thanh-Dat Truong, Chi~Nhan Duong, Hoang~Anh Pham, Bhiksha Raj, Ngan Le, Khoa Luu, et~al.
\newblock The right to talk: An audio-visual transformer approach.
\newblock In {\em ICCV}, 2021.

\bibitem{tsai2016video}
Yi-Hsuan Tsai, Ming-Hsuan Yang, and Michael~J Black.
\newblock Video segmentation via object flow.
\newblock In {\em CVPR}, 2016.

\bibitem{aist-dance-db}
Shuhei Tsuchida, Satoru Fukayama, Masahiro Hamasaki, and Masataka Goto.
\newblock Aist dance video database: Multi-genre, multi-dancer, and multi-camera database for dance information processing.
\newblock In {\em ISMIR}, 2019.

\bibitem{tulyakov2018mocogan}
Sergey Tulyakov, Ming-Yu Liu, Xiaodong Yang, and Jan Kautz.
\newblock Mocogan: Decomposing motion and content for video generation.
\newblock In {\em CVPR}, 2018.

\bibitem{unterthiner2019fvd}
Thomas Unterthiner, Sjoerd van Steenkiste, Karol Kurach, Rapha{\"e}l Marinier, Marcin Michalski, and Sylvain Gelly.
\newblock Fvd: A new metric for video generation.
\newblock 2019.

\bibitem{vaswani2017attention}
Ashish Vaswani, Noam Shazeer, Niki Parmar, Jakob Uszkoreit, Llion Jones, Aidan~N Gomez, {\L}ukasz Kaiser, and Illia Polosukhin.
\newblock Attention is all you need.
\newblock {\em NeurIPS}, 30, 2017.

\bibitem{vedantam2015cider}
Ramakrishna Vedantam, C Lawrence~Zitnick, and Devi Parikh.
\newblock Cider: Consensus-based image description evaluation.
\newblock In {\em CVPR}, 2015.

\bibitem{vinyals2015show}
Oriol Vinyals, Alexander Toshev, Samy Bengio, and Dumitru Erhan.
\newblock Show and tell: A neural image caption generator.
\newblock In {\em CVPR}, pages 3156--3164, 2015.

\bibitem{vougioukas2019video}
Konstantinos Vougioukas, Pingchuan Ma, Stavros Petridis, and Maja Pantic.
\newblock Video-driven speech reconstruction using generative adversarial networks.
\newblock {\em arXiv:1906.06301}, 2019.

\bibitem{vougioukas2020realistic}
Konstantinos Vougioukas, Stavros Petridis, and Maja Pantic.
\newblock Realistic speech-driven facial animation with gans.
\newblock {\em IJCV}, 128(5):1398--1413, 2020.

\bibitem{WahCUB_200_2011}
C. Wah, S. Branson, P. Welinder, P. Perona, and S. Belongie.
\newblock {The Caltech-UCSD Birds-200-2011 Dataset}.
\newblock Technical Report CNS-TR-2011-001, California Institute of Technology, 2011.

\bibitem{walker2015dense}
Jacob Walker, Abhinav Gupta, and Martial Hebert.
\newblock Dense optical flow prediction from a static image.
\newblock In {\em ICCV}, pages 2443--2451, 2015.

\bibitem{wallace2002histochemical}
Mark~N Wallace, Peter~W Johnston, and Alan~R Palmer.
\newblock Histochemical identification of cortical areas in the auditory region of the human brain.
\newblock {\em Experimental Brain Research}, 143(4):499--508, 2002.

\bibitem{wang2023overwriting}
Angelina Wang and Olga Russakovsky.
\newblock Overwriting pretrained bias with finetuning data.
\newblock In {\em ICCV}, 2023.

\bibitem{wang2018reconstruction}
Bairui Wang, Lin Ma, Wei Zhang, and Wei Liu.
\newblock Reconstruction network for video captioning.
\newblock In {\em CVPR}, 2018.

\bibitem{wang2017adversarial}
Bokun Wang, Yang Yang, Xing Xu, Alan Hanjalic, and Heng~Tao Shen.
\newblock Adversarial cross-modal retrieval.
\newblock In {\em ACMM}, pages 154--162, 2017.

\bibitem{wang2016comprehensive}
Kaiye Wang, Qiyue Yin, Wei Wang, Shu Wu, and Liang Wang.
\newblock A comprehensive survey on cross-modal retrieval.
\newblock {\em arXiv:1607.06215}, 2016.

\bibitem{wang2019hallucinating}
Lei Wang, Piotr Koniusz, and Du~Q Huynh.
\newblock Hallucinating idt descriptors and i3d optical flow features for action recognition with cnns.
\newblock In {\em ICCV}, pages 8698--8708, 2019.

\bibitem{wang2018learning}
Liwei Wang, Yin Li, Jing Huang, and Svetlana Lazebnik.
\newblock Learning two-branch neural networks for image-text matching tasks.
\newblock {\em IEEE TPAMI}, 41(2):394--407, 2018.

\bibitem{wang2017diverse}
Liwei Wang, Alexander Schwing, and Svetlana Lazebnik.
\newblock Diverse and accurate image description using a variational auto-encoder with an additive gaussian encoding space.
\newblock In {\em NeurIPS}, 2017.

\bibitem{wang2019neighbourhood}
Peng Wang, Qi Wu, Jiewei Cao, Chunhua Shen, Lianli Gao, and Anton van~den Hengel.
\newblock Neighbourhood watch: Referring expression comprehension via language-guided graph attention networks.
\newblock In {\em CVPR}, 2019.

\bibitem{wang2019language}
Weining Wang, Yan Huang, and Liang Wang.
\newblock Language-driven temporal activity localization: A semantic matching reinforcement learning model.
\newblock In {\em CVPR}, pages 334--343, 2019.

\bibitem{wang2019vatex}
Xin Wang, Jiawei Wu, Junkun Chen, Lei Li, Yuan-Fang Wang, and William~Yang Wang.
\newblock Vatex: A large-scale, high-quality multilingual dataset for video-and-language research.
\newblock In {\em ICCV}, 2019.

\bibitem{wang2021t2vlad}
Xiaohan Wang, Linchao Zhu, and Yi Yang.
\newblock T2vlad: global-local sequence alignment for text-video retrieval.
\newblock In {\em CVPR}, pages 5079--5088, 2021.

\bibitem{wang2019unos}
Yang Wang, Peng Wang, Zhenheng Yang, Chenxu Luo, Yi Yang, and Wei Xu.
\newblock Unos: Unified unsupervised optical-flow and stereo-depth estimation by watching videos.
\newblock In {\em CVPR}, pages 8071--8081, 2019.

\bibitem{wang2019camp}
Zihao Wang, Xihui Liu, Hongsheng Li, Lu Sheng, Junjie Yan, Xiaogang Wang, and Jing Shao.
\newblock Camp: Cross-modal adaptive message passing for text-image retrieval.
\newblock In {\em ICCV}, pages 5764--5773, 2019.

\bibitem{wang2022multi}
Zeyu Wang, Yu Wu, Karthik Narasimhan, and Olga Russakovsky.
\newblock Multi-query video retrieval.
\newblock {\em arXiv:2201.03639}, 2022.

\bibitem{wold1987principal}
Svante Wold, Kim Esbensen, and Paul Geladi.
\newblock Principal component analysis.
\newblock {\em Chemometrics and intelligent laboratory systems}, 2(1-3):37--52, 1987.

\bibitem{wu2018you}
Qi Wu, Peng Wang, Chunhua Shen, Ian Reid, and Anton Van Den~Hengel.
\newblock Are you talking to me? reasoned visual dialog generation through adversarial learning.
\newblock In {\em CVPR}, 2018.

\bibitem{wu2020jazz}
Shih-Lun Wu and Yi-Hsuan Yang.
\newblock The jazz transformer on the front line: Exploring the shortcomings of ai-composed music through quantitative measures.
\newblock {\em arXiv:2008.01307}, 2020.

\bibitem{wu2022snoc}
Yu Wu, Lu Jiang, and Yi Yang.
\newblock Switchable novel object captioner.
\newblock {\em IEEE TPAMI}, pages 1--1, 2022.

\bibitem{wu2021explore}
Yu Wu and Yi Yang.
\newblock Exploring heterogeneous clues for weakly-supervised audio-visual video parsing.
\newblock In {\em CVPR}, 2021.

\bibitem{wu2018decoupled}
Yu Wu, Linchao Zhu, Lu Jiang, and Yi Yang.
\newblock Decoupled novel object captioner.
\newblock In {\em ACMM}, 2018.

\bibitem{wu2019dual}
Yu Wu, Linchao Zhu, Yan Yan, and Yi Yang.
\newblock Dual attention matching for audio-visual event localization.
\newblock In {\em ICCV}, 2019.

\bibitem{wu2020comprehensive}
Zonghan Wu, Shirui Pan, Fengwen Chen, Guodong Long, Chengqi Zhang, and S~Yu Philip.
\newblock A comprehensive survey on graph neural networks.
\newblock {\em IEEE TNNLS}, 2020.

\bibitem{wu20153d}
Zhirong Wu, Shuran Song, Aditya Khosla, Fisher Yu, Linguang Zhang, Xiaoou Tang, and Jianxiong Xiao.
\newblock 3d shapenets: A deep representation for volumetric shapes.
\newblock In {\em CVPR}, 2015.

\bibitem{xiao2017weakly}
Fanyi Xiao, Leonid Sigal, and Yong Jae~Lee.
\newblock Weakly-supervised visual grounding of phrases with linguistic structures.
\newblock In {\em CVPR}, pages 5945--5954, 2017.

\bibitem{xiao2021next}
Junbin Xiao, Xindi Shang, Angela Yao, and Tat-Seng Chua.
\newblock Next-qa: Next phase of question-answering to explaining temporal actions.
\newblock In {\em CVPR}, 2021.

\bibitem{xie2018visual}
Ning Xie, Farley Lai, Derek Doran, and Asim Kadav.
\newblock Visual entailment task for visually-grounded language learning.
\newblock {\em arXiv preprint arXiv:1811.10582}, 2018.

\bibitem{xie2019visual}
Ning Xie, Farley Lai, Derek Doran, and Asim Kadav.
\newblock Visual entailment: A novel task for fine-grained image understanding.
\newblock {\em arXiv:1901.06706}, 2019.

\bibitem{xu2016ask}
Huijuan Xu and Kate Saenko.
\newblock Ask, attend and answer: Exploring question-guided spatial attention for visual question answering.
\newblock In {\em ECCV}. Springer, 2016.

\bibitem{xu2016msr}
Jun Xu, Tao Mei, Ting Yao, and Yong Rui.
\newblock Msr-vtt: A large video description dataset for bridging video and language.
\newblock In {\em CVPR}, pages 5288--5296, 2016.

\bibitem{xu2015show}
Kelvin Xu, Jimmy Ba, Ryan Kiros, Kyunghyun Cho, Aaron Courville, Ruslan Salakhudinov, Rich Zemel, and Yoshua Bengio.
\newblock Show, attend and tell: Neural image caption generation with visual attention.
\newblock In {\em ICML}, 2015.

\bibitem{xu2023multimodal}
Peng Xu, Xiatian Zhu, and David~A Clifton.
\newblock Multimodal learning with transformers: A survey.
\newblock {\em IEEE TPAMI}, 2023.

\bibitem{xu2018attngan}
Tao Xu, Pengchuan Zhang, Qiuyuan Huang, Han Zhang, Zhe Gan, Xiaolei Huang, and Xiaodong He.
\newblock Attngan: Fine-grained text to image generation with attentional generative adversarial networks.
\newblock In {\em CVPR}, 2018.

\bibitem{xuan2020cross}
Hanyu Xuan, Zhenyu Zhang, Shuo Chen, Jian Yang, and Yan Yan.
\newblock Cross-modal attention network for temporal inconsistent audio-visual event localization.
\newblock In {\em AAAI}, pages 279--286, 2020.

\bibitem{yadav2021speech}
Ravindra Yadav, Ashish Sardana, Vinay~P Namboodiri, and Rajesh~M Hegde.
\newblock Speech prediction in silent videos using variational autoencoders.
\newblock In {\em ICASSP}, pages 7048--7052. IEEE, 2021.

\bibitem{yang2022tubedetr}
Antoine Yang, Antoine Miech, Josef Sivic, Ivan Laptev, and Cordelia Schmid.
\newblock Tubedetr: Spatio-temporal video grounding with transformers.
\newblock In {\em CVPR}, 2022.

\bibitem{yang2022avqa}
Pinci Yang, Xin Wang, Xuguang Duan, Hong Chen, Runze Hou, Cong Jin, and Wenwu Zhu.
\newblock Avqa: A dataset for audio-visual question answering on videos.
\newblock In {\em ACMM}, 2022.

\bibitem{yang2019dynamic}
Sibei Yang, Guanbin Li, and Yizhou Yu.
\newblock Dynamic graph attention for referring expression comprehension.
\newblock In {\em ICCV}, 2019.

\bibitem{yang2017deep}
Xitong Yang, Palghat Ramesh, Radha Chitta, Sriganesh Madhvanath, Edgar~A Bernal, and Jiebo Luo.
\newblock Deep multimodal representation learning from temporal data.
\newblock In {\em CVPR}, pages 5447--5455, 2017.

\bibitem{yang2020improving}
Zhengyuan Yang, Tianlang Chen, Liwei Wang, and Jiebo Luo.
\newblock Improving one-stage visual grounding by recursive sub-query construction.
\newblock In {\em ECCV}, pages 387--404. Springer, 2020.

\bibitem{yang2019fast}
Zhengyuan Yang, Boqing Gong, Liwei Wang, Wenbing Huang, Dong Yu, and Jiebo Luo.
\newblock A fast and accurate one-stage approach to visual grounding.
\newblock In {\em ICCV}, pages 4683--4693, 2019.

\bibitem{yang2016stacked}
Zichao Yang, Xiaodong He, Jianfeng Gao, Li Deng, and Alex Smola.
\newblock Stacked attention networks for image question answering.
\newblock In {\em CVPR}, 2016.

\bibitem{you2016image}
Quanzeng You, Hailin Jin, Zhaowen Wang, Chen Fang, and Jiebo Luo.
\newblock Image captioning with semantic attention.
\newblock In {\em CVPR}, 2016.

\bibitem{yu2016automatic}
Dong Yu and Li Deng.
\newblock {\em Automatic speech recognition}, volume~1.
\newblock Springer, 2016.

\bibitem{yu2021mm}
Jiashuo Yu, Ying Cheng, Rui-Wei Zhao, Rui Feng, and Yuejie Zhang.
\newblock Mm-pyramid: Multimodal pyramid attentional network for audio-visual event localization and video parsing.
\newblock {\em arXiv:2111.12374}, 2021.

\bibitem{yu2016modeling}
Licheng Yu, Patrick Poirson, Shan Yang, Alexander~C Berg, and Tamara~L Berg.
\newblock Modeling context in referring expressions.
\newblock In {\em ECCV}, pages 69--85. Springer, 2016.

\bibitem{yu2018rethinking}
Zhou Yu, Jun Yu, Chenchao Xiang, Zhou Zhao, Qi Tian, and Dacheng Tao.
\newblock Rethinking diversified and discriminative proposal generation for visual grounding.
\newblock {\em arXiv:1805.03508}, 2018.

\bibitem{yuan2022explainability}
Hao Yuan, Haiyang Yu, Shurui Gui, and Shuiwang Ji.
\newblock Explainability in graph neural networks: A taxonomic survey.
\newblock {\em IEEE TPAMI}, 2022.

\bibitem{yuan2021multimodal}
Xin Yuan, Zhe Lin, Jason Kuen, Jianming Zhang, Yilin Wang, Michael Maire, Ajinkya Kale, and Baldo Faieta.
\newblock Multimodal contrastive training for visual representation learning.
\newblock In {\em CVPR}, pages 6995--7004, 2021.

\bibitem{yun2021pano}
Heeseung Yun, Youngjae Yu, Wonsuk Yang, Kangil Lee, and Gunhee Kim.
\newblock Pano-avqa: Grounded audio-visual question answering on 360deg videos.
\newblock In {\em ICCV}, 2021.

\bibitem{zhan2019spatial}
Fangneng Zhan, Hongyuan Zhu, and Shijian Lu.
\newblock Spatial fusion gan for image synthesis.
\newblock In {\em CVPR}, pages 3653--3662, 2019.

\bibitem{zhang2017stackgan}
Han Zhang, Tao Xu, Hongsheng Li, Shaoting Zhang, Xiaogang Wang, Xiaolei Huang, and Dimitris~N Metaxas.
\newblock Stackgan: Text to photo-realistic image synthesis with stacked generative adversarial networks.
\newblock In {\em CVPR}, 2017.

\bibitem{zhang2018stackgan++}
Han Zhang, Tao Xu, Hongsheng Li, Shaoting Zhang, Xiaogang Wang, Xiaolei Huang, and Dimitris~N Metaxas.
\newblock Stackgan++: Realistic image synthesis with stacked generative adversarial networks.
\newblock {\em IEEE TPAMI}, 2018.

\bibitem{zhang2021multi}
Mingxing Zhang, Yang Yang, Xinghan Chen, Yanli Ji, Xing Xu, Jingjing Li, and Heng~Tao Shen.
\newblock Multi-stage aggregated transformer network for temporal language localization in videos.
\newblock In {\em CVPR}, pages 12669--12678, 2021.

\bibitem{zhang2021flow}
Zhimeng Zhang, Lincheng Li, Yu Ding, and Changjie Fan.
\newblock Flow-guided one-shot talking face generation with a high-resolution audio-visual dataset.
\newblock In {\em CVPR}, pages 3661--3670, 2021.

\bibitem{zhang2020does}
Zhu Zhang, Zhou Zhao, Yang Zhao, Qi Wang, Huasheng Liu, and Lianli Gao.
\newblock Where does it exist: Spatio-temporal video grounding for multi-form sentences.
\newblock In {\em CVPR}, 2020.

\bibitem{zhao2019sound}
Hang Zhao, Chuang Gan, Wei-Chiu Ma, and Antonio Torralba.
\newblock The sound of motions.
\newblock In {\em ICCV}, 2019.

\bibitem{zhao2018sound}
Hang Zhao, Chuang Gan, Andrew Rouditchenko, Carl Vondrick, Josh McDermott, and Antonio Torralba.
\newblock The sound of pixels.
\newblock In {\em ECCV}, 2018.

\bibitem{zhao2019object}
Zhong-Qiu Zhao, Peng Zheng, Shou-tao Xu, and Xindong Wu.
\newblock Object detection with deep learning: A review.
\newblock {\em IEEE TNNLS}, 30(11):3212--3232, 2019.

\bibitem{zhen2019deep}
Liangli Zhen, Peng Hu, Xu Wang, and Dezhong Peng.
\newblock Deep supervised cross-modal retrieval.
\newblock In {\em CVPR}, pages 10394--10403, 2019.

\bibitem{zhou2021pose}
Hang Zhou, Yasheng Sun, Wayne Wu, Chen~Change Loy, Xiaogang Wang, and Ziwei Liu.
\newblock Pose-controllable talking face generation by implicitly modularized audio-visual representation.
\newblock In {\em CVPR}, pages 4176--4186, 2021.

\bibitem{zhou2019grounded}
Luowei Zhou, Yannis Kalantidis, Xinlei Chen, Jason~J Corso, and Marcus Rohrbach.
\newblock Grounded video description.
\newblock In {\em CVPR}, 2019.

\bibitem{zhou2018visual}
Yipin Zhou, Zhaowen Wang, Chen Fang, Trung Bui, and Tamara~L Berg.
\newblock Visual to sound: Generating natural sound for videos in the wild.
\newblock In {\em CVPR}, pages 3550--3558, 2018.

\bibitem{zhu2020domain}
Jiapeng Zhu, Yujun Shen, Deli Zhao, and Bolei Zhou.
\newblock In-domain gan inversion for real image editing.
\newblock In {\em ECCV}. Springer, 2020.

\bibitem{zhu2019dm}
Minfeng Zhu, Pingbo Pan, Wei Chen, and Yi Yang.
\newblock Dm-gan: Dynamic memory generative adversarial networks for text-to-image synthesis.
\newblock In {\em CVPR}, 2019.

\bibitem{yezhu2022quantizedgan}
Ye Zhu, Kyle Olszewski, Yu Wu, Panos Achlioptas, Menglei Chai, Yan Yan, and Sergey Tulyakov.
\newblock Quantized gan for complex music generation from dance videos.
\newblock In {\em ECCV}, 2022.

\bibitem{zhu2023boundary}
Ye Zhu, Yu Wu, Zhiwei Deng, Olga Russakovsky, and Yan Yan.
\newblock Boundary guided learning-free semantic control with diffusion models.
\newblock {\em NeurIPS}, 2023.

\bibitem{zhu2021learning}
Ye Zhu, Yu Wu, Hugo Latapie, Yi Yang, and Yan Yan.
\newblock Learning audio-visual correlations from variational cross-modal generation.
\newblock In {\em ICCASP}, 2021.

\bibitem{yezhu2022cdcd}
Ye Zhu, Yu Wu, Kyle Olszewski, Jian Ren, Sergey Tulyakov, and Yan Yan.
\newblock Discrete contrastive diffusion for cross-modal and conditional generation.
\newblock In {\em ICLR}, 2023.

\bibitem{zhu2020describing}
Ye Zhu, Yu Wu, Yi Yang, and Yan Yan.
\newblock Describing unseen videos via multi-modal cooperative dialog agents.
\newblock In {\em ECCV}, 2020.

\bibitem{zhu2021saying}
Ye Zhu, Yu Wu, Yi Yang, and Yan Yan.
\newblock Saying the unseen: Video descriptions via dialog agents.
\newblock In {\em TPAMI}, 2021.

\end{thebibliography}
}

\vspace{-2.0in}
\begin{IEEEbiography}[{\includegraphics[width=0.9in,height=1in,clip,keepaspectratio]{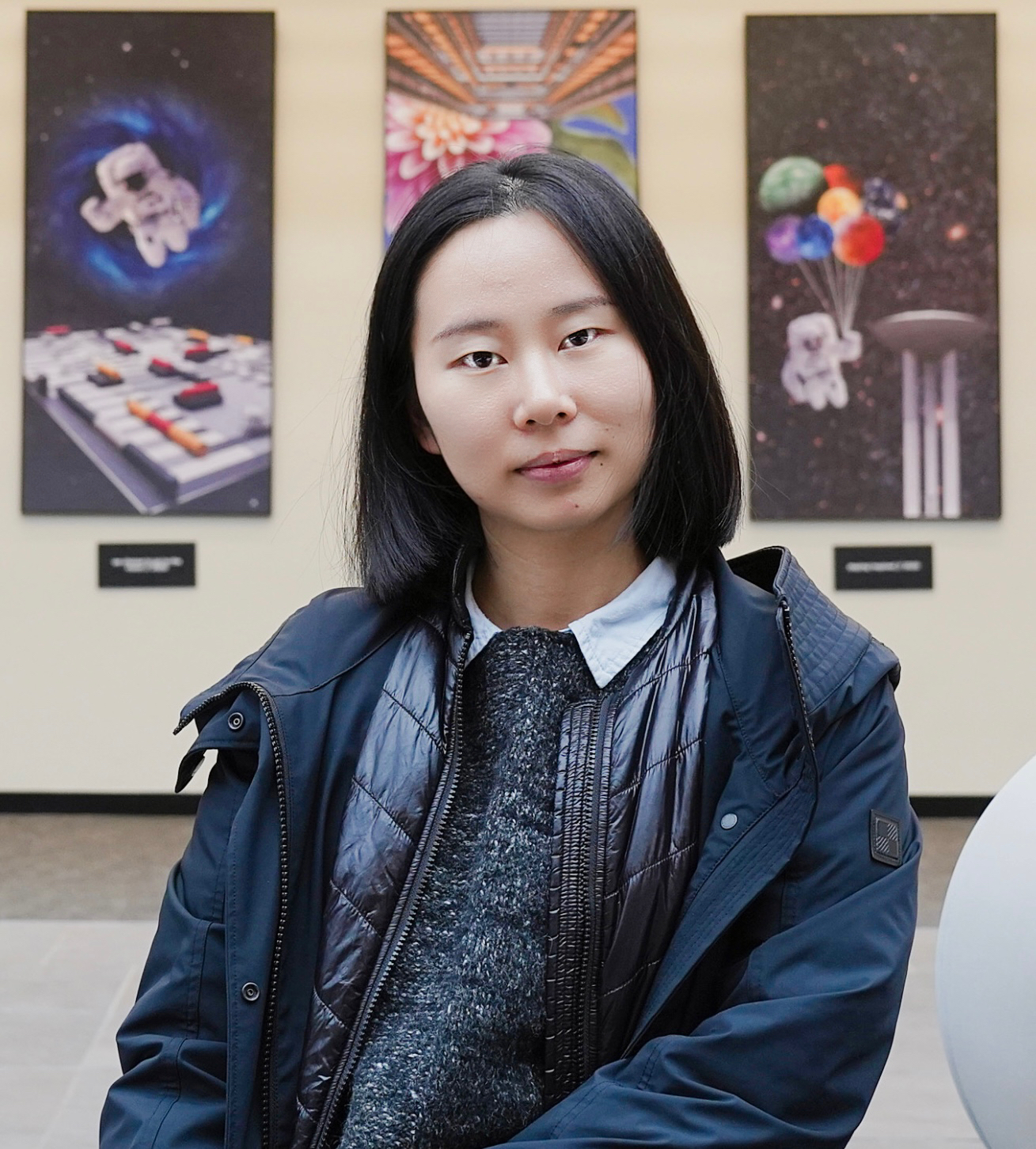}}]{Ye Zhu}
is currently a postdoctoral researcher in Computer Science at Princeton University, USA.
She holds the Ph.D. degree in Computer Science from Illinois Institute of Technology, USA, in 2023.
She received her B.S. and M.S. degrees from Shanghai Jiao Tong University, China. Her research interests mainly include multimodal learning and generation.
\end{IEEEbiography}
\vspace{-2.2in}
\begin{IEEEbiography}[{\includegraphics[width=0.9in,height=1in,clip,keepaspectratio]{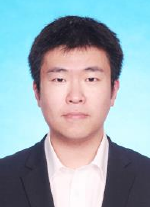}}]{Yu Wu}
is professor with the School of Computer Science at Wuhan University, China. 
His research interests are multi-modal perception and video understanding.
He was the recipient of Google PhD Fellowship 2020. He was the Workshop Chair of CVPR 2023, and also served as Area Chair for CVPR 2023, NeurIPs 2023.
\end{IEEEbiography}
\vspace{-2.2in}
\begin{IEEEbiography}[{\includegraphics[width=0.9in,height=1in,clip,keepaspectratio]{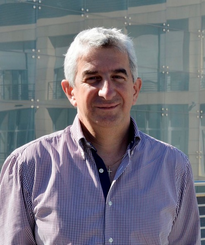}}]{Nicu Sebe} is professor with the University of Trento, Italy. His research covers the areas of multimedia information retrieval and human behavior understanding. He was the Program Chair of the International Conference on
Image and Video Retrieval in 2007 and 2010, ACM Multimedia 2007 and 2011. He was the Program Chair of ICCV 2017 and ECCV 2016, and a General Chair of ACM ICMR 2017. 
\end{IEEEbiography}
\vspace{-2.2in}
\begin{IEEEbiography}[{\includegraphics[width=0.9in,height=1in,clip,keepaspectratio]{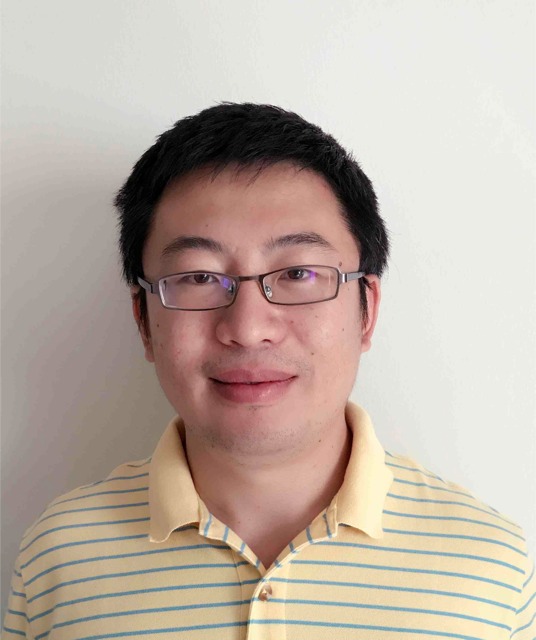}}]{Yan Yan} is currently Gladwin Development Chair Assistant Professor with the Department of Computer Science, Illinois Institute of Technology. He received the Ph.D. degree in computer science from the University of Trento. 
His research interests include computer vision, machine learning, and multimedia.
\end{IEEEbiography}

\vspace{-1.8in}

\clearpage


\vspace{-0.15in}
\appendices


\section{Multimodal Datasets}
\label{sec:dataset}

Datasets play an essential role in deep learning research but in the context of multimodal learning, the datasets usually require data and annotations that cover several modalities.
In this section, we summarize different multimodal datasets, categorized by their real-world or synthetic natures, and introduce their basic information including the data scale, annotations, and applicable tasks.
Table~\ref{tab:data} lists the overall information for the presented multimodal datasets.

Here we present the multimodal datasets \ye{listed in Table~\ref{tab:data} in the main paper}.

\subsection{Real-World Dataset}
\label{subsec:real_data}

The real-world data are essential for multimodal networks to be applied in practical scenarios.

\noindent \textbf{MSCOCO} dataset~\cite{lin2014microsoft}
is a large-scale benchmark dataset widely used for various computer vision research problems such as object detection and segmentation. It includes 330k images, among which over 200k of them are labeled. 
In terms of the multimodal attributes, MSCOCO includes the textual captions for each image and therefore can be applied to vision and text related tasks such as image captioning and text-to-image synthesis. 

\noindent \textbf{Conceptual Captions} (CC) dataset~\cite{sharma2018conceptual,changpinyo2021conceptual} provides image-caption pairs extracted from Internet. The training split of the CC dataset has more than 3M images with a token vocabulary size of 51k. Each caption has an average length 10.3 words (tokens). The validation and testing splits include more than 15k images.
As one of the largest open access image-caption datasets, the CC dataset is often used for large-scale model pretraining. 
Note that several large-scale pre-trained models (\textit{e.g.}, CLIP~\cite{radford2021clip}) do not release their training dataset.

\noindent \ye{\textbf{LAION} datasets~\cite{schuhmann2021laion,schuhmann2022laion} are recent open large-scale datasets designed for vision and language representation learning and generation. Released Laion datasets have two version of LAION-400M~\cite{schuhmann2021laion} and LAION-5B~\cite{schuhmann2022laion}. The former one includes 400 million image-text pairs used in CLIP training~\cite{radford2021clip}, and the later one features a larger collection of filtered image-text pairs that have been used for recent large text-to-image generation models such as GLIDE~\cite{nichol2021glide} and StableDiffusion~\cite{Rombach_2022_CVPR}.
}

\noindent \textbf{CUB200} dataset~\cite{WahCUB_200_2011} contains 11,788 images of 200 bird species. The training and testing splits include 5,994 and 5,794 images respectively. The textual descriptions of each bird image are later annotated by~\cite{reed2016learning}. Compared to the previous large vision and text datasets such as MSCOCO and CC, CUB200 focus on a very specific and fine-grained category of visual information.   

\noindent \textbf{VisDial} dataset~\cite{das2017visual} contains images and the related question answer dialog annotations, which are designed for machines to maintain a meaningful dialog given the visual information as input. The 120k images of VisDial come from MSCOCO and each has a corresponding dialog annotation that consist of 10 question and answer pairs, resulting in 1.2M rounds of dialog interactions. It is a benchmark dataset used for vision and text related tasks such as visual dialog~\cite{das2017visual}.

\noindent \textbf{GuessWhat?!} is a large-scale dataset consisting of 800k visual question and answer pairs on 66k images in total~\cite{de2017guesswhat}.
Specifically, the data are collected from a cooperative two-player game named \textit{GuessWhat?!}, in which both players see the image. 
One player of the game - the oracle - is assigned a random object from the given image, while the other player - the questioner - does not know the target object assigned. The game objective is for the questioner to locate the target object via dialog interactions with the oracle. 
This is a dataset usually applicable in the visual dialog task.

\noindent \ye{\textbf{VQA} dataset~\cite{antol2015vqa} is initially a benchmark dataset that contains real world images with their corresponding open-ended questions. VQA 2.0~\cite{goyal2017making} enriches the original version with 10 ground truth and 3 plausible answers to questions of 265,016 images, providing a more balanced VQA dataset for studying the VQA task.
} 

\noindent \ye{\textbf{OK-VQA} dataset~\cite{marino2019ok}, in which the \emph{OK} stands for Outside Knowledge, is a benchmark dataset for a specific VQA task setting that requires external knowledge to answer. While the answers to the raised questions in classic VQA task can be inferred purely based on the given images, this dataset is designed for the scenario where the answers should be provided based on both image content and some external knowledge.
}

\noindent \textbf{NExT-QA}~\cite{xiao2021next} is a benchmark dataset with videos and question-answer dialog. The main difference of this video-text dataset lies within the fact that the NExT-QA is claimed to host causal and temporal action reasoning in the video question answering process.
It includes a total of 5440 videos with an average length of 44 seconds. For each video, the dataset provides 10 rounds of question and answer pairs as the dialog annotations, resulting in a total of 52k manually labeled question and answer pairs. 
The question and answer pairs are further grouped into clusters with causal, temporal, or descriptive attributes. 

\begin{table*}[ht]
    \centering
    \scalebox{0.95}{
    \begin{tabular}{ccccccc}
    \hline
        Dataset & Nature & Vision & Audio & Text & Others & Typical Applications  \\ \hline
        MSCOCO~\cite{lin2014microsoft} & R & image & - & caption & body keypoints &  image captioning 
        /text-to-image synthesis\\
        CC~\cite{sharma2018conceptual,changpinyo2021conceptual} & R & image & - & caption & -  & image-text \\
        LAION~\cite{schuhmann2021laion,schuhmann2022laion} & R & image & - & caption & - & image-text \\
        CUB200~\cite{WahCUB_200_2011} & R & image & - & caption & -  & text-to-image synthesis \\
        VisDial~\cite{davis2018visual} & R & image & - & dialog & - & visual dialog \\
        GuessWhat~\cite{de2017guesswhat}& R & image & - & dialog & - & visual dialog\\
        VQA~\cite{antol2015vqa,goyal2017making} & R & images & - & questions & -  & VQA \\
        OK-VQA~\cite{marino2019ok} & R & images & - & dialog & - & VQA\\
        
        NExT-QA~\cite{xiao2021next} & R & video & - & dialog & - & VQA \\
        MSVD~\cite{chen2011collecting} & R & video & - & caption & - & video captioning \\
        MSR-VTT~\cite{xu2016msr} & R & video & - & caption & - & video captioning \\ 
        Vatex~\cite{wang2019vatex} & R & video & - & caption & - & video captioning\\
        ActivityNet Captions~\cite{krishna2017dense} & R & video & - & caption & - & video captioning \\
        \re{Visual Genome~\cite{krishna2017visualgenome}} & \re{R} & \re{images} & - & \re{caption} & \re{graph} &  \re{VQA\&captioning\&grounding} \\
        \re{WIT~\cite{srinivasan2021wit}} & \re{R} &  \re{images} & - & \re{multilingual texts} & - & \re{multimodal retrieval} \\
        Crema-d~\cite{cao2014crema} & R & video & vocal audio & - & - & emotion perception \\
        LipRead~\cite{chung2017lip} & R & video & vocal audio & - & - & speech recognition\\
        \re{ASL Alphabet~\cite{akash_nagaraj_2018}} & \re{R} & \re{images} & - & \re{sign language} & - &  \re{sign language recognition\&generation}\\
        \re{How2ign~\cite{duarte2021how2sign}} & \re{R} & \re{videos} & - & \re{sign language} & - & \re{sign language recognition\&generation} \\
        AVSD~\cite{hori2019end,alamri2019audio} & R & video & ambient & dialog & - & video captioning and visual dialog \\
        AVE~\cite{tian2018audio} & R & video & ambient & - & & audio-visual event localization\\
        \re{EPIC-SOUNDS~\cite{huh2023epic}} & \re{R} & \re{videos} & \re{ambient} & - & - & \re{audio-visual action recognition}\\
        AIST~\cite{aist-dance-db} & R & video &  music & - & - & dance-music\\
        AIST++~\cite{aistplus} & R & video & music & - & 2D/3D/SMPL  & dance-music \\
        TikTok Dance-Music~\cite{yezhu2022quantizedgan}& R & video & music & - & 2D & dance-music \\ 
        MUSIC~\cite{zhao2018sound,zhao2019sound} & R & video & music & - & - &  video-music\\
        FACC~\cite{harwath2015deep} & R & image & speech & caption & - & sanity check for multimodal learning\\
        LLP~\cite{tian2020avvp} & R & video & ambient & - & - & audio-visual video parsing \\
        CLEVR~\cite{johnson2017clevr} & S & image & - & dialog & scene graph &  visual question answering\\
        MUGEN~\cite{hayes2022mugen} & S & video & music & caption & -  & video-music-text \\
        ModelNet~\cite{wu20153d} & S & image & - & - & point clouds/meshes & 3D related tasks \\
        ShapeNet~\cite{chang2015shapenet} & S & image & - &  - & point clouds/meshes & 3D related tasks\\
        \hline
    \end{tabular}}
    \caption{\textbf{Overview of various multimodal datasets.} \emph{R} and \emph{S} indicates the real-world and synthetic natures, respectively.  We only list the typical applications related to multi- and cross-modality problems in this table. Some large-scale benchmark dataset such as MSCOCO are also widely used for other tasks such as object segmentation and detection. \ye{Detailed introduction for the above datasets can be found in Appendix~\ref{sec:dataset}.}
    }
    \label{tab:data}
\end{table*}

\noindent \textbf{MSVD} (Microsoft Research Video Description)~\cite{chen2011collecting} is a dataset consists of about 120K English descriptions for more than 2,000 video clips. It is a dataset first collected and proposed for the machine translation usage, but later also widely used for the video captioning task.

\noindent \textbf{MSR-VTT} (MSR-Video to Text)~\cite{xu2016msr} is a large-sclae video benchmark for video understanding. It contains in total 10,000 web video clips with 41.2 hours and 2000K clip-sentence pairs. Each video clip is annotated with 20 natural sentences. 
MSR-VTT dataset is collected based on 257 popular queries, covering 20 representative categories.
It is mostly adopted in video captioning task.

\noindent \ye{\textbf{Vatex} dataset~\cite{wang2019vatex} features a large-scale multilingual dataset for video captioning. It contains 41,250 videos and 825,000 captions in English and Chinese, with over 206,000 English-Chinese translation pairs. Similar to MSR-VTT~\cite{xu2016msr}, this dataset is proposed for video captioning in the context of multimodal learning.
}

\noindent \ye{\textbf{ActivityNet Capations} dataset~\cite{krishna2017dense,zhou2019grounded} includes 20k untrimmed YouTube videos of ActivityNet~\cite{caba2015activitynet} with 100k caption annotations. With an average video length of 120s, the videos from ActivityNet Captions usually have over 3 recognized activities, and each of those activities is annotated by a caption description with an average of 13.5 words. This dataset can be used for various video and language related tasks including video captioning and grounding. 
}

\noindent \re{\textbf{Video Genome}~\cite{krishna2017visualgenome} is a realworld dataset that contains 108K images annotated with objects, attributes, and pairwise relationships between objects in the form of graphs.
For the language side, it integrates 5.4M region descriptions and 1.7M visual question answers.
It can be adopted for multimodal tasks such as visual question answering and grounding.
}

\noindent \re{\textbf{WIT} dataset~\cite{srinivasan2021wit} is a recent Wikipedia-based image text dataset for multimodal multilingual learning. It covers 37.5M image-text examples with 11.5M unique images across 108 languages.
It is mainly proposed for multimodal retrieval task.
}

\noindent \ye{\textbf{Crema-d} is an early audio-visual dataset~\cite{cao2014crema} designed for multimodal emotion expression and perception. Specifically, this dataset includes facial and vocal emotional expressions in sentences spoken with different emotional states (\textit{e.g.}, happy, sad, and anger). 
}

\noindent \textbf{LipRead} dataset introduced in~\cite{chung2017lip} is proposed for words/speech recognition given talking faces. For the visual information, the videos of talking faces are collected from TV broadcasts, and then annotated with over a million word instances.

\noindent \re{\textbf{ASL Alphabet}~\cite{akash_nagaraj_2018} is a dataset for sign language, which collects images of alphabets from the American Sign Language. It is grouped into 29 classes, of which 26 alphabets from A to Z and 3 other classes of SPACE, DELETE, NOTHING.
It is a dataset specially designed for sign language-related applications.
}

\noindent \re{\textbf{How2sign}~\cite{duarte2021how2sign} is another dataset for sign language-related tasks. It is a multimodal and multiview continuous ASL dataset.
How2sign includes more than 80 hours of sign language videos and a set of corresponding modalities including speech, English transcripts, and depth.
}

\noindent \textbf{AVSD} (Audio Visual Scene Aware Dialog)~\cite{hori2019end,alamri2019audio} is a dataset that includes vision, audio and text data attributes. The visual information includes videos from the Charades dataset~\cite{sigurdsson2016hollywood}, and the text data include both caption and dialog annotations. For each dialog of a given video, there are ten rounds of question-answer pairs. This dataset can be applied in both video captioning~\cite{zhu2020describing,zhu2021saying} or video-based question answering~\cite{schwartz2019simple}.

\noindent \textbf{AVE.} (Audio Visual Event) dataset~\cite{tian2018audio} contains 4143 unconstrained 10s videos covering 28 event categories such as dog barking, frying food, and playing guitar.
Most videos contain visual events that span over the full 10 seconds, and the visual event within the video is accompanied by its corresponding ambient sound.
This dataset is proposed to study the audio-visual event localization problems~\cite{tian2018audio,wu2019dual,zhu2021learning}.

\noindent \re{\textbf{EPIC-SOUNDS}~\cite{huh2023epic}
builds on top of egocentric videos from EPIC-KITCHENS-100~\cite{damen2018scaling}. It collects audio annocations capturing temporal extents and class labels within the audio stream, including 75.9k segments audible events and actions across 44 classes. This dataset can be used for audio-visual related tasks such as audio-visual localization and recognition.
}

\noindent \textbf{AIST}~\cite{aist-dance-db} is a dataset with professional dance videos with paired dance music, which is also one of the first large-scale datasets for street dances. It contains 13,940 videos with 1,618 dances, accompanied by 60 music pieces. The total video length is about 118.1 hours. The dances in the AISTcover 10 different genres such as break, pop, house, and street jazz. 
The dance choreographers are performed by professional dancers and the videos are recorded in a studio environment from 9 camera poses.
This dataset was initially collected for the purpose of promoting dance information processing, but has been later utilized in more cross-modal scenarios related to dance and music.

Additionally, \textbf{AIST++}~\cite{aistplus} is a subset of AIST with more refined dance motion annotations. Specifically, this dataset provides motion annotations in form of Skinkked Multi-Person Linear (SPML) representations~\cite{SMPL:2015}, 3D human body keypoints, and 2D human keypoints, proposed for the music-to-dance generation task~\cite{aistplus} that seek to generate dance motions from given musical audio input.  

\noindent \textbf{TikTok Dance-Music}~\cite{yezhu2022quantizedgan} includes dance videos with paired music collected from the TikTok social platform. This dataset includes 445 video clips with an average length of 12.3 seconds with 85 different songs.
Similar to the AIST++~\cite{aistplus}, it also contains the motion attributes, represented by the 2D human skeleton data.
As another dataset for the dance and music cross-modality generation task, the major difference between AIST++ and TikTok Dance-Music lies within the fact that the former has clearer settings, while the latter is rather ``in the wild", and therefore also more challenging when utilized in concrete dance and music related tasks. 

\noindent \textbf{MUSIC}~\cite{zhao2018sound,zhao2019sound} is a dedicated dataset for music instrumental performance videos. It includes 685 untrimmed videos of musical solos and duets with 11 instrumental types such as guitar, cello, and violin. For each video, there is also the corresponding music sound. 
The dataset has been initially proposed for visual-audio grounding problems, such as the spatial sound localization within the videos~\cite{zhao2018sound}.

\noindent \textbf{Flicker 8k} has 8000 natural images with caption descriptions. Each sentence has 5 textual descriptions, resulting in 40000 captions in total. 
The \textbf{Flicker Audio Caption Corpus}~\cite{harwath2015deep} provides the speech audio of the 40000 spoken captions.
As a small dataset with vision, audio and text annotations, it is often used for preliminary experiments for sanity checks in multimodal learning related works.   

\noindent \textbf{Look, Listen, and Parse} (LLP)~\cite{tian2020avvp} is a subset of AudioSet~\cite{gemmeke2017audioset}, which contains 11,849 YouTube video clips categorized into 25 event classes. Each video is 10s length, with a corresponding ambient sound that characterizes at least one specific event such as vehicle sounds or human activities. There are in total 6626 event annotations, among which 4131 for audio and 2495 for the visual, paired with second-wise temporal boundaries. Similar to AVE~\cite{tian2018audio}, this dataset was originally proposed for audio-visual tasks such as audio-visual video parsing (AVVP) (see details in Sec.~\ref{sec:discriminative}).

\vspace{-0.1in}
\subsection{Synthetic Dataset}
\label{subsec:syn_data}

Compared to real-world datasets, synthetic datasets usually include generated data with less noise and are often used for the diagnostic purposes of analyzing the reasoning abilities of model networks. 

\noindent \textbf{CLEVR}~\cite{johnson2017clevr} is a diagnostic synthetic dataset for visual reasoning and compositional language parsing. It contains images with geometrical shapes of cubes, cylinders, and spheres in different colors and materials. The main dataset includes 70k, 15k, and 15k images for the training, validation, and test sets, respectively. In addition to the images, CLVER also provides question and answer annotations.

\noindent \textbf{MUGEN}~\cite{hayes2022mugen} is a recently proposed large-scale synthetic dataset for multimodal studies with vision, audio and text data attributes. The data of MUGEN are collected from the open-source platform game CoinRun~\cite{cobbe2019quantifying}, which includes 375k video clips of 3.2 seconds, paired music sequences and corresponding human-annotated text descriptions.

\noindent 
\ye{\textbf{ModelNet}~\cite{wu20153d} is a synthetic 3D dataset with multiple 3D annotations in the form of point clouds and CAD-generated meshes. It contains two variants, ModelNet10 and ModelNet40, with different numbers of object categories. It is a widely used benchmark dataset for 3D related multimodal tasks.}

\noindent 
\ye{\textbf{ShapeNet}~\cite{chang2015shapenet} features another large-scale 3D repository for CAD models categorized into 16 common object classes. It contains over 300M individual meshes with 220,000 classified into 3,135 classes annotated using WordNet hypernym-hyponym relationships, and is also widely used in 3D research field.
}

\section{Evaluation for Synthesized Data}
\label{app_sec:evaluation_generation}

The evaluations for the generative tasks have always been an important aspect to consider. 
We can categorize the existing evaluations metrics based on different taxonomy. 
In this survey, we focus on two criteria for classifying those metrics: whether a metric is subjective or objective, or if a metric is used to measure the general quality of the data or the faithfulness between the generated data and the given input.
For the first criterion, subjective evaluations often involve human users to assess and rate the quality, while the objective metrics are usually automatically computed without the intervention from humans.
As for the second criterion, while the evaluation for general quality has been widely considered for unconditional generative setting, the faithfulness is especially important in case of cross-modality generation since we want to synthesize data that well corresponds to the input data (\textit{e.g.}, the generated ambient sound is expected to match the scenario from the video). 
We summarize some commonly adopted metrics for different synthesized data and their features in Table~\ref{tab:metrics}.

\begin{table*}[t]
    \centering
    \scalebox{1.1}{
    \begin{tabular}{ccccc}
    \toprule
    Data   & Metrics &  Gen./Cross I-O & Obj./ Sub. & Measurements\\ \hline
    Image   & IS~\cite{salimans2016improved} &  Gen. & Obj. & quality\&diversity \\
    Image & FID~\cite{heusel2017gans} & Gen. & Obj. & quality \\
    Video & FVD~\cite{unterthiner2019fvd} & Gen. & Obj. & quality \\
    Image\&Texts & ClipScore~\cite{hessel2021clipscore} & Cross I-O & Obj. & text-image correspondence \\
    Image\&Video & Classification Acc.~\cite{liu2022compositional} & Cross I-O & Obj. & label correspondence \\
    General audio & SNR~\cite{plapous2006improvedsnr} & Gen. & Obj. & signal-to-noise ratio \\
    General audio & MOS~\cite{zhu2019dm} & Gen.\& Cross I-O & Sub. & crafted instruction \\
    Music & Beat Scores~\cite{zhu2019dm} & Cross I-O & Obj. & beats correspondence\\
    Music & Genre Accuracy~\cite{zhu2019dm} & Cross I-O & Obj. & genre diversity \\
    Music & Pitch Class Histogram Entropy~\cite{wu2020jazz} & Gen. & Obj. & music tonality \\
    Music & Grooving Pattern Similarity~\cite{wu2020jazz} & Gen. & Obj& music rhythm \\
    Music & Structureness Indicator~\cite{wu2020jazz} & Gen. & Obj. & music repetitive structure \\
    Speech & STOI~\cite{taal2010short} & Gen. & Obj. & speech audio intelligibility\\
    Speech & ESTOI~\cite{jensen2016algorithm} & Gen. & Obj. & speech audio intelligibility \\
    Speech & PESQ~\cite{rix2001perceptual}& Gen. & Obj. & perception evaluation of speech quality\\
    Speech & WER~\cite{rix2001perceptual} & Cross I-O & Obj. & speech-word accuracy\\
    Text & BLEU~\cite{papineni2002bleu} & Gen. & Obj. & text similarity \\
    Text & METEOR~\cite{banerjee2005meteor} & Gen. & Obj. & text. similarity  \\
    Text & ROUGE~\cite{lin2004rouge} & Gen. & Obj. & text similarity \\
    Text & CIDEr~\cite{vedantam2015cider} & Cross I-O & Obj. & text similarity\\
    Text & SPICE~\cite{anderson2016spice} & Cross I-O & Obj. & text-image semantic similarity \\
    \bottomrule
    \end{tabular}}
    \caption{\textbf{Common evaluation metrics for different synthesized data in generative tasks.} \textit{Gen./Cross I-O} means whether the metric is used for evaluating the general quality of the synthesized data or the input-output correspondence for the cross-modality scenarios. \textit{Obj./Sub.} stands for the metric is either objective or subjective evaluation.}
    \label{tab:metrics}
    \vspace{-0.2in}
\end{table*}

\subsection{Visual Data Evaluation}
To measure the general quality of generated vision data including images and videos, the most commonly used objective metrics include the Inception Scores (IS)~\cite{salimans2016improved} and Fréchet Inception Distance (FID)~\cite{heusel2017gans}. 
The IS measures the general reality and diversity for the generated images using a pre-trained Inception v3 image classification model~\cite{szegedy2016rethinking}, where also originated the name of ``Inception Score". 
Specifically, the idea of IS relies on two premises: 
On the one hand, for a synthesized image $x$ to be realistic, it should be assigned to a certain image class discriminatively with a high conditional probability, and thus a conditional label distribution $p(y|x)$ should have a low entropy with $y$ as the class label. 
On the other hand, while measuring the diversity of generated images, IS considers the marginal probabilities of all the labels to be widely spread over all class labels, therefore leading to a high entropy. 
IS combines both above criteria while calculating the quantitative scores.
A higher IS indicates better quality for the generative results.
The FID score is another metric widely used to assess the general quality of synthesized images.
It calculates the distance on the feature vector levels, by first computing the image feature vectors using the Inception models~\cite{szegedy2016rethinking} from both original and generated images. 
A lower FID score implies better quality of the generated images with higher similarity to real images.
Similar to the FID for image evaluations, FVD (Fréchet Video Distance)~\cite{unterthiner2019fvd} is designed to measure the fidelity of generated videos.

In addition to the above objective metrics, subjective evaluations conducted by human users are also used in literature to judge the general quality of the synthesized visual data. However, unlike the objective metrics such as IS and FID, subjective metrics are rather hand crafted case by case. 
For example, DALLE-2~\cite{ramesh2022dalle2} perform extensive human evaluations on the generated images from text input in terms of the photorealism, caption similarity, and sample diversity.

In the meanwhile, the cross-modality generation for visual data often takes certain conditional information from other modality as input (\textit{e.g.}, the text-to-image generation), in which case we hope the model captures the connections between multiple modalities and generate coherent data align with input.  
To this end, the ClipScore~\cite{hessel2021clipscore} is used as an objective metric to measure the relevance between the given text description and the synthesized image~\cite{yezhu2022cdcd,tang2022improved}.  
The ClipScore works by considering the similarity on the feature level of a text and image pair via the pre-trained CLIP model~\cite{radford2021clip}.
In other conditional generative setting such as class-conditioned image generations, existing works also adopt the pre-tarined classifier-based scores for measuring whether the synthesized image well corresponds to the given class label~\cite{liu2022compositional}.


\subsection{Audio Data Evaluation}
The evaluation of the generated audio data differs in terms of the specific audio type.

For the music audio, the general quality of the synthesized music often relies on subjective evaluations with human testers~\cite{melgan,yezhu2022quantizedgan,di2021video,jukebox}. Some other metrics adapted from the digital signal processing field can also be used for the same purpose, such as the Signal-to-Noise Ratio (SNR), which quantifies the strength of the desired audio signal relative to the background noise. 
However, in the context of input-output faithfulness under the cross-modality setting, recent works propose several objective quantitative metrics as evaluation.
Beats coverage and beats hit scores are used in~\cite{yezhu2022quantizedgan,yezhu2022cdcd} to measure the consistency of the musical rhythms in respect to the given video input for the human body dance motions.  
A classifier-based retrieval accuracy score is adopted for considering the diversity of the genera\cite{yezhu2022quantizedgan}.
In~\cite{wu2020jazz}, the Pitch Class Histogram Entropy is another metric for evaluating the music in tonality, the Grooving Pattern Similarity is used to measures the rhythmicity, and the Structureness Indicator is proposed to evaluate the repetitive structure of music audio.

In terms of the speech audio, the evaluation system is quite different from the music area. 
Due to its natural connections with language, the quantitative metrics used are often derived from the speech processing area.
Classic examples include STOI (Short-Time Objective Intelligibility measure)~\cite{taal2010short}, ESTOI (extended STOI)~\cite{jensen2016algorithm}, PESQ (Perceptual Evaluation of Speech Quality)~\cite{rix2001perceptual} and WER (Word Error Rate).
Specifically, STOI and ESTOI imply the quality of speech audio intelligibility, PESQ is an indicator for the perceptual quality, and WER counts the correct work prediction in comparison to the ground truth. Human based Mean Opinion Scores (MOS) test have also been proposed to asses the quality in more recent studies~\cite{tan2022naturalspeech}.

As for the ambient audio, existing works on this topic largely rely on the human based subjective evaluations given specific assessing criteria and instructions for different aspects such as the sound-video alignment and real-or-fake determination. Few works also report the cross-entropy loss as quantitative scores as metric scores~\cite{zhou2018visual}.

\subsection{Textual Data Evaluation}
The evaluations for texts are relatively better developed in the Natural Language Processing field, and many of those metrics are also adopted in assessing the quality of generated texts for generative tasks.

In most cases, the metrics for evaluating the generated textual data rely on the reference text, in other words, we compute the scores by comparing the generated words with the ground truth annotations.
Classic examples of such objective metrics includes the BiLingual Evaluation Understudy (BLEU) scores~\cite{papineni2002bleu}, the Metric for Evaluation of Translation with Explicit ORdering (METEOR)~\cite{banerjee2005meteor}, the Recall-Oriented Understudy for Gisting Evaluation (ROUGE)~\cite{lin2004rouge}, the Consensus-based Image Description Evaluation (CIDEr)~\cite{vedantam2015cider}, and the Semantic Propositional Image Caption Evaluation  (SPICE)~\cite{anderson2016spice}.
The BLEU scores are initially designed for evaluating machine translated texts between different languages, and are believed to have high correlations with human judgements. The most common version of BLEU scores look at the ``n-gram'' precision, which counts the number of matches for the ``n-gram"" words in the generated sentences and the ground truth descriptions. The BLEU scores range from 0 to 1, and a larger value indicates higher similarity to the reference.
The METEOR is another widely used metric in machine translation literature, which additionally considers the alignment between evaluated and reference sentences in terms of harmonic mean of unigram precision and recall. 
Similar to BLEU, the ROUGE scores are proposed to measure the similarity between the translated textual summary and the reference summary by considering the n-grams matches. A ROUGE value close to 0 indicates poor similarity, while the value close to 1 indicates the opposite. 
The CIDEr is a metric specifically proposed for measuring the quality of textual description with regard to the images. The characteristic of the CIDEr metric is that it compares the generated sentences with a set of ground truth sentences written by humans, and thus shows higher agreement with consensus with subjective evaluations performed by human evaluators.  
The SPICE is also designed specifically for the purpose of caption evaluations as the CIDEr. Specifically, this metric takes the semantic propositional content into account, and calculates the score defined over the scene graph. Similar to other metrics, a SPICE value closer to 1 implies better quality of the generated textual sentences with respect to ground truth captions and given images.
In more recent literature, the CLIPScore~\cite{hessel2021clipscore} is proposed as a reference-free metric for text and image related tasks using the distance computation from the feature space of the pre-trained CLIP model~\cite{radford2021clip}.

\end{document}